\documentclass{article}

\usepackage{microtype}
\usepackage{graphicx}
\usepackage{subcaption}
\usepackage{booktabs} 

\usepackage{hyperref}

\usepackage[preprint]{icml2026}

\usepackage{amsmath}
\usepackage{amssymb}
\usepackage{mathtools}
\usepackage{amsthm}

\usepackage[capitalize,noabbrev]{cleveref}

\theoremstyle{plain}

\theoremstyle{definition}

\theoremstyle{remark}

\usepackage[textsize=tiny]{todonotes}

\usepackage{times}
\usepackage{latexsym}

\usepackage[T1]{fontenc}

\usepackage[utf8]{inputenc}

\usepackage{microtype}

\usepackage{inconsolata}

\usepackage{graphicx}

\usepackage{graphicx}
\usepackage{listings}
\usepackage{booktabs}   
\usepackage{tabularx}   
\newcolumntype{Y}{>{\raggedright\arraybackslash}X}  %
\usepackage{inconsolata}
\usepackage{array}
\usepackage{pifont}
\usepackage{tabularx}
\usepackage{adjustbox}
\usepackage{multirow}
\usepackage{caption}
\usepackage{enumitem}
\usepackage{xspace}
\usepackage{tcolorbox}
\usepackage{booktabs,subcaption,amsfonts,dcolumn}
\usepackage{url}
\usepackage{amsmath}
\usepackage{amsthm,amsfonts,amssymb,bm,stmaryrd, bbm}
\usepackage{color,xcolor,colortbl}
\usepackage{CJKutf8}
\usepackage{makecell}
\usepackage{booktabs}
\usepackage{graphicx}
\usepackage{listings}
\usepackage{stfloats}
\usepackage{float}
\usepackage{subcaption}
\usepackage{placeins}
\usepackage{dsfont}
\usepackage{algorithm}
\usepackage{algpseudocode}
\algrenewcommand\algorithmicrequire{\textbf{Input:}}
\algrenewcommand\algorithmicensure{\textbf{Output:}}

\definecolor{warning}{HTML}{C80000}
\definecolor{myyellow}{rgb}{0.98, 0.94, 0.75}
\definecolor{mygreen}{HTML}{538234}
\definecolor{myblue}{HTML}{0070C0}
\definecolor{textblue}{HTML}{4E95D9}
\definecolor{textred}{HTML}{C00000}
\definecolor{mypurple}{RGB}{224, 65, 245}
\definecolor{myorange}{RGB}{209, 136, 17}
\definecolor{Mycolor1}{HTML}{BAD8F2}
\definecolor{Mycolor2}{HTML}{DDEEFA}
\definecolor{adobe}{HTML}{D83220}

\newcommand{\ours}{\textsc{ThinkRouter}}

\newtcbox{\hlsecondarytab}{on line, box align=base, colback=blue!15,colframe=white,size=fbox,arc=3pt, before upper=\strut, top=-2pt, bottom=-4pt, left=-2pt, right=-2pt, boxrule=0pt}

\icmltitlerunning{\ours: Efficient Reasoning via Routing Thinking between Latent and Discrete Spaces}

\begin{document}

\twocolumn[
  \icmltitle{\ours: Efficient Reasoning via Routing\\Thinking between Latent and Discrete Spaces}




  \begin{icmlauthorlist}
    \icmlauthor{Xin Xu$^\dagger$}{yyy}
    \icmlauthor{Tong Yu}{comp}
    \icmlauthor{Xiang Chen}{comp}
    \icmlauthor{Haoliang Wang}{comp}
    \icmlauthor{Julian McAuley}{yyy}
    \icmlauthor{Saayan Mitra}{comp}
  \end{icmlauthorlist}

  \icmlaffiliation{yyy}{UC San Diego}
  \icmlaffiliation{comp}{Adobe Research}

  \icmlcorrespondingauthor{Saayan Mitra}{smitra@adobe.com}

  \icmlkeywords{Latent reasoning, ICML}

  \vskip 0.3in
]



\printAffiliationsAndNotice{$^\dagger$<xinxucs@ucsd.edu>}  

\begin{abstract}
Recent work explores latent reasoning to improve reasoning efficiency by replacing explicit reasoning trajectories with continuous representations in a latent space, yet its effectiveness varies across settings.
Analysis of model confidence dynamics under latent reasoning reveals that thinking trajectories ending in incorrect answers contain fewer low-confidence steps than those ending in correct answers.
Meanwhile, we suggest that soft embeddings aggregated by multiple low-confidence thinking alternatives may introduce and propagate noise, leading to high confidence in unreliable reasoning trajectories.
Motivated by these observations, \textbf{\ours}, an inference-time confidence-aware routing mechanism is proposed to avoid high confidence and noise for efficient reasoning.
\ours\ routes thinking to the discrete token space when model confidence is low, and to the latent space otherwise.
Extensive experiments on STEM reasoning and coding benchmarks across diverse large reasoning models demonstrate that \ours\ outperforms explicit CoT, random routing, and latent reasoning baselines in terms of accuracy, achieving an average improvement of 19.70 points in Pass@1, while reducing generation length by up to 15.55\%.
Further comprehensive analysis reveals that \ours\ can calibrate errors arising from explicit CoT and latent reasoning, and accelerates end-of-thinking token generation by globally lowering model confidence.

\end{abstract}

\section{Introduction}

Large language models (LLMs) have demonstrated promising reasoning capabilities to solve complex problems \citep{huang-chang-2023-towards, mmlu_pro}.
A key driver is explicit chain-of-thought (CoT), which emulates human thinking by generating intermediate reasoning trajectories in natural language \citep{cot,cot_mystery,chu-etal-2024-navigate}.
Recent work uses reinforcement learning (RL) to train LLMs to reason with thinking trajectories before giving answers \citep{lrm_survey1, lrm_survey}.
Such reasoning-intensive training produces large reasoning models (LRMs), e.g., OpenAI o1 \citep{openaio1} and Qwen3 \citep{qwen3}, which have demonstrated strong reasoning performance on hard tasks, such as mathematics and coding \citep{claude3modelcard, swebench, gemini}. 
While explicit trajectories improve reasoning accuracy and interpretability, they limit models' expressive bandwidth \citep{survey2}.
Meanwhile, long thinking chains substantially increase inference cost and response latency \citep{l1, longcot_survey}.
These developments highlight the two goals for \textbf{efficient reasoning}, i.e., improving reasoning accuracy while reducing generation length.

To target this goal, recent work has explored LLM reasoning in a latent space, shifting reasoning from discrete tokens to latent representations \citep{survey1}.
For example, Coconut \citep{coconut}, CCoT \citep{ccot}, and LightThinker \citep{lightthinker} construct several soft tokens to represent long thoughts to reduce tokens but require tuning, and their effectiveness varies across settings, where they even drop performance in some cases.
\textit{Soft Thinking} \citep{soft-thinking}, a training-free method by calculating token-probability-weighted soft embeddings, is proposed and can raise the performance ceiling (\S \ref{sec:reasoning_space}). However, the underlying reason for its effectiveness has not fully explored \citep{DBLP:journals/corr/abs-2508-03440}.
Meanwhile, few works \citep{swireasoning} study whether hybrid reasoning between latent space and discrete spaces will help efficient reasoning.

Therefore, we explore training-free LRM reasoning in hybrid reasoning spaces for efficient reasoning in this work.
Since \textit{Soft Thinking} performs much better than explicit CoT, we first analyze latent-only reasoning through LRM confidence dynamics with \textit{Soft Thinking} (\S \ref{sec:pre_motivation}).
The maximum next-token probability is used as the proxy of LRM confidence \citep{DBLP:conf/iclr/HendrycksG17, DBLP:conf/icml/GuoPSW17}. 
We observe that the reasoning trajectories for incorrect answer predictions have fewer low-confidence steps than those for correct answers.
We also hypothesize that if the maximum next-token probability is low, the soft embedding is an aggregation of multiple low-confidence incompatible thinking alternatives, introducing representational noise.
Such noise may propagate and accumulate across successive latent reasoning steps, leading the model to commit to an inadequately supported solution with high confidence.
These observations motivate us to propose a new, efficient reasoning solution that prevents LRMs from becoming highly confident and from representational noise.

Therefore, we propose \textbf{\ours}, an inference-time mechanism that routes LRM thinking between the discrete token space and the latent space based on LRM confidence (\S \ref{sec:method}).
Specifically, for each time step during thinking, when the maximum next-token probability is lower than a routing threshold, \ours\ routes thinking to the discrete space where one next token is sampled to avoid introducing much noise and mitigate confidence.
Otherwise, LRM conducts thinking in the latent space where a probability-weighted soft embedding (following \textit{Soft Thinking}) is calculated.
\ours\ is evaluated (\S \ref{sec:exp}) on LRMs with diverse scales (1.5B - 32B) and architectures (Qwen3 \citep{qwen3} and gpt-oss \citep{gpt-oss}) and datasets with different domains (STEM reasoning and coding).
Extensive experiments illustrate that \textbf{\ours\ outperforms discrete CoT, \textit{Soft Thinking}, and random routing in accuracy}, improving average Pass@1 by up to 19.70 points,  while reducing generation length comparably with the baselines.
Moreover, we analyze the underlying reason for \ours's effectiveness.
\ours\ can correct errors from explicit CoT and \textit{Soft Thinking}.
\ours\ can also increase the ratio of low-confidence time steps during thinking, indicating that \ours\ can prevent LRMs from becoming highly confident in incorrect solutions to improve accuracy.
Meanwhile, we find that the steps immediately preceding the end-of-thinking (EOT) token generation are characterized by sharply declining or relatively low confidence, suggesting that confidence mitigation \ours\ brings can accelerate the triggering of the EOT token. 
As a result, \ours\ effectively shortens the generation length.
Overall, \ours\ is a simple yet effective method for efficient reasoning.
Our contributions in this work are:
\begin{itemize}
    \item LRM confidence dynamics under \textit{Soft Thinking} (latent-only reasoning) is explored. We observe that incorrect predictions involve fewer low-confidence time steps than correct ones during thinking, and hypothesize that soft embeddings aggregated from multiple low-confidence alternatives are noisy. These motivate us to avoid high confidence and noise to improve reasoning.
    \item We introduce \ours, an inference-time mechanism for efficient reasoning by routing thinking between a discrete token space and a latent space based on LRM confidence.
    \item The extensive experiments show that \ours\ outperforms explicit CoT, random routing, and \textit{Soft thinking}, achieving much accuracy gains across models and tasks, and competitive generation length reduction.
    \item Comprehensive analysis reveals that \ours\ can perform corrective calibration for errors arising from explicit CoT and latent reasoning with \textit{Soft Thinking}, and accelerates end-of-thinking token generation by globally lowering model confidence.
\end{itemize}

\section{Related Work}
\paragraph{Latent Reasoning}
Recent work explores moving reasoning from discrete CoT tokens to latent thoughts.
Coconut \citep{coconut} uses the last hidden states as input embeddings and tunes models to reason with several soft tokens that represent longer reasoning steps to reduce tokens.
However, there is a mismatch between the last hidden states and the input embeddings.
Some works, such as \citet{DBLP:journals/corr/abs-2311-01460}, CCoT~\citep{ccot}, LaRS~\citep{lars}, SoftCoT~\citep{softcot}, CODI~\citep{codi}, CoT2~\citep{cot2}, and SIM-COT~\citep{SIM-CoT}, construct a mapping or use distillation to learn compressed latent thoughts from discrete chain-of-thoughts to align them and effectively internalize step-by-step traces into dense representations.
CoLaR~\citep{CoLaR} and \citet{soft_token_hard_truth} use RL to optimize latent reasoning behavior.
\citet{loop1, loop2} use looped language models to scale latent reasoning by iteratively refining hidden states with shared parameters.
These latent-reasoning methods improve LRM reasoning efficiency but rely on costly training or distillation.
Soft Thinking~\citep{soft-thinking} tries training-free latent reasoning by calculating a next-token-probability-weighted soft embedding.
However, few works explore the underlying mechanism of latent reasoning~\citep{DBLP:journals/corr/abs-2505-12514}.

\paragraph{Hybrid Reasoning}
Recent work investigates hybrid reasoning, where LLM reasoning performs among different thinking modes. HRPO \citep{HRPO} enables switching between latent and discrete reasoning through an RL-learned gating mechanism. In parallel, Thinkless \citep{thinkless}, AdaptThink \citep{zhang-etal-2025-adaptthink, adaptthink}, LHRM \citep{LHRM}, \citet{qiao-etal-2025-agentic}, and MixReasoning \cite{mixreasoning} learn policies that decide long thinking or non-thinking. 
However, all of these methods rely on RL or additional training to acquire the hybrid reasoning behaviors. 
To date, there is few training-free mechanism that can explicitly control whether reasoning is carried out in latent space or discrete space at inference time \citep{swireasoning}.

\begin{figure*}[htbp]
    \centering
    \begin{subfigure}[b]{0.41\textwidth}
        \centering
        \includegraphics[width=\textwidth]{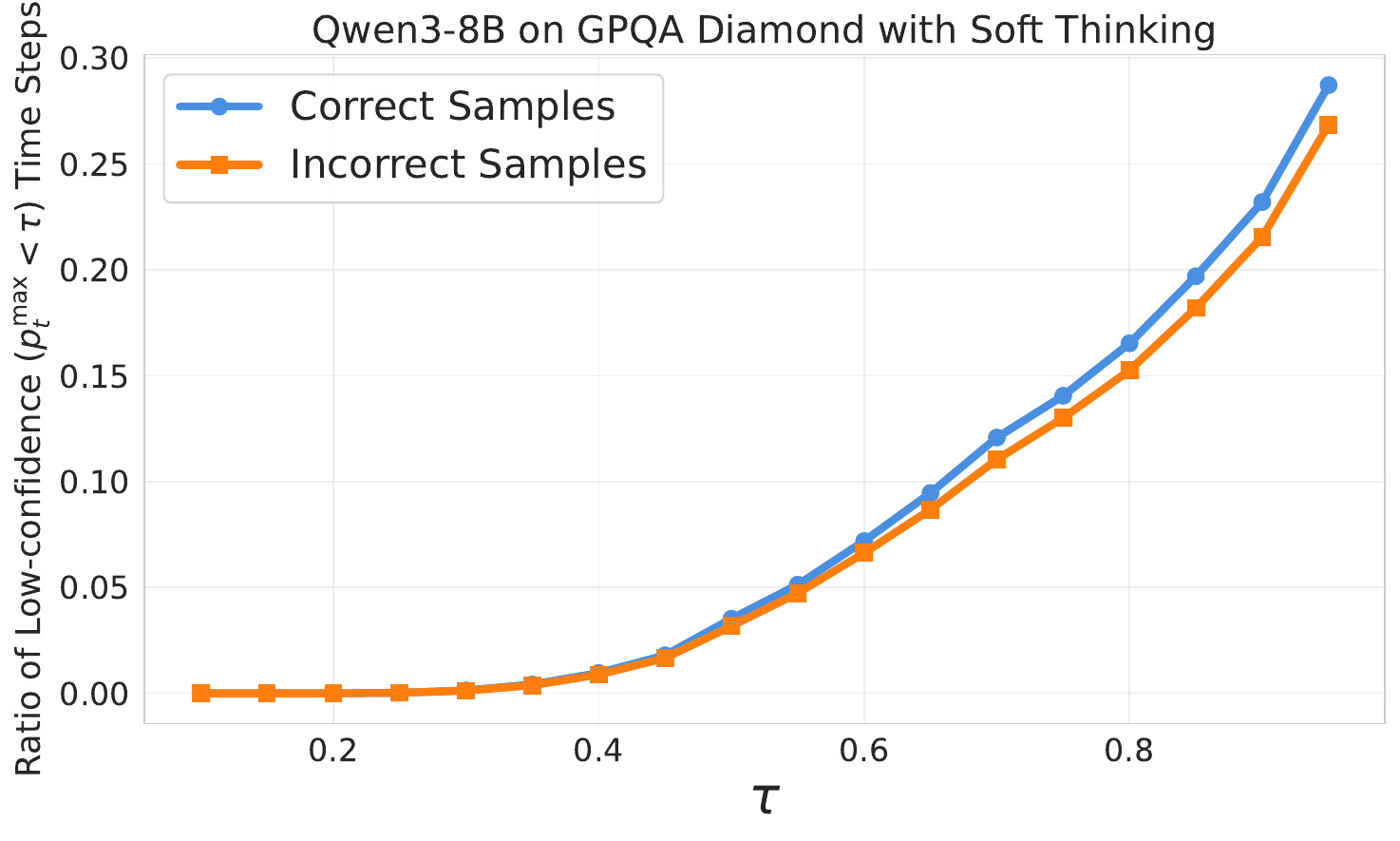}
    \end{subfigure}
    \hspace{0.05\textwidth}
    \begin{subfigure}[b]{0.41\textwidth}
        \centering
        \includegraphics[width=\textwidth]{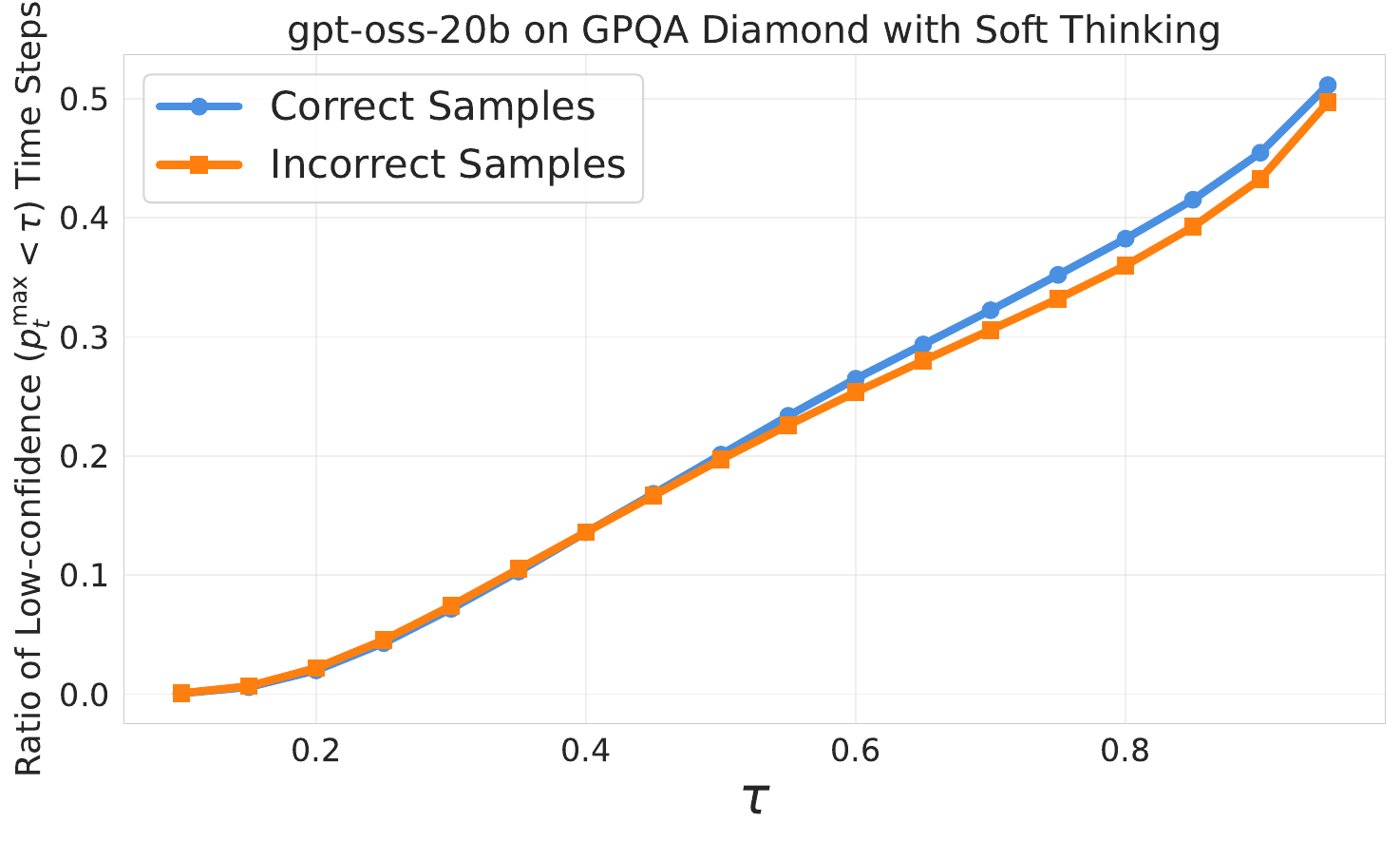}
    \end{subfigure}
    \\
    \caption{Ratio of low-confidence time steps ($p_t^{\max}<\tau$) within reasoning trajectories under \textit{Soft Thinking} (latent-only reasoning) on GPQA Diamond. The incorrect predictions are associated with fewer low-confidence thinking time steps than the correct predictions.
    }
    \label{fig:low_confidence_ratio}
\end{figure*}

\section{Preliminary}

\subsection{LRM Reasoning in Discrete and Latent Spaces}
\label{sec:reasoning_space}
Given an input query $x_{1:Q} = \{x_1, x_2, \ldots, x_Q\}$, a large reasoning model $\mathcal{M}$ first has a thinking process by producing a reasoning trajectory $r_{1:M} = \{r_1, r_2, \ldots, r_M\}$ until a special end-of-thinking (EOT) token is generated.
And then it generates a final answer $y_{1:N} = \{y_1, y_2, \ldots, y_N\}$.
We denote $\mathcal{V}$ as the LRM's vocabulary of size $|\mathcal{V}|$, and $E  \in \mathbb{R}^{|\mathcal{V}|\times d}$ as the token embeddings.
For each token $v \in \mathcal{V}$, its embedding is $E[v] \in \mathbb{R}^d$.

\paragraph{Reasoning in a Discrete Space}
At each time step $t$ within thinking, a discrete token $r_{t}$ is sampled from the next-token probability distribution $p_t$ (after temperature scaling) over the vocabulary, conditioned on the input and previous generated tokens:
\begin{equation}
    r_{t}\sim p_t = \text{LRM}(E[x_{1:Q}],E[r_{1:t-1}]) \in \Delta^{|\mathcal{V}|-1}
\end{equation}

\paragraph{Reasoning in a Latent Space}
During thinking, a soft token embedding $\tilde{e}_t$ is calculated at each time step $t$.
In this work, we follow \textit{Soft Thinking} \citep{soft-thinking} to calculate a probability-weighted soft token embedding using the top-$j$ probabilities:
\begin{equation}
\label{eq:2}
    \hat{p}_t = \text{\textsc{Sample}}(p_t), p_t = \text{LRM}(E[x_{1:Q}], \tilde{e}_{1:t-1}) \in \Delta^{|\mathcal{V}|-1}
\end{equation}
\begin{equation}
\label{eq:3}
    \mathcal V_t^{\text{top-j}} = \operatorname{Top-J}(\hat{p}_t)
\end{equation}
\begin{equation}
\label{eq:soft_embedding}
    \tilde{e}_{t} = \sum_{v \in \mathcal{V}_t^{\text{top-j}}} \tilde{p}_t[v] E[v] \in \mathbb{R}^d,\ \tilde{p}_t[v] = \frac{\hat{p}_t[v]}{\sum_{u \in V_t^{\text{top-j}}} \hat{p}_t[u]}
\end{equation}
\begin{equation}
    \tilde{e}_{1:t} := \tilde{e}_{1:t-1} \| \tilde{e}_{t}
\end{equation}
where \textsc{Sample} is the sampling operation that applies top-k, top-p, and min-p filtering with renormalization (Appendix \ref{app:baseline}), and $\operatorname{Top-J}(p_t)$ is the set of tokens with top-$j$ highest probabilities under the distribution $p_t$. 
Unlike reasoning in a discrete space, which collapses the probability mass onto one single token, thereby committing to one explicit reasoning path, latent reasoning operates in a latent space, allowing LRMs to integrate multiple potential reasoning paths in parallel, thereby maintaining the information over possible thoughts \citep{coconut, soft-thinking}.

After thinking, the model generates a final answer $y$ with standard decoding in a discrete space:
\begin{equation}
     y_{t}\sim p_t \\ = \text{LRM}(E[x_{1:Q}],E[r_{1:M}], E[y_{1:t-1}]) \in \Delta^{|\mathcal{V}|-1}
\end{equation}

\subsection{Model Confidence with Latent Reasoning}
\label{sec:pre_motivation}
We first examine LRM behaviors to figure out the difference between correct and incorrect generations under latent-only reasoning with \textit{Soft Thinking}, which outperforms explicit CoT.
Specifically, we analyze whether LRMs exhibit systematically different confidence patterns within reasoning trajectories when producing correct versus incorrect answers. 
Two LRMs from different families and scales (Qwen3-8B~\citep{qwen3} and gpt-oss-20b~\citep{gpt-oss}) are evaluated on two representative reasoning tasks spanning different domains: STEM reasoning with GPQA Diamond~\citep{gpqa} and code generation with HumanEval~\citep{humaneval}. 
We conduct \textit{Soft Thinking} and record the next-token probability distribution over the vocabulary at each time step. 
The maximum next-token probability, i.e. $p_t^{\max} = \max\limits_{v\in \mathcal{V}}p_t[v]$ as a proxy for the model confidence is monitored \citep{DBLP:conf/icml/GalG16, DBLP:journals/tacl/JiangXAN20, DBLP:journals/tacl/VashurinFVRVTPXSGPBNPS25}. 
Figure \ref{fig:low_confidence_ratio} and \ref{fig:low_ratio_gptoss} report the ratios of low-confidence time steps ($p_t^{\max} <\tau$) in reasoning trajectories separately for correct and incorrect answers LRMs generate, where $\tau \in [0.1, 0.95]$ (Appendix \ref{app:pre_exp}).

\begin{figure*}[htbp]
    \centering    \includegraphics[width=0.93\linewidth]{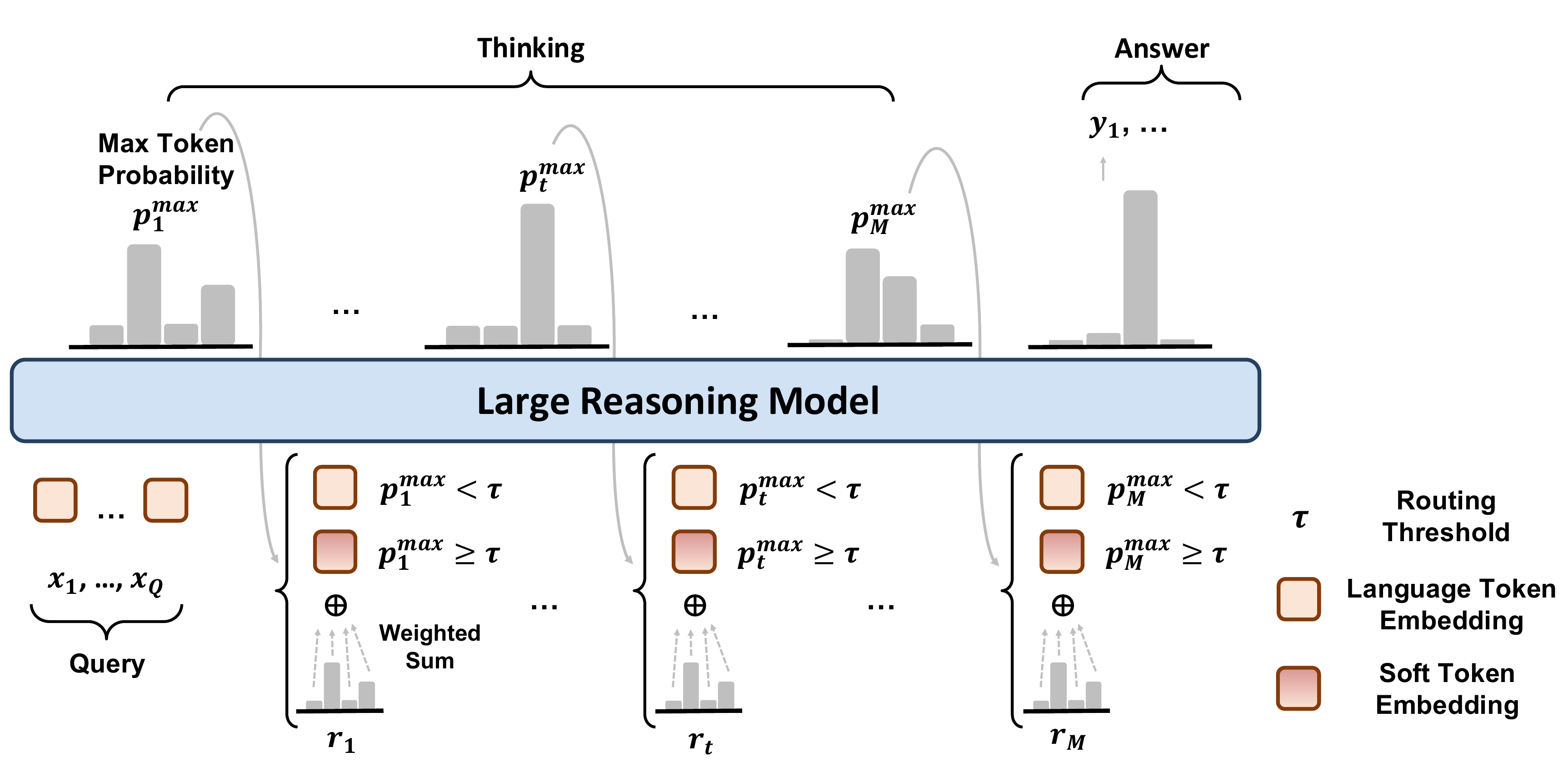}
    \caption{Overview of \ours. During thinking, when the maximum next-token probability $p_t^{\max}$ is lower than the routing threshold $\tau$, \ours\ routes thinking to the discrete token space; otherwise, \ours\ routes thinking to the latent space by calculating a probability-weighted soft embedding.}
    \label{fig:method}
\end{figure*}

From these figures, we observe that when LRMs ultimately produce an incorrect answer, their reasoning trajectories contain fewer low-confidence steps than those ending in a correct answer in most cases, especially when $\tau \in [0.4, 0.9]$, indicating an association between unsuccessful reasoning and relatively high confidence.
This gives us insights into mitigating high confidence to improve reasoning.
Meanwhile, it is found that reasoning trajectories of correct predictions often maintain steps with relatively low $p_t^{\max}$.
A relatively low $p_t^{\max}$ indicates that $p_t$ assigns comparable probability mass to multiple alternative continuations. 
In such cases, the soft embedding is formed by aggregating these low-confidence alternatives.
However, we hypothesize that these alternatives may correspond to distinct or even mutually incompatible reasoning directions.
Aggregating such alternatives may yield a representation that is semantically diffuse or poorly grounded, potentially introducing noise \citep{DBLP:conf/iclr/AroraLM17,DBLP:conf/iclr/YangDSC18, DBLP:conf/nips/LimEHM21, DBLP:conf/emnlp/HaoGMHWWH23,self-consistency}.
We further conjecture that such noise may accumulate over successive latent reasoning steps.
As such representations are propagated in the latent space, a LRM can drift toward spurious or incoherent reasoning, eventually assigning higher confidence to directions that are not well supported, which provides a potential explanation for why LRMs tend to have fewer low-confidence steps in reasoning trajectories of incorrect answers \cite{DBLP:conf/nips/BengioVJS15, DBLP:conf/emnlp/Schmidt19, DBLP:journals/tois/HuangYMZFWCPFQL25, DBLP:journals/corr/abs-2509-06770, DBLP:journals/corr/abs-2509-04664, DBLP:journals/corr/abs-2505-13143}.
Overall, these observations reveal potential failure modes that limit efficient reasoning and motivate us to prevent LRMs from high confidence and avoid integrating multiple low-confidence alternatives.

\section{\ours}
\label{sec:method}

Therefore, we propose \textbf{\ours}, an inference-time mechanism by routing thinking between discrete and latent spaces based on LRM confidence, as shown in Figure \ref{fig:method}.
Specifically, \ours\ prevent unreliable thinking alternatives from being jointly explored in a latent space.
At each time step $t$ within thinking, \ours\ determines the reasoning space based on the maximum next-token probability under the next-token probability distribution:
\begin{equation}
\begin{array}{c}
\text{Thinking} \\
\text{Space}
\end{array}
\begin{cases}
\text{Discrete Token Space with }r_t & \text{if } p_t^{\max} < \tau, \\[6pt]
\displaystyle \text{Latent Space with }\tilde{e}_{t} & \text{if } p_t^{\max} \ge \tau,
\end{cases}
\end{equation}
If the maximum next-token probability $p_t^\text{max} < \tau$, where $\tau$ is a routing threshold, indicating that all alternatives are of low confidence, the thinking operates with one discrete token to avoid aggregating multiple incompatible or noisy thinking alternatives and prevent LRMs from becoming highly confident by committing to a low-confidence alternative.
Conversely, when $p_t^{\max} \ge \tau$, reasoning proceeds in the latent space where a soft token embedding $\tilde{e}_{t}$ is calculated to represent a mixture of multiple plausible reasoning paths, allowing richer exploration in the latent concept space following \textit{Soft Thinking}. 
Overall, LRM reasoning with \ours\ is implemented as Algorithm \ref{alg:thinkrouter} (To facilitate presentation, we omit the case where the model outputs exceed the LRM's maximum generation length). More details, such as \textsc{ColdStop}, \textsc{MultinomialSample}, \textsc{Decode}, etc., are described in Appendix \ref{app:algorithm}.

\begin{algorithm}[ht!]
\caption{\ours}
\label{alg:thinkrouter}
\begin{algorithmic}[1] 
\Require Query $x_{1:Q}$, LRM $\mathcal{M}$, Routing Threshold $\tau$
\Ensure Answer $y$
\State $\mathcal{R} \gets [\,]$ \Comment{Embeddings}
\While{true}    \Comment{Thinking}
    \State $p_t \gets \mathcal{M}(E[x_{1:Q}],\, \mathcal{R})$   \Comment{Temperature Scaling}
    \State $p_t^{\max} \gets \max_{v\in \mathcal{V}} p_t[v]$
    \State $\hat{p}_t \gets$ \textsc{Sample}($p_t$) \Comment{Top-k/Top-p/Min-p Renorm}
    \If{$p_t^{\max}<\tau$} \Comment{Discrete Space}
        \State $r_t \gets \textsc{MultinomialSample}(\hat{p}_t)$ 
        \State $\mathcal{R} \gets \mathcal{R} \,\Vert\, E[r_t]$
    \Else \Comment{Latent Space}
        \State $\mathcal V_t^{\text{top-j}} = \operatorname{Top-J}(\hat{p}_t)$ \Comment{Eq. (\ref{eq:3})}
        \State $\tilde{p}_t[v] = \frac{\hat{p}_t[v]}{\sum_{u \in V_t^{\text{top-j}}} \hat{p}_t[u]}$
        \State $\tilde{e}_{t} = \sum_{v \in \mathcal{V}_t^{\text{top-j}}} \tilde{p}_t[v] E[v] \in \mathbb{R}^d$ \Comment{Eq. (\ref{eq:soft_embedding})}
        \State $\mathcal{R} \gets \mathcal{R} \,\Vert\, \tilde e_t$
        \State $r_t \leftarrow \arg\max_{v \in \mathcal{V}} \hat{p}_t[v]$
    \EndIf
    \If{\textsc{ColdStop}($p_t$)}
        \State $r_t = \textit{EOT Token}$
        \State \textbf{break}
    \EndIf
    \If{$r_t = \textit{EOT Token}$}
        \State \textbf{break}
    \EndIf
\EndWhile
\State $y \gets \textsc{Decode}(\mathcal{M}, E[x_{1:Q}], \mathcal{R})$
\State \Return $y$
\end{algorithmic}
\end{algorithm}

\section{Experiments and Results}
\label{sec:exp}
\subsection{Setups}
\label{sec:experi_setup}
\paragraph{Datasets and Metrics}
\ours~ is comprehensively evaluated on five reasoning benchmarks in different domains, including $i)$ STEM reasoning: AIME 2024 \citep{aime}, AIME 2025 \citep{aime}, and GPQA Diamond \citep{gpqa}, and $ii)$ coding: HumanEval \citep{humaneval}, and MBPP \citep{mbpp}.
The details are in Appendix \ref{app:datasets}.
Pass@1 is used as the accuracy of the models' generated answers, where Pass@$k=1 -\binom{n-c}{k}/\binom{n}{k}$. 
We set $k=1$ so that Pass@1$=\frac{c}{n}$, following \textit{Soft Thinking}.
For each sample, the LRM generates $n=3$ candidate answers with three seeds $\{0,7,42\}$.
$c$ is the number of correct answers.
We report the average Pass@1 over all samples in each test set.
Meanwhile, we use the token number of thinking trajectories and final answer outputs as the generation length and report the average generation length across all samples in each test set to measure token costs.

\paragraph{Models}
We select four large reasoning models, Qwen3-(1.7B, 8B, 32B) \citep{qwen3}, and gpt-oss-20b \citep{gpt-oss}, with different model scales and architectures to evaluate our method, which aims to illustrate the generality and robustness of \ours.

\paragraph{Baselines}
\ours\ are compared with four baselines for evaluation.
Following \textit{Soft Thinking} \citep{soft-thinking}, we apply two baselines in the discrete token space: CoT with the sampling strategies and with greedy decoding.
\textit{Soft Thinking} serves as the baseline in the latent space.
We also include a Random Routing baseline for sanity check, which randomly selects the thinking space at each time step to assess the effectiveness of the confidence-aware mechanism.
More details are described in Appendix \ref{app:baseline}.
\paragraph{Implementation}
All experiments are implemented with SGLang \citep{sglang} and NVIDIA H100 80G GPUs, following \textit{Soft Thinking}. 
For the routing threshold $\tau$, we use 10 samples randomly selected from each dataset as a validation set and perform a grid search within \{0.4, 0.5, 0.6, 0.7, 0.8, 0.9\} on the validation set to find the optimal $\tau$. Then we conduct evaluation with the optimal $\tau$ on the rest samples in each dataset (More details are in Appendix \ref{app:implementation}).

Tables \ref{tab:math} and \ref{tab:code} report \ours's performance on STEM reasoning and coding benchmarks, respectively. 
For each benchmark, the metrics are calculated over the remaining samples after excluding the 10 samples used for grid search.
To ensure the robustness of our evaluation, we also report results on all samples of each benchmark in Tables \ref{tab:math_all} and \ref{tab:code_all}, provided in Appendix \ref{app:fulldataset}.

\begin{table*}[ht]
  \centering
  \caption{Pass@1 (\%) and generation length on \textbf{STEM reasoning} benchmarks for \ours\ and the baselines across different models. \textcolor{textblue}{Blue} and \textcolor{textred}{red} values indicate performance \textcolor{textblue}{improvements} and \textcolor{textred}{degradations}, respectively, relative to CoT (sampling). \textbf{Bold} values highlight the best-performing baseline within the same model and benchmark setting.}
  \scalebox{0.70}{
    \begin{tabular}{l|cccccccc|ccccc}
    \toprule
          & \multicolumn{8}{c}{\textbf{Pass@1 (\%)\ $\uparrow$}}                    & \multicolumn{5}{c}{\textbf{Generation Length\ $\downarrow$}} \\
\cmidrule{2-14}          & \multicolumn{2}{c}{\textbf{AIME 2024}} & \multicolumn{2}{c}{\textbf{AIME 2025}} & \multicolumn{2}{c}{\textbf{GPQA Diamond}} & \multicolumn{2}{c}{\textbf{Average}} & \textbf{AIME 2024} & \textbf{AIME 2025} & \textbf{GPQA Diamond} & \multicolumn{2}{c}{\textbf{Average}}   \\
    \midrule
          & \multicolumn{13}{c}{\textbf{Qwen3-1.7B}} \\
    \midrule
    CoT (sampling) & 46.67 &       & 25.00 &       & 38.48 &       & 36.71 &       & 18433.02 & 19146.13 & \textbf{8601.55} & 15393.57 &  \\
    CoT (greedy) & 63.33 & \textcolor{textblue}{$\uparrow$16.67} & 26.67 & \textcolor{textblue}{$\uparrow$1.67}  & 35.28 & \textcolor{textred}{$\downarrow$3.19} & 41.76 & \textcolor{textblue}{$\uparrow$5.05}  & 19189.85 & 21794.50 & 12583.13 & 17855.83 & \textcolor{textred}{$\uparrow$16.00\%} \\
    \textit{Soft Thinking} & 55.00 & \textcolor{textblue}{$\uparrow$8.33}  & 43.33 & \textcolor{textblue}{$\uparrow$18.33} & 43.26 & \textcolor{textblue}{$\uparrow$4.79}  & 47.20 & \textcolor{textblue}{$\uparrow$10.48} & 17424.30 & 18835.50 & 9076.58 & 15112.13 & \textcolor{textblue}{$\downarrow$1.83\%} \\
    Random Routing & 60.00 & \textcolor{textblue}{$\uparrow$13.33} & 38.33 & \textcolor{textblue}{$\uparrow$13.33} & 44.33 & \textcolor{textblue}{$\uparrow$5.85}  & 47.55 & \textcolor{textblue}{$\uparrow$10.84} & 17577.37 & 20223.57 & 9134.88 & 15645.27 & \textcolor{textred}{$\uparrow$1.64\%} \\
    \textbf{\ours} & \textbf{71.67} & \textcolor{textblue}{\textbf{$\uparrow$25.00}} & \textbf{51.67} & \textcolor{textblue}{\textbf{$\uparrow$26.67}} & \textbf{45.92} & \textcolor{textblue}{\textbf{$\uparrow$7.45}}  & \textbf{56.42} & \textcolor{textblue}{\textbf{$\uparrow$19.70}} & \textbf{15863.13} & \textbf{18504.38} & 8899.38 & \textbf{14422.30} & \textcolor{textblue}{\textbf{$\downarrow$6.31\%}} \\
    \midrule
          & \multicolumn{13}{c}{\textbf{Qwen3-8B}} \\
    \midrule
    CoT (sampling) & 76.67 &       & 71.67 &       & 59.04 &       & 69.13 &       & 14138.82 & 20042.08 & 8285.81 & 14155.57 &  \\
    CoT (greedy) & \textbf{86.67} & \textcolor{textblue}{\textbf{$\uparrow$10.00}} & \textbf{81.67} & \textcolor{textblue}{\textbf{$\uparrow$10.00}} & 60.64 & \textcolor{textblue}{$\uparrow$1.60}  & 76.32 & \textcolor{textblue}{$\uparrow$7.20}  & 15474.35 & 20388.60 & 10605.57 & 15489.51 & \textcolor{textred}{$\uparrow$9.42\%} \\
    \textit{Soft Thinking} & 85.00 & \textcolor{textblue}{$\uparrow$8.33}  & 75.00 & \textcolor{textblue}{$\uparrow$3.33}  & 62.94 & \textcolor{textblue}{$\uparrow$3.90}  & 74.31 & \textcolor{textblue}{$\uparrow$5.19}  & \textbf{13338.65} & 19297.30 & 8041.19 & 13559.05 & \textcolor{textblue}{$\downarrow$4.21\%} \\

    Random Routing & 85.00 & \textcolor{textblue}{$\uparrow$8.33}  & 78.33 & \textcolor{textblue}{$\uparrow$6.67}  & 65.96 & \textcolor{textblue}{$\uparrow$6.91}  & 76.43 & \textcolor{textblue}{$\uparrow$7.30}  & 14854.15 & 19978.65 & 8778.94 & 14537.25 & \textcolor{textred}{$\uparrow$2.70\%} \\
    
    \textbf{\ours} & \textbf{86.67} & \textcolor{textblue}{\textbf{$\uparrow$10.00}} & 80.00 & \textcolor{textblue}{$\uparrow$8.33}  & \textbf{74.82} & \textcolor{textblue}{\textbf{$\uparrow$15.78}} & \textbf{80.50} & \textcolor{textblue}{\textbf{$\uparrow$11.37}} & 13661.87 & \textbf{18756.07} & \textbf{5470.79} & \textbf{12629.57} & \textcolor{textblue}{\textbf{$\downarrow$10.78\%}} \\
    \midrule
           & \multicolumn{13}{c}{\textbf{Qwen3-32B}} \\
    
\midrule    CoT (sampling) & 76.67 &       & 78.33 &       & 66.67 &       & 73.89 &       & 12508.73 & 17758.15 & 5733.72 & 12000.20 &  \\
    CoT (greedy) & 75.00 & \textcolor{textred}{$\downarrow$1.67} & 76.67 & \textcolor{textred}{$\downarrow$1.67} & 65.78 & \textcolor{textred}{$\downarrow$0.89} & 72.48 & \textcolor{textred}{$\downarrow$1.41} & 12809.85 & \textbf{16162.40} & 8264.37 & 12412.21 & \textcolor{textred}{$\uparrow$3.43\%} \\
    \textit{Soft Thinking} & \textbf{91.67} & \textcolor{textblue}{\textbf{$\uparrow$15.00}} & 78.33 & 0.00  & 72.87 & \textcolor{textblue}{$\uparrow$6.21}  & 80.96 & \textcolor{textblue}{$\uparrow$7.07}  & 11890.85 & 17573.90 & 5671.31 & 11712.02 & \textcolor{textblue}{$\downarrow$2.40\%} \\
    Random Routing & 90.00 & \textcolor{textblue}{$\uparrow$13.33} & 80.00 & \textcolor{textblue}{$\uparrow$1.67}  & 75.18 & \textcolor{textblue}{$\uparrow$8.51}  & 81.73 & \textcolor{textblue}{$\uparrow$7.84}  & 11698.17 & 18147.18 & 5845.89 & 11897.08 & \textcolor{textblue}{$\downarrow$0.86\%} \\
    \textbf{\ours} & \textbf{91.67} & \textcolor{textblue}{\textbf{$\uparrow$15.00}} & \textbf{86.67} & \textcolor{textblue}{\textbf{$\uparrow$8.33}} & \textbf{76.42} & \textcolor{textblue}{\textbf{$\uparrow$9.75}}  & \textbf{86.58} & \textcolor{textblue}{\textbf{$\uparrow$12.70}} & \textbf{11810.12} & 16208.12 & \textbf{5590.69} & \textbf{11202.97} & \textcolor{textblue}{\textbf{$\downarrow$6.64\%}} \\
    \midrule
    & \multicolumn{13}{c}{\textbf{gpt-oss-20b}} \\
    \midrule
    CoT (sampling) & 78.33 &       & 73.33 &       & 64.18 &       & 71.95 &       & 10293.37 & 14243.98 & 4265.47 & 9600.94 &  \\
    CoT (greedy) & 76.67 & \textcolor{textred}{$\downarrow$1.67} & 73.33 & 0.00  & 66.84 & \textcolor{textblue}{$\uparrow$2.66}  & 72.28 & \textcolor{textblue}{$\uparrow$0.33}  & 13524.60 & 17569.45 & 8316.85 & 13136.97 & \textcolor{textred}{$\uparrow$36.83\%} \\
    \textit{Soft Thinking} & 75.00 & \textcolor{textred}{$\downarrow$3.33} & 70.00 & \textcolor{textred}{$\downarrow$3.33} & 65.25 & \textcolor{textblue}{$\uparrow$1.06}  & 70.08 & \textcolor{textred}{$\downarrow$1.87} & \textbf{5769.30} & \textbf{5381.90} & 3247.56 & \textbf{4799.59} & \textcolor{textblue}{\textbf{$\downarrow$50.01\%}} \\
    Random Routing & 93.33 & \textcolor{textblue}{$\uparrow$15.00} & 80.00 & \textcolor{textblue}{$\uparrow$6.67}  & 65.25 & \textcolor{textblue}{$\uparrow$1.06}  & 79.53 & \textcolor{textblue}{$\uparrow$7.58}  & 6942.24 & 10609.87 & 3116.52 & 6889.54 & \textcolor{textblue}{$\downarrow$28.24\%} \\
    \textbf{\ours} & \textbf{91.67} & \textcolor{textblue}{$\uparrow$\textbf{13.33}} & \textbf{88.33} & \textcolor{textblue}{$\uparrow$\textbf{15.00}} & \textbf{71.63} & \textcolor{textblue}{$\uparrow$\textbf{7.45}} & \textbf{83.88} & \textcolor{textblue}{$\uparrow$\textbf{11.93}} & 8624.70 & 12762.00 & \textbf{2937.28} & 8107.99 & \textcolor{textblue}{$\downarrow$15.55\%} \\
    \bottomrule
    \end{tabular}}
  \label{tab:math}%
\end{table*}%

\begin{table*}[ht!]
  \centering
  \caption{Pass@1 (\%) and generation length on \textbf{coding} benchmarks for \ours\ and the baselines across different models.}
  \scalebox{0.73}{
    \begin{tabular}{l|cccccc|cccc}
    \toprule
          & \multicolumn{6}{c}{\textbf{Pass@1 (\%) $\uparrow$}}    & \multicolumn{4}{c}{\textbf{Generation Length $\downarrow$}} \\
\cmidrule{2-11}          & \multicolumn{2}{c}{\textbf{HumanEval}} & \multicolumn{2}{c}{\textbf{MBPP}} & \multicolumn{2}{c}{\textbf{Average}} & \textbf{HumanEval} & \textbf{MBPP}  & \multicolumn{2}{c}{\textbf{Average}} \\
    \midrule
          & \multicolumn{10}{c}{\textbf{Qwen3-1.7B}} \\
    \midrule
    CoT (sampling) & 78.57 &       & 74.22 &       & 76.40 &       & 4193.49 & \textbf{3901.11} & 4047.30 &  \\
    CoT (greedy) & 72.73 & \textcolor{textred}{$\downarrow$5.84} & 71.26 & \textcolor{textred}{$\downarrow$2.97} & 71.99 & \textcolor{textred}{$\downarrow$4.41} & 5894.18 & 5629.36 & 5761.77 & \textcolor{textred}{$\uparrow$42.36\%} \\
    \textit{Soft Thinking} & 77.92 & \textcolor{textred}{$\downarrow$0.65} & 72.87 & \textcolor{textred}{$\downarrow$1.35} & 75.40 & \textcolor{textred}{$\downarrow$1.00} & \textbf{3729.96} & 4036.15 & \textbf{3883.06} & \textcolor{textblue}{\textbf{$\downarrow$4.06\%}} \\
    Random Routing & 77.92 & \textcolor{textred}{$\downarrow$0.65} & 72.06 & \textcolor{textred}{$\downarrow$2.16} & 74.99 & \textcolor{textred}{$\downarrow$1.40} & 3878.50 & 4011.40 & 3944.95 & \textcolor{textblue}{$\downarrow$2.53\%} \\
    \textbf{\ours} & \textbf{81.82} & \textcolor{textblue}{\textbf{$\uparrow$3.25}}  & \textbf{75.57} & \textcolor{textblue}{\textbf{$\uparrow$1.35}}  & \textbf{78.70} & \textcolor{textblue}{\textbf{$\uparrow$2.30}}  & 4057.11 & 3913.83 & 3985.47 & \textcolor{textblue}{$\downarrow$1.53\%} \\
    \midrule
          & \multicolumn{10}{c}{\textbf{Qwen3-8B}} \\
    \midrule
    CoT (sampling) & 76.19 &       & 94.06 &       & 85.13 &       & 4066.13 & 3412.00 & 3739.07 &  \\
    CoT (greedy) & 71.43 & \textcolor{textred}{$\downarrow$4.76} & 89.07 & \textcolor{textred}{$\downarrow$4.99} & 80.25 & \textcolor{textred}{$\downarrow$4.88} & 5900.47 & 4975.66 & 5438.06 & \textcolor{textred}{$\uparrow$45.44\%} \\
    \textit{Soft Thinking} & 72.73 & \textcolor{textred}{$\downarrow$3.46} & 91.50 & \textcolor{textred}{$\downarrow$2.56} & 82.11 & \textcolor{textred}{$\downarrow$3.01} & \textbf{3510.87} & \textbf{2894.60} & \textbf{3202.74} & \textcolor{textblue}{\textbf{$\downarrow$14.34\%}} \\
    Random Routing & 77.92 & \textcolor{textblue}{$\uparrow$1.73}  & 94.60 & \textcolor{textblue}{$\uparrow$0.54}  & 86.26 & \textcolor{textblue}{$\uparrow$1.14}  & 3823.40 & 3297.15 & 3560.28 & \textcolor{textblue}{$\downarrow$4.78\%} \\
    \textbf{\ours} & \textbf{79.44} & \textcolor{textblue}{\textbf{$\uparrow$3.25}}  & \textbf{94.47} & \textcolor{textblue}{\textbf{$\uparrow$0.40}}  & \textbf{86.95} & \textcolor{textblue}{\textbf{$\uparrow$1.83}}  & 3704.82 & 3162.93 & 3433.88 & \textcolor{textblue}{$\downarrow$8.16\%} \\
    \midrule
          & \multicolumn{10}{c}{\textbf{Qwen3-32B}} \\
    \midrule
    CoT (sampling) & 67.32 &       & 95.14 &       & 81.23 &       & 3558.52 & 2623.08 & 3090.80 &  \\
    CoT (greedy) & \textbf{72.08} & \textcolor{textblue}{\textbf{$\uparrow$4.76}}  & 95.55 & \textcolor{textblue}{$\uparrow$0.40}  & \textbf{83.81} & \textcolor{textblue}{\textbf{$\uparrow$2.58}}  & 3662.56 & 2709.68 & 3186.12 & \textcolor{textred}{$\uparrow$3.08\%} \\
    \textit{Soft Thinking} & 69.48 & \textcolor{textblue}{$\uparrow$2.16}  & 96.36 & \textcolor{textblue}{$\uparrow$1.21}  & 82.92 & \textcolor{textblue}{$\uparrow$1.69}  & 3573.73 & \textbf{2419.82} & \textbf{2996.78} & \textcolor{textblue}{\textbf{$\downarrow$3.04\%}} \\
    Random Routing & 69.91 & \textcolor{textblue}{$\uparrow$2.60}  & 96.36 & \textcolor{textblue}{$\uparrow$1.21}  & 83.13 & \textcolor{textblue}{$\uparrow$1.91}  & 3582.90 & 2600.39 & 3091.65 & \textcolor{textred}{$\uparrow$0.03\%} \\
    \textbf{\ours} & 69.40 & \textcolor{textblue}{$\uparrow$2.08}  & \textbf{96.49} & \textcolor{textblue}{\textbf{$\uparrow$1.35}}  & 82.94 & \textcolor{textblue}{$\uparrow$1.72}  & \textbf{3511.16} & 2521.86 & 3016.51 & \textcolor{textblue}{$\downarrow$2.40\%} \\
    \midrule
    & \multicolumn{10}{c}{\textbf{gpt-oss-20b}} \\
    \midrule
    CoT (sampling) & 75.97 &       & 97.17 &       & 86.57 &       & 1054.40 & 895.63 & 975.02 &  \\
    CoT (greedy) & 72.73 & \textcolor{textred}{$\downarrow$3.25} & 95.95 & \textcolor{textred}{$\downarrow$1.21} & 84.34 & \textcolor{textred}{$\downarrow$2.23} & 1929.97 & 1433.84 & 1681.90 & \textcolor{textred}{$\uparrow$72.50\%} \\
    \textit{Soft Thinking} & 79.00 & \textcolor{textblue}{$\uparrow$3.03}  & 94.33 & \textcolor{textred}{$\downarrow$2.83} & 86.78 & \textcolor{textblue}{$\uparrow$0.21}  & \textbf{958.12} & 815.64 & 886.88 & \textcolor{textblue}{$\downarrow$9.04\%} \\
    Random Routing & 78.22 & \textcolor{textblue}{$\uparrow$2.25}  & 96.22 & \textcolor{textred}{$\downarrow$0.94} & 87.22 & \textcolor{textblue}{$\uparrow$0.65}  & 967.73 & \textbf{690.38} & \textbf{829.06} & \textcolor{textblue}{\textbf{$\downarrow$14.97\%}} \\
    \textbf{\ours} & \textbf{79.22} & \textcolor{textblue}{\textbf{$\uparrow$3.25}}  & 96.09 & \textcolor{textblue}{\textbf{$\uparrow$0.92}} & 87.55 & \textcolor{textblue}{\textbf{$\uparrow$0.98}}  & 1017.05 & 781.05 & 899.05 & \textcolor{textblue}{$\downarrow$7.79\%} \\

    \bottomrule
    \end{tabular}}
  \label{tab:code}%
\end{table*}%

\subsection{Main Results}
\label{sec:result}

\paragraph{Pass@1 Accuracy Improvement} 
Across almost all benchmarks, \ours\ outperforms the baselines.
For STEM reasoning as shown in Table \ref{tab:math}, \ours\ yields notable gains.
For example, \ours\ improves the average Pass@1 of CoT (sampling) by up to \textcolor{textblue}{+19.70 points} in Qwen3-1.7B.
Although \textit{Soft Thinking} also improves accuracy in some settings, its effectiveness varies between tasks and models.
Comparatively, \ours\ consistently achieves average Pass@1 gains over \textit{Soft Thinking} of \textcolor{textblue}{9.22, 6.18, 5.63, 13.80} points on Qwen3-1.7B, 8B, 32B, and gpt-oss-20b respectively. 
Importantly, even in cases where \textit{Soft Thinking} reduces the accuracy of CoT, \ours\ remains much more effective.
For instance, for gpt-oss-20b, \textit{Soft Thinking} degrades the Pass@1 by \textcolor{textred}{3.33} points on the difficult AIME 2025 compared to CoT (sampling), whereas \ours\ still achieves \textcolor{textblue}{15.00}-point improvement, indicating that \textit{Soft Thinking} can amplify noise in reasoning trajectories, leading to unreliable answers, while confidence-aware routing can avoid such a failure mode.
A similar pattern is observed on coding benchmarks as shown in Table \ref{tab:code}.
While \textit{Soft Thinking} causes accuracy drops in most cases, \ours\ consistently improves Pass@1.
Moreover, we observe an ordering in Pass@1 performance in most cases: \ours\ > Random Routing > \textit{Soft Thinking}, which suggests that routing is beneficial for efficient reasoning, and incorporating confidence into routing is much more effective.
Overall, these results show that \ours\ yields greater and more robust accuracy gains than the baselines.

\vspace{-3mm}
\paragraph{Generation Length Reduction} \ours\ achieves competitive generation length reduction across all benchmarks and models compared with the baselines.
Especially on STEM reasoning benchmarks, \ours\ reduces the average generation lengths relative to Soft Thinking by \textcolor{textblue}{-4.56\%, -6.86\%, -4.35\%} on Qwen3-1.7B, 8B, 32B, respectively.
On coding benchmarks, \ours\ exhibits comparable generation length reduction performance with \textit{Soft Thinking}, consistently reducing generation length relative to CoT.
Although Random Routing increases the average generation length in some settings, \ours\ always produce shorter outputs than CoT, demonstrating the significance of confidence awareness.

\subsection{Error Calibration}
\label{sec:calibration}
\begin{figure}[ht!]
    \centering
    \begin{subfigure}{0.9\linewidth}
        \centering
        \includegraphics[width=\linewidth]{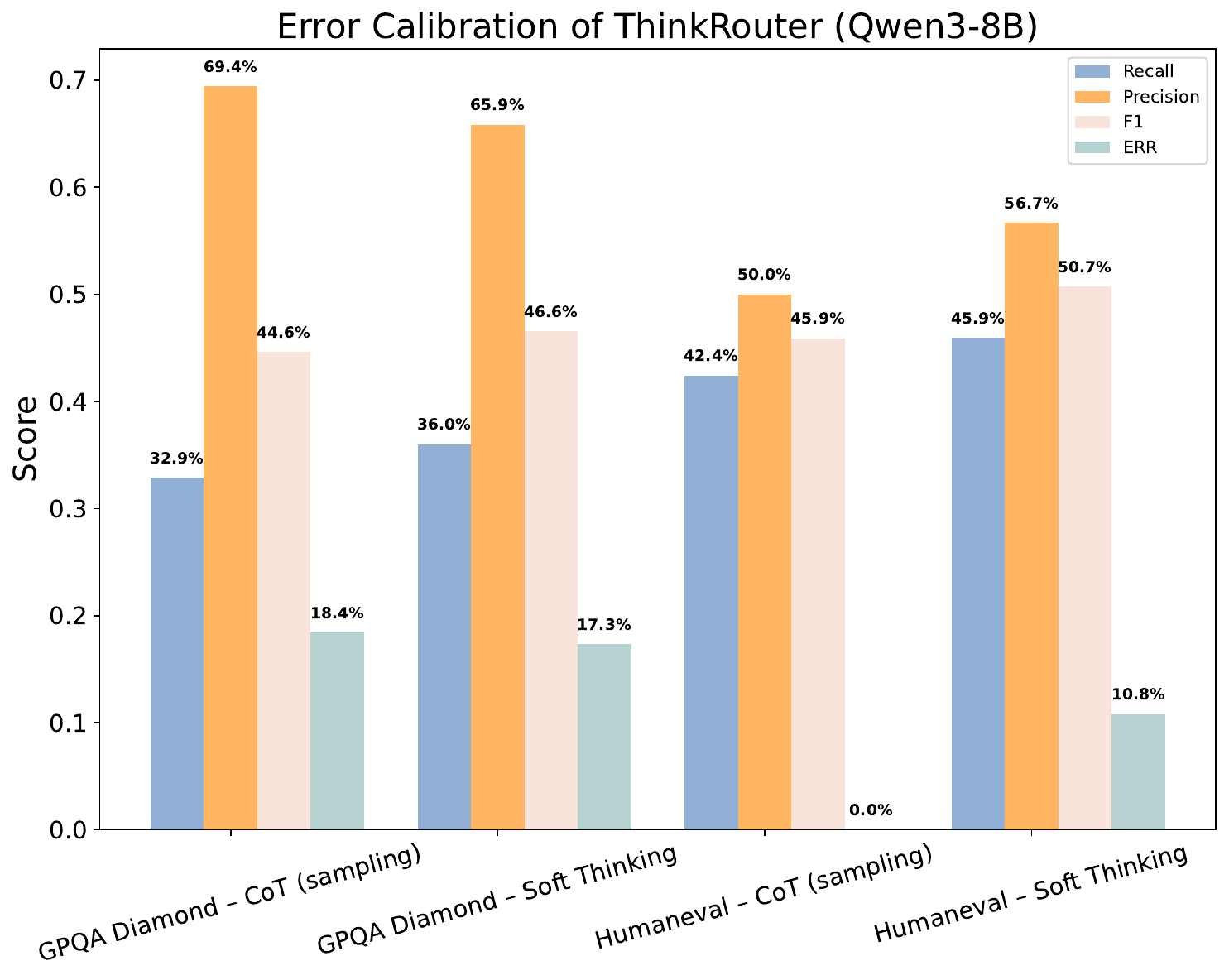}
        \vspace{-6mm}
        \caption{Qwen3-8B.}
        \label{fig:qwen3_matrix}
    \end{subfigure}

    \begin{subfigure}{0.9\linewidth}
        \centering
        \includegraphics[width=\linewidth]{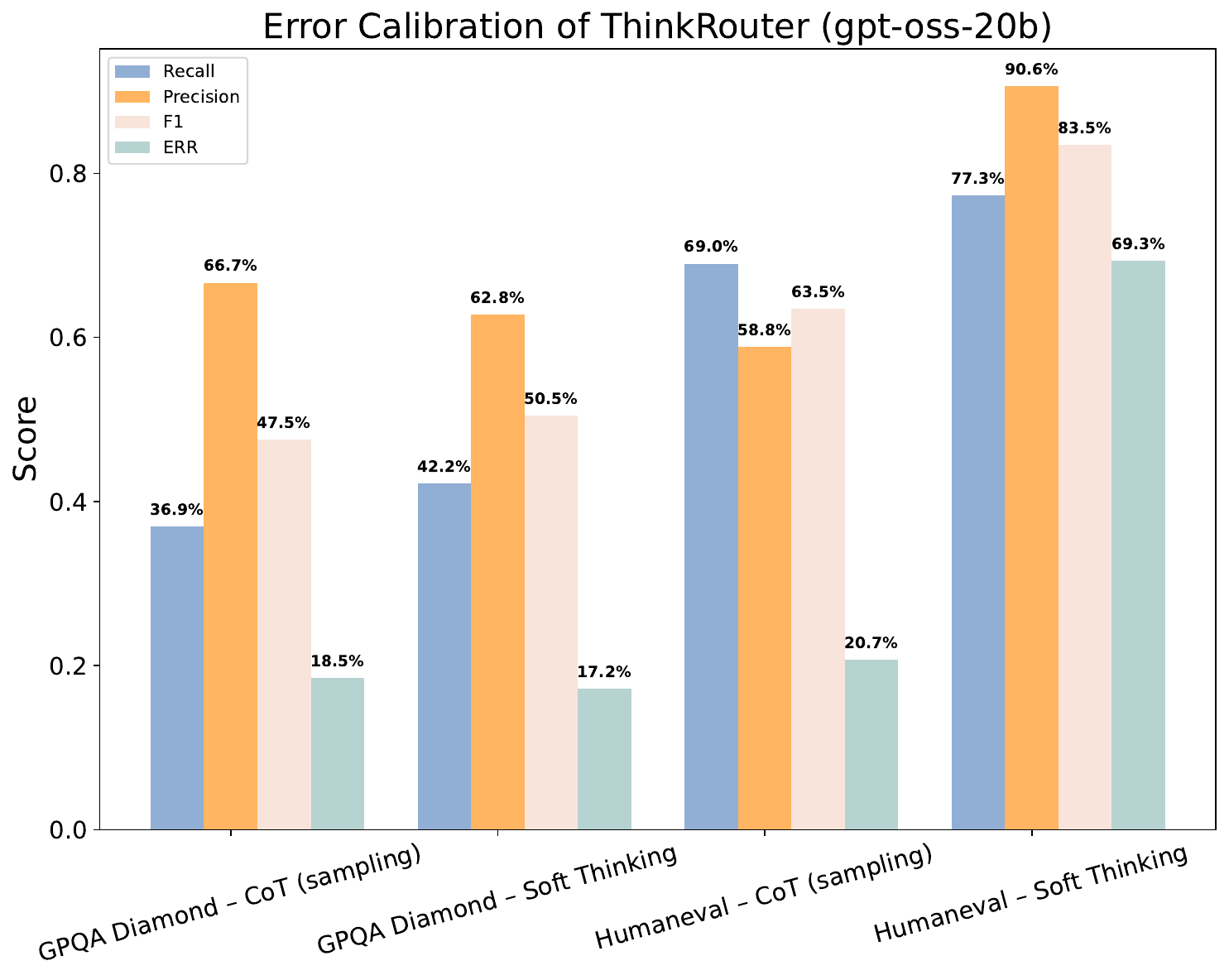}
        \vspace{-6mm}
        \caption{gpt-oss-20b.}
        \label{fig:gpt_matrix}
    \end{subfigure}

    \caption{Error calibration of \ours.}
    \label{fig:calibration}
\end{figure}

\begin{table*}[ht!]
\centering
\caption{Confusion-like matrix for analyzing calibration behavior in \ours.}
\scalebox{0.88}{
\begin{tabular}{l c c}
\toprule
 & \textbf{\ours\ - Correct} & \textbf{\ours\ - Incorrect} \\
\midrule
\textbf{Baseline - Correct} & TN (Baseline correct $\rightarrow$ \ours\ correct) & FP (Baseline correct $\rightarrow$ \ours\ incorrect) \\
\textbf{Baseline - Incorrect} & TP (Baseline incorrect $\rightarrow$ \ours\ correct) & FN (Baseline incorrect $\rightarrow$ \ours\ incorrect) \\
\bottomrule
\end{tabular}}
\label{tab:confusion}
\end{table*}
To better understand where the accuracy gains of \ours\ come from, we construct a confusion-style matrix as shown in Table \ref{tab:confusion} to analyze the error calibration capability between the baselines and \ours.
The matrix enumerates all possible combinations of calibration for two reasoning methods on the same samples.
We use the following three metrics to measure the error calibration capability of \ours.
Specifically, we use Recall as \textit{Fix Rate} ($=\frac{TP}{TP + FN}$) to measure error coverage, which is the proportion of baseline errors that are successfully corrected by \ours. 
\textit{Precision} ($=\frac{TP}{TP+FP}$) measures the reliability of calibrations without over-correction.
\textit{F1} ($=2 \cdot \frac{\text{Precision} \cdot \text{Recall}}{\text{Precision} + \text{Recall}}$) captures the balance between reliability and coverage. 
Meanwhile, we define the \textit{Error Reduction Rate (ERR $=\frac{\text{Errors (Baseline)} - \text{Errors (\ours)}}{\text{Errors (Baseline)}}$)} quantifies the net proportion of errors eliminated compared with the baseline. 
We report these metrics in Figure \ref{fig:calibration}.
For each test instance, we run inference three times with different random seeds (\{0, 7, 42\}) and determine the final answers by majority voting \citep{self-consistency}.
The results demonstrate that \ours\ consistently calibrate incorrect answers across different models and benchmarks.
According to \textit{Fix Rate}, up to 77.3\% errors of the baselines are successfully corrected by \ours.
The precision remains consistently high (up to 90.6\%), showing that \ours\ avoids aggressive over-correction.
More importantly, all \textit{ERR} is $\geq0$, illustrating that \ours\ can consistently perform corrective calibration without net error amplification.


\subsection{Why \ours\ Improves Reasoning Performance}
\label{sec:analysis}

\paragraph{LRM Confidence}
\begin{figure*}[htbp]
    \centering
    \begin{subfigure}[b]{0.46\textwidth}
        \centering
        \includegraphics[width=\textwidth]{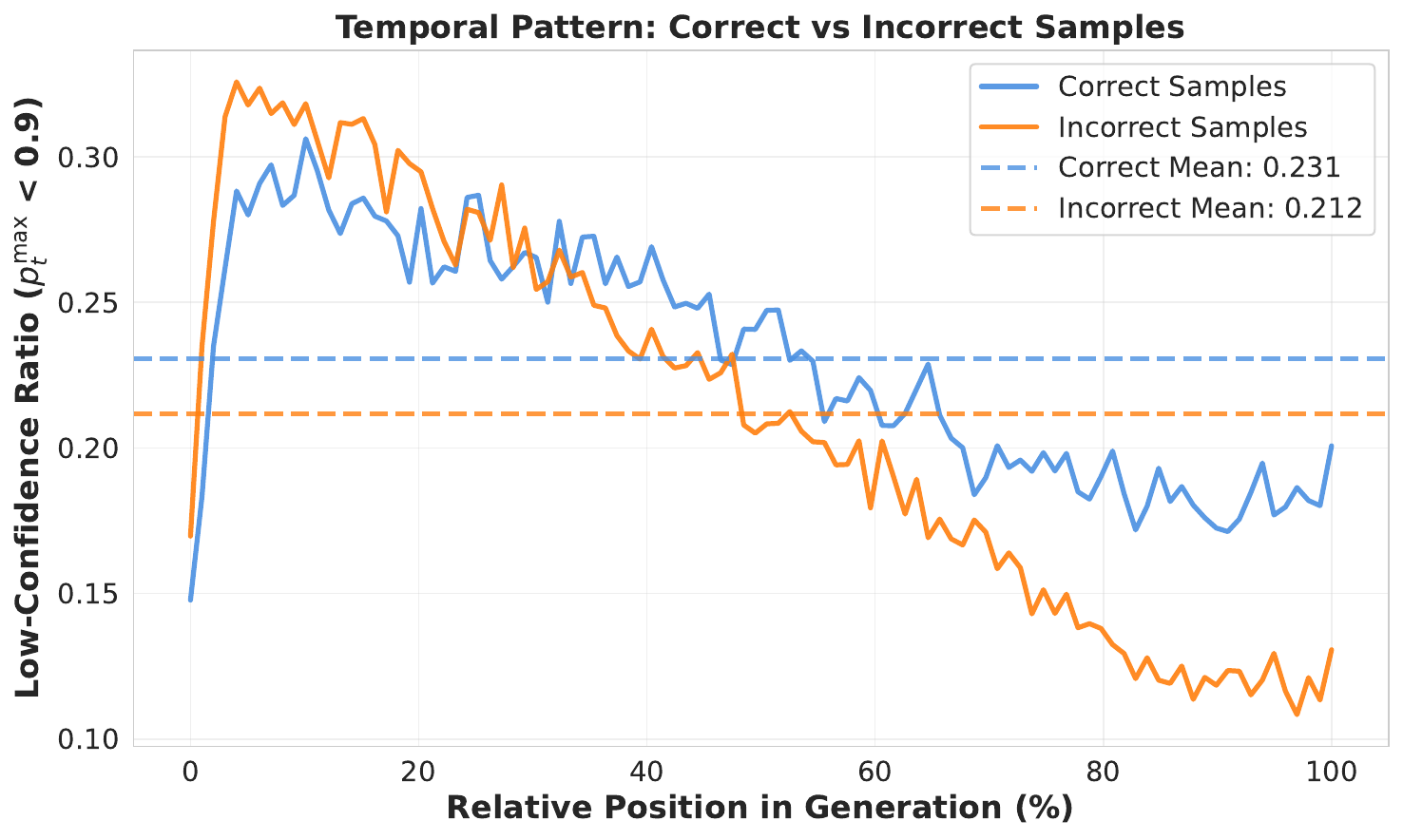}
        \caption{ \textit{Soft Thinking} (Latent-only Reasoning).}
    \end{subfigure}
    \hspace{0.03\textwidth}
    \begin{subfigure}[b]{0.46\textwidth}
        \centering
        \includegraphics[width=\textwidth]{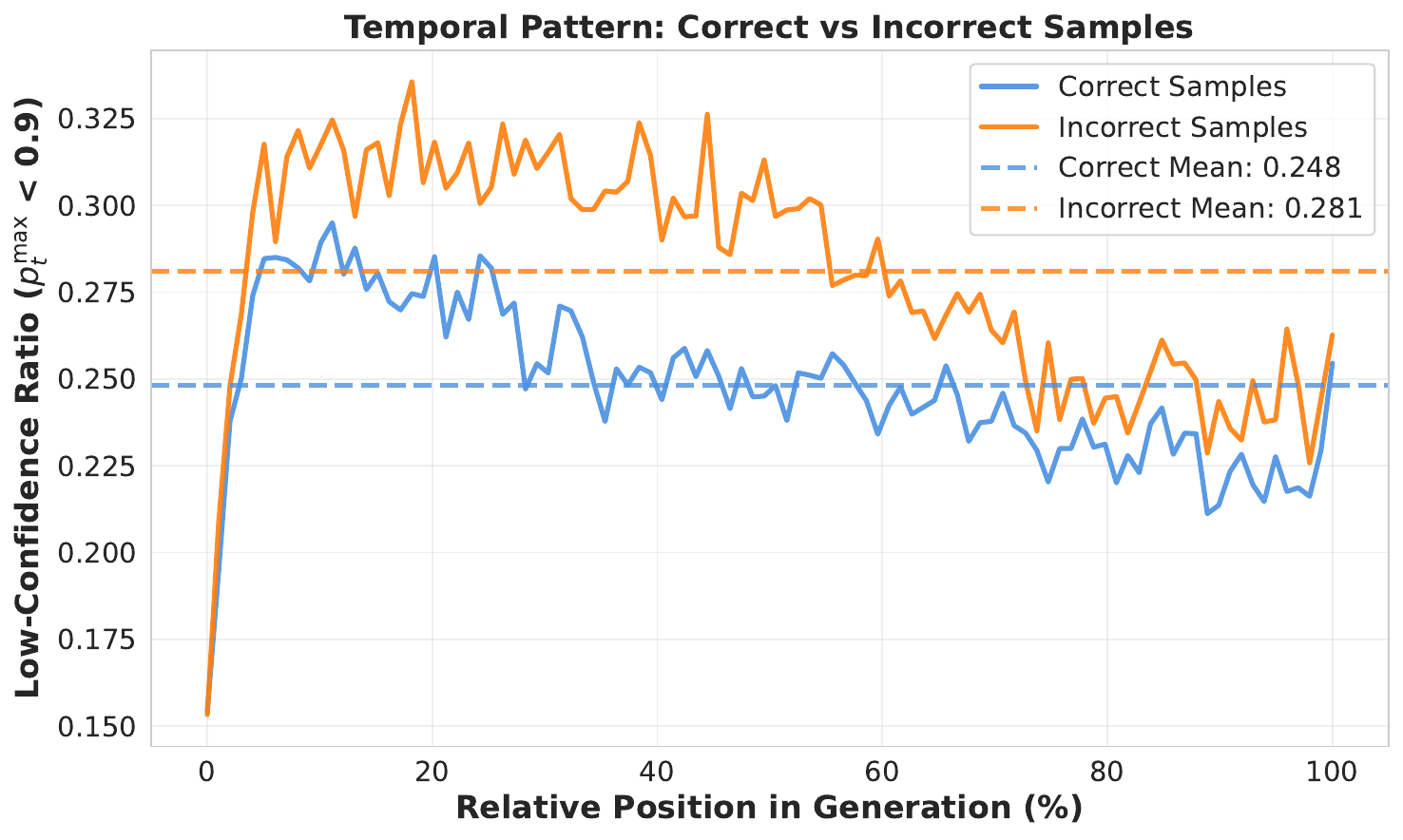}
        \caption{\ours.}
    \end{subfigure}
    \caption{Low-confidence time step ratio (\%) across generation steps on Qwen3-8B with GPQA Diamond.}
    \label{fig:temporal_pattern1}
\end{figure*}
To demystify how \ours\ works and whether it follows our motivation mentioned in \S \ref{sec:pre_motivation}, we analyze the confidence dynamics as thinking progresses. 
The details are described in Appendix \ref{app:low-confidence}.
Figure \ref{fig:temporal_pattern1}, \ref{fig:temporal_pattern2}, \ref{fig:temporal_pattern3}, and \ref{fig:temporal_pattern4}  show the ratios of low-confidence time steps for \textit{Soft Thinking} (latent-only reasoning) and our confidence-aware routing.
For \textit{Soft Thinking}, as thinking progresses, LRMs assign a higher $p_t^{\max}$ to incorrect solutions than correct ones, especially for the last period of thinking.
This trend is similar to the observations in \S \ref{sec:pre_motivation}.
Comparing \ours\ with \textit{Soft Thinking}, we find that \ours\ consistently increases the ratios of low-confidence steps across different models and datasets, especially for the samples LRMs give incorrect answers to, which suggests that \ours\ prevents LRMs from prematurely collapsing into thinking with relatively high confidence.
Furthermore, across all four figures, we observe that under \ours, the confidence trajectories of correct and incorrect solutions become increasingly closer as generation progresses, compared to \textit{Soft Thinking}.
This indicates that confidence-aware routing stabilizes inference-time confidence dynamics and mitigates the divergence of incorrect confidence trajectories from correct ones.
Overall, by controlling LRM confidence dynamics through these two ways, \ours\ improves reasoning accuracy.

\begin{figure*}
    \centering
    \begin{subfigure}[b]{0.33\textwidth}
        \centering
        \includegraphics[width=\textwidth]{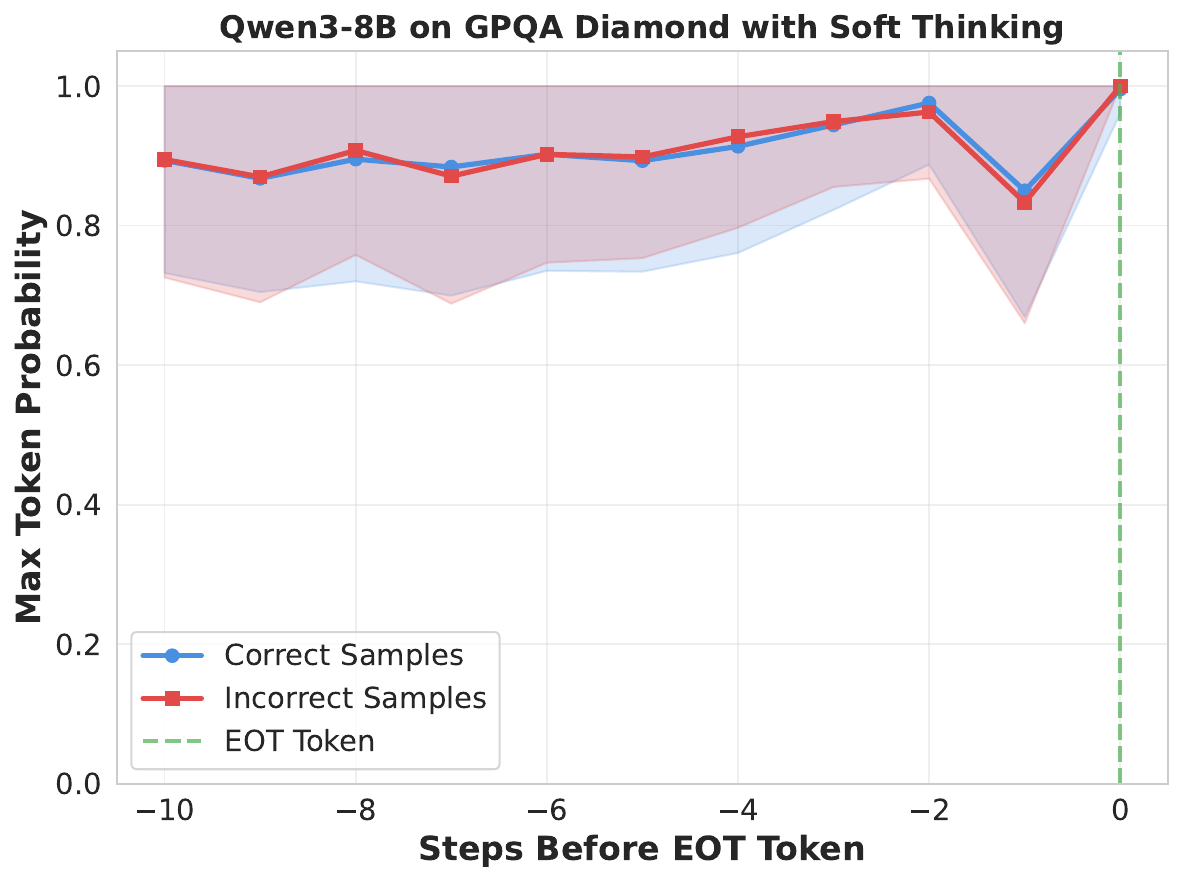}
    \end{subfigure}
    \begin{subfigure}[b]{0.33\textwidth}
        \centering
        \includegraphics[width=\textwidth]{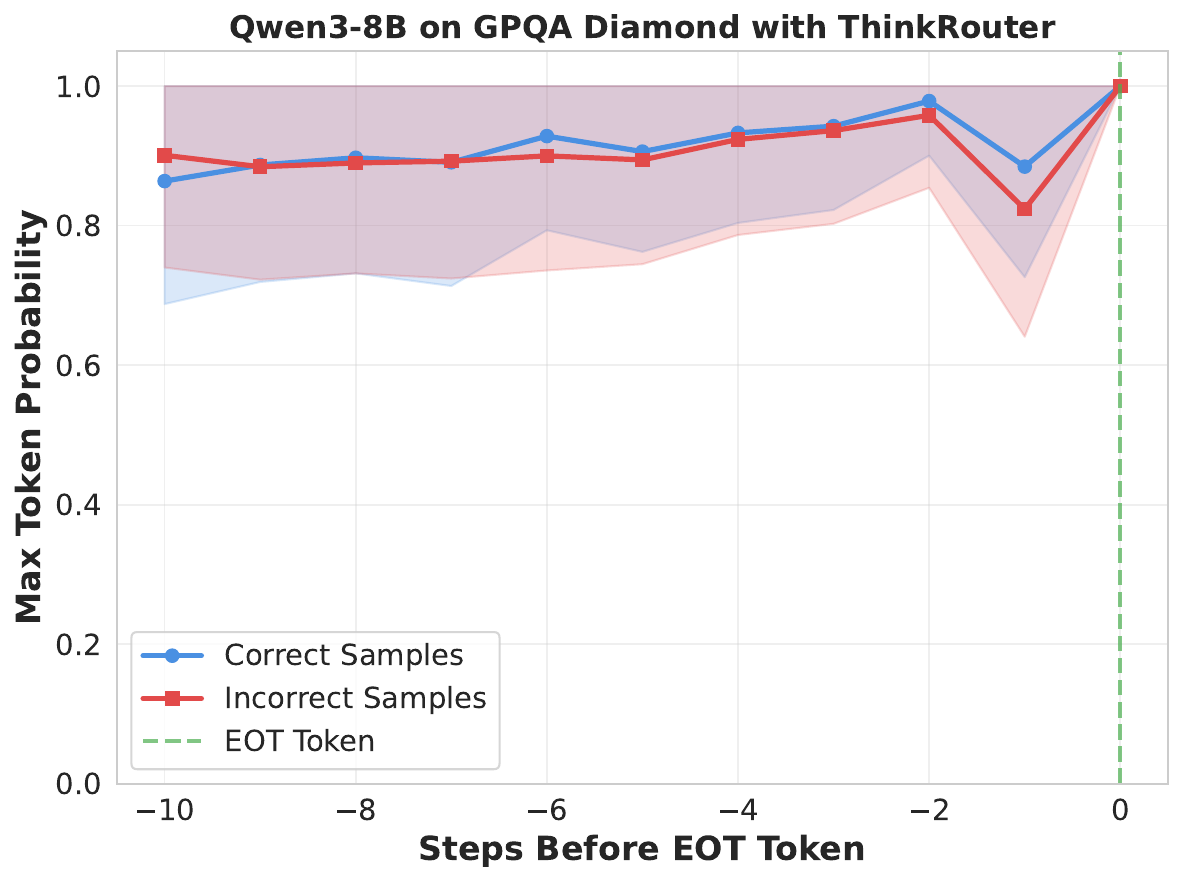}
    \end{subfigure}
    \begin{subfigure}[b]{0.33\textwidth}
        \centering
        \includegraphics[width=\textwidth]{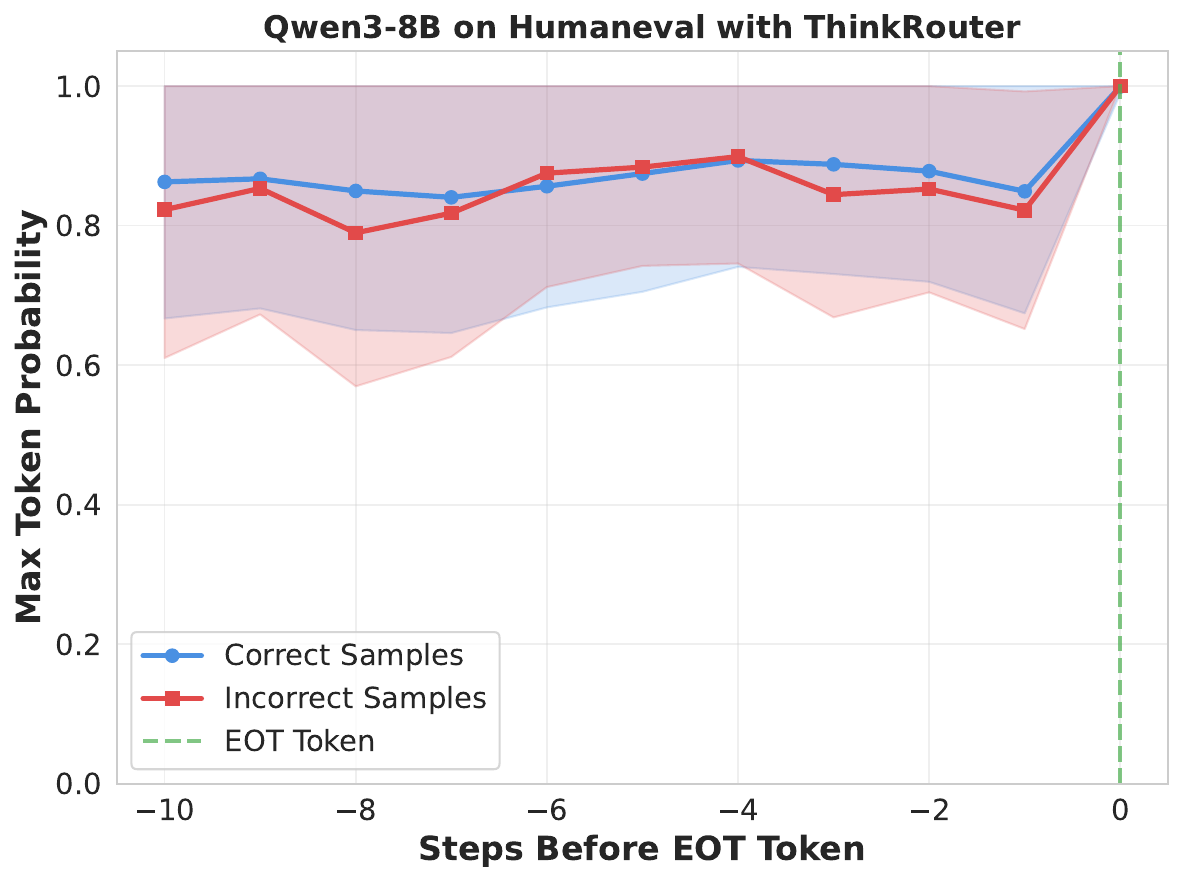}
    \end{subfigure}
    \caption{$p_t^{\max}$ of last 10 time steps before the end-of-thinking token for Qwen3-8B.}
    \label{fig:eot1}
\end{figure*}

\paragraph{Thinking Stop}
We explore why \ours\ can reduce the length of reasoning trajectories.
We firstly count the thinking stop modes as shown in Table \ref{tab:thinking_stop}, which suggests that \ours\ substantially reduces the proportion of thinking courses that terminate via \textit{Cold Stop}.
Since \textit{Cold Stop} is triggered under sustained overconfident token distributions (Figure \ref{fig:codestop_qwen3}), this shift indicates that \ours\ is effective at mitigating LRM confidence during thinking and promoting earlier, well-formed termination. 
For further understanding, we examine confidence evolution over the last ten steps preceding EOT token generation. 
Figure \ref{fig:eot1} and \ref{fig:eot2} show that EOT tokens are typically generated when the maximum next-token probability undergoes a noticeable drop or remains relatively low.
This suggests that \ours, by lowering and regularizing confidence across thinking, accelerates the conditions under which the EOT token becomes likely, thereby shortening the overall reasoning trajectory. 
Moreover, comparing samples with correct and incorrect answers reveals that incorrect samples generally exhibit longer reasoning trajectories than correct ones (Figure \ref{fig:generation_length_qwen3}, \ref{fig:generation_length_gptoss}).
By improving reasoning accuracy and reducing the fraction of such error-prone trajectories, \ours\ further decreases the generation length.

\subsection{Reasoning trajectory at Routing Times}
\label{sec:trajectory_analysis}
There are some examples of top-3 next-token probability distributions during thinking in Figures \ref{fig:prob_qwen3_gpqa}, \ref{fig:prob_qwen3_humaneval}, and \ref{fig:prob_gptoss_humaneval}. 
Inspecting a large number of examples, we observe that the time steps when \ours\ routes thinking to the discrete space mostly correspond to \textit{thinking tokens} with low confidence \citep{DBLP:journals/corr/abs-2506-02867}, which, for example, express $i)$ transitions 
(e.g., `then', `but', `alternatively'), $ii)$ execution (e.g., `let', `provide', `verify', `calculate'), and $iii)$ task-specific symbolics (e.g., mathematical LaTeX notations like `\$' and `\textbackslash', units like `kcal' and `mo'). 
According to \citet{DBLP:journals/corr/abs-2506-02867, DBLP:journals/corr/abs-2506-01939}, these \textit{thinking tokens} have mutual information peaks with the gold answers, which are critical to LRMs' reasoning performance. 
\ours\ routes such critical time to the discrete token space, suggesting that \ours\ does not route thinking arbitrarily, but selectively intervenes at time steps that are both semantically decisive and structurally important for reasoning, thereby improving the reliability of thinking.

\section{Conclusion}
In this work, we propose \ours, an inference-time mechanism for efficient reasoning that routes thinking between the discrete token space and the latent space based on LRM confidence.
Extensive experiments across diverse LRMs and benchmarks demonstrate that \ours\ can robustly improve reasoning accuracy and reduce generation length, and highlight the significance of \ours's confidence awareness.
Furthermore, we comprehensively analyze the underlying reasons for the effectiveness of \ours\ through LRM confidence dynamics.
We observe that \ours\ can globally lower model confidence.
\ours\ can calibrate errors arising from CoT and \textit{Soft Thinking} and accelerate the trigger of end-of-thinking token generation.
We hope our work can provide insights into efficient LRM reasoning and reasoning behaviors in the future.



\section*{Impact Statement}

\paragraph{LLM Usage} Two-family open-sourced LLMs, Qwen3 and gpt-oss-20b, are used for experiments (details in Section \ref{sec:experi_setup} and Appendix \ref{app:exper}).
GPT-4.1 serves as an LLM judge to help with the evaluation of STEM reasoning benchmarks.
GPT-5.2 is used to polish the writing.

This paper presents work whose goal is to advance the field of machine learning. There are many potential societal consequences of our work, none of which we feel must be specifically highlighted here.





\bibliography{example_paper}

@article{DBLP:journals/corr/abs-2506-01939,
  author       = {Shenzhi Wang and
                  Le Yu and
                  Chang Gao and
                  Chujie Zheng and
                  Shixuan Liu and
                  Rui Lu and
                  Kai Dang and
                  Xionghui Chen and
                  Jianxin Yang and
                  Zhenru Zhang and
                  Yuqiong Liu and
                  An Yang and
                  Andrew Zhao and
                  Yang Yue and
                  Shiji Song and
                  Bowen Yu and
                  Gao Huang and
                  Junyang Lin},
  title        = {Beyond the 80/20 Rule: High-Entropy Minority Tokens Drive Effective
                  Reinforcement Learning for {LLM} Reasoning},
  journal      = {CoRR},
  volume       = {abs/2506.01939},
  year         = {2025},
  url          = {https://doi.org/10.48550/arXiv.2506.01939},
  doi          = {10.48550/ARXIV.2506.01939},
  eprinttype    = {arXiv},
  eprint       = {2506.01939},
  bibsource    = {dblp computer science bibliography, https://dblp.org}
}

@article{DBLP:journals/corr/abs-2506-02867,
  author       = {Chen Qian and
                  Dongrui Liu and
                  Haochen Wen and
                  Zhen Bai and
                  Yong Liu and
                  Jing Shao},
  title        = {Demystifying Reasoning Dynamics with Mutual Information: Thinking
                  Tokens are Information Peaks in {LLM} Reasoning},
  journal      = {CoRR},
  volume       = {abs/2506.02867},
  year         = {2025},
  url          = {https://doi.org/10.48550/arXiv.2506.02867},
  doi          = {10.48550/ARXIV.2506.02867},
  eprinttype    = {arXiv},
  eprint       = {2506.02867},
  bibsource    = {dblp computer science bibliography, https://dblp.org}
}

@article{HRPO,
  author       = {Zhenrui Yue and
                  Bowen Jin and
                  Huimin Zeng and
                  Honglei Zhuang and
                  Zhen Qin and
                  Jinsung Yoon and
                  Lanyu Shang and
                  Jiawei Han and
                  Dong Wang},
  title        = {Hybrid Latent Reasoning via Reinforcement Learning},
  journal      = {CoRR},
  volume       = {abs/2505.18454},
  year         = {2025},
  url          = {https://doi.org/10.48550/arXiv.2505.18454},
  doi          = {10.48550/ARXIV.2505.18454},
  eprinttype    = {arXiv},
  eprint       = {2505.18454},
  bibsource    = {dblp computer science bibliography, https://dblp.org}
}

@article{thinkless,
  author       = {Gongfan Fang and
                  Xinyin Ma and
                  Xinchao Wang},
  title        = {Thinkless: {LLM} Learns When to Think},
  journal      = {CoRR},
  volume       = {abs/2505.13379},
  year         = {2025},
  url          = {https://doi.org/10.48550/arXiv.2505.13379},
  doi          = {10.48550/ARXIV.2505.13379},
  eprinttype    = {arXiv},
  eprint       = {2505.13379},
  bibsource    = {dblp computer science bibliography, https://dblp.org}
}

@article{adaptthink,
  author       = {Xu Wan and
                  Wei Wang and
                  Wenyue Xu and
                  Wotao Yin and
                  Jie Song and
                  Mingyang Sun},
  title        = {AdapThink: Adaptive Thinking Preferences for Reasoning Language Model},
  journal      = {CoRR},
  volume       = {abs/2506.18237},
  year         = {2025},
  url          = {https://doi.org/10.48550/arXiv.2506.18237},
  doi          = {10.48550/ARXIV.2506.18237},
  eprinttype    = {arXiv},
  eprint       = {2506.18237},
  bibsource    = {dblp computer science bibliography, https://dblp.org}
}

@article{mixreasoning,
  author       = {Haiquan Lu and
                  Gongfan Fang and
                  Xinyin Ma and
                 Qi Li and
                 Xinchao Wang},
  title        = {MixReasoning: Switching Modes to Think},
  journal      = {CoRR},
  volume       = {abs/2510.06052},
  year         = {2025},
  url          = {https://arxiv.org/abs/2510.06052},
  doi          = {10.48550/ARXIV.2510.06052},
  eprinttype    = {arXiv},
  eprint       = {2510.06052}
}

@article{LHRM,
  author       = {Lingjie Jiang and
                  Xun Wu and
                  Shaohan Huang and
                  Qingxiu Dong and
                  Zewen Chi and
                  Li Dong and
                  Xingxing Zhang and
                  Tengchao Lv and
                  Lei Cui and
                  Furu Wei},
  title        = {Think Only When You Need with Large Hybrid-Reasoning Models},
  journal      = {CoRR},
  volume       = {abs/2505.14631},
  year         = {2025},
  url          = {https://doi.org/10.48550/arXiv.2505.14631},
  doi          = {10.48550/ARXIV.2505.14631},
  eprinttype    = {arXiv},
  eprint       = {2505.14631},
  bibsource    = {dblp computer science bibliography, https://dblp.org}
}

@article{soft-thinking,
  author       = {Zhen Zhang and
                  Xuehai He and
                  Weixiang Yan and
                  Ao Shen and
                  Chenyang Zhao and
                  Shuohang Wang and
                  Yelong Shen and
                  Xin Eric Wang},
  title        = {Soft Thinking: Unlocking the Reasoning Potential of LLMs in Continuous
                  Concept Space},
  journal      = {CoRR},
  volume       = {abs/2505.15778},
  year         = {2025},
  url          = {https://doi.org/10.48550/arXiv.2505.15778},
  doi          = {10.48550/ARXIV.2505.15778},
  eprinttype    = {arXiv},
  eprint       = {2505.15778},
  bibsource    = {dblp computer science bibliography, https://dblp.org}
}

@article{coconut,
  author       = {Shibo Hao and
                  Sainbayar Sukhbaatar and
                  DiJia Su and
                  Xian Li and
                  Zhiting Hu and
                  Jason Weston and
                  Yuandong Tian},
  title        = {Training Large Language Models to Reason in a Continuous Latent Space},
  journal      = {CoRR},
  volume       = {abs/2412.06769},
  year         = {2024},
  url          = {https://doi.org/10.48550/arXiv.2412.06769},
  doi          = {10.48550/ARXIV.2412.06769},
  eprinttype    = {arXiv},
  eprint       = {2412.06769},
  bibsource    = {dblp computer science bibliography, https://dblp.org}
}

@misc{aime,
  title        = {American Invitational Mathematics Examination},
  author       = {{AIME}},
  howpublished = {\url{https://artofproblemsolving.com/wiki/index.php/American_Invitational_Mathematics_Examination}},

}

@article{gpqa,
  author       = {David Rein and
                  Betty Li Hou and
                  Asa Cooper Stickland and
                  Jackson Petty and
                  Richard Yuanzhe Pang and
                  Julien Dirani and
                  Julian Michael and
                  Samuel R. Bowman},
  title        = {{GPQA:} {A} Graduate-Level Google-Proof Q{\&}A Benchmark},
  journal      = {CoRR},
  volume       = {abs/2311.12022},
  year         = {2023},
  url          = {https://doi.org/10.48550/arXiv.2311.12022},
  doi          = {10.48550/ARXIV.2311.12022},
  eprinttype    = {arXiv},
  eprint       = {2311.12022},
  bibsource    = {dblp computer science bibliography, https://dblp.org}
}

@article{humaneval,
  author       = {Mark Chen and
                  Jerry Tworek and
                  Heewoo Jun and
                  Qiming Yuan and
                  Henrique Pond{\'{e}} de Oliveira Pinto and
                  Jared Kaplan and
                  Harri Edwards and
                  Yuri Burda and
                  Nicholas Joseph and
                  Greg Brockman and
                  Alex Ray and
                  Raul Puri and
                  Gretchen Krueger and
                  Michael Petrov and
                  Heidy Khlaaf and
                  Girish Sastry and
                  Pamela Mishkin and
                  Brooke Chan and
                  Scott Gray and
                  Nick Ryder and
                  Mikhail Pavlov and
                  Alethea Power and
                  Lukasz Kaiser and
                  Mohammad Bavarian and
                  Clemens Winter and
                  Philippe Tillet and
                  Felipe Petroski Such and
                  Dave Cummings and
                  Matthias Plappert and
                  Fotios Chantzis and
                  Elizabeth Barnes and
                  Ariel Herbert{-}Voss and
                  William Hebgen Guss and
                  Alex Nichol and
                  Alex Paino and
                  Nikolas Tezak and
                  Jie Tang and
                  Igor Babuschkin and
                  Suchir Balaji and
                  Shantanu Jain and
                  William Saunders and
                  Christopher Hesse and
                  Andrew N. Carr and
                  Jan Leike and
                  Joshua Achiam and
                  Vedant Misra and
                  Evan Morikawa and
                  Alec Radford and
                  Matthew Knight and
                  Miles Brundage and
                  Mira Murati and
                  Katie Mayer and
                  Peter Welinder and
                  Bob McGrew and
                  Dario Amodei and
                  Sam McCandlish and
                  Ilya Sutskever and
                  Wojciech Zaremba},
  title        = {Evaluating Large Language Models Trained on Code},
  journal      = {CoRR},
  volume       = {abs/2107.03374},
  year         = {2021},
  url          = {https://arxiv.org/abs/2107.03374},
  eprinttype    = {arXiv},
  eprint       = {2107.03374},
  bibsource    = {dblp computer science bibliography, https://dblp.org}
}

@article{mbpp,
  author       = {Jacob Austin and
                  Augustus Odena and
                  Maxwell I. Nye and
                  Maarten Bosma and
                  Henryk Michalewski and
                  David Dohan and
                  Ellen Jiang and
                  Carrie J. Cai and
                  Michael Terry and
                  Quoc V. Le and
                  Charles Sutton},
  title        = {Program Synthesis with Large Language Models},
  journal      = {CoRR},
  volume       = {abs/2108.07732},
  year         = {2021},
  url          = {https://arxiv.org/abs/2108.07732},
  eprinttype    = {arXiv},
  eprint       = {2108.07732},
  bibsource    = {dblp computer science bibliography, https://dblp.org}
}

@article{qwen3,
  author       = {An Yang and
                  Anfeng Li and
                  Baosong Yang and
                  Beichen Zhang and
                  Binyuan Hui and
                  Bo Zheng and
                  Bowen Yu and
                  Chang Gao and
                  Chengen Huang and
                  Chenxu Lv and
                  Chujie Zheng and
                  Dayiheng Liu and
                  Fan Zhou and
                  Fei Huang and
                  Feng Hu and
                  Hao Ge and
                  Haoran Wei and
                  Huan Lin and
                  Jialong Tang and
                  Jian Yang and
                  Jianhong Tu and
                  Jianwei Zhang and
                  Jian Yang and
                  Jiaxi Yang and
                  Jingren Zhou and
                  Junyang Lin and
                  Kai Dang and
                  Keqin Bao and
                  Kexin Yang and
                  Le Yu and
                  Lianghao Deng and
                  Mei Li and
                  Mingfeng Xue and
                  Mingze Li and
                  Pei Zhang and
                  Peng Wang and
                  Qin Zhu and
                  Rui Men and
                  Ruize Gao and
                  Shixuan Liu and
                  Shuang Luo and
                  Tianhao Li and
                  Tianyi Tang and
                  Wenbiao Yin and
                  Xingzhang Ren and
                  Xinyu Wang and
                  Xinyu Zhang and
                  Xuancheng Ren and
                  Yang Fan and
                  Yang Su and
                  Yichang Zhang and
                  Yinger Zhang and
                  Yu Wan and
                  Yuqiong Liu and
                  Zekun Wang and
                  Zeyu Cui and
                  Zhenru Zhang and
                  Zhipeng Zhou and
                  Zihan Qiu},
  title        = {Qwen3 Technical Report},
  journal      = {CoRR},
  volume       = {abs/2505.09388},
  year         = {2025},
  url          = {https://doi.org/10.48550/arXiv.2505.09388},
  doi          = {10.48550/ARXIV.2505.09388},
  eprinttype    = {arXiv},
  eprint       = {2505.09388},
  bibsource    = {dblp computer science bibliography, https://dblp.org}
}

@article{gpt-oss,
  author       = {Sandhini Agarwal and
                  Lama Ahmad and
                  Jason Ai and
                  Sam Altman and
                  Andy Applebaum and
                  Edwin Arbus and
                  Rahul K. Arora and
                  Yu Bai and
                  Bowen Baker and
                  Haiming Bao and
                  Boaz Barak and
                  Ally Bennett and
                  Tyler Bertao and
                  Nivedita Brett and
                  Eugene Brevdo and
                  Greg Brockman and
                  S{\'{e}}bastien Bubeck and
                  Che Chang and
                  Kai Chen and
                  Mark Chen and
                  Enoch Cheung and
                  Aidan Clark and
                  Dan Cook and
                  Marat Dukhan and
                  Casey Dvorak and
                  Kevin Fives and
                  Vlad Fomenko and
                  Timur Garipov and
                  Kristian Georgiev and
                  Mia Glaese and
                  Tarun Gogineni and
                  Adam P. Goucher and
                  Lukas Gross and
                  Katia Gil Guzman and
                  John Hallman and
                  Jackie Hehir and
                  Johannes Heidecke and
                  Alec Helyar and
                  Haitang Hu and
                  Romain Huet and
                  Jacob Huh and
                  Saachi Jain and
                  Zach Johnson and
                  Chris Koch and
                  Irina Kofman and
                  Dominik Kundel and
                  Jason Kwon and
                  Volodymyr Kyrylov and
                  Elaine Ya Le and
                  Guillaume Leclerc and
                  James Park Lennon and
                  Scott Lessans and
                  Mario Lezcano Casado and
                  Yuanzhi Li and
                  Zhuohan Li and
                  Ji Lin and
                  Jordan Liss and
                  Lily Liu and
                  Jiancheng Liu and
                  Kevin Lu and
                  Chris Lu and
                  Zoran Martinovic and
                  Lindsay McCallum and
                  Josh McGrath and
                  Scott McKinney and
                  Aidan McLaughlin and
                  Song Mei and
                  Steve Mostovoy and
                  Tong Mu and
                  Gideon Myles and
                  Alexander Neitz and
                  Alex Nichol and
                  Jakub Pachocki and
                  Alex Paino and
                  Dana Palmie and
                  Ashley Pantuliano and
                  Giambattista Parascandolo and
                  Jongsoo Park and
                  Leher Pathak and
                  Carolina Paz and
                  Ludovic Peran and
                  Dmitry Pimenov and
                  Michelle Pokrass and
                  Elizabeth Proehl and
                  Huida Qiu and
                  Gaby Raila and
                  Filippo Raso and
                  Hongyu Ren and
                  Kimmy Richardson and
                  David Robinson and
                  Bob Rotsted and
                  Hadi Salman and
                  Suvansh Sanjeev and
                  Max Schwarzer and
                  D. Sculley and
                  Harshit Sikchi and
                  Kendal Simon and
                  Karan Singhal and
                  Yang Song and
                  Dane Stuckey and
                  Zhiqing Sun and
                  Philippe Tillet and
                  Sam Toizer and
                  Foivos Tsimpourlas and
                  Nikhil Vyas and
                  Eric Wallace and
                  Xin Wang and
                  Miles Wang and
                  Olivia Watkins and
                  Kevin Weil and
                  Amy Wendling and
                  Kevin Whinnery and
                  Cedric Whitney and
                  Hannah Wong and
                  Lin Yang and
                  Yu Yang and
                  Michihiro Yasunaga and
                  Kristen Ying and
                  Wojciech Zaremba and
                  Wenting Zhan and
                  Cyril Zhang and
                  Brian Zhang and
                  Eddie Zhang and
                  Shengjia Zhao},
  title        = {gpt-oss-120b {\&} gpt-oss-20b Model Card},
  journal      = {CoRR},
  volume       = {abs/2508.10925},
  year         = {2025},
  url          = {https://doi.org/10.48550/arXiv.2508.10925},
  doi          = {10.48550/ARXIV.2508.10925},
  eprinttype    = {arXiv},
  eprint       = {2508.10925},
  bibsource    = {dblp computer science bibliography, https://dblp.org}
}

@inproceedings{softcot,
    title = "{S}oft{C}o{T}: Soft Chain-of-Thought for Efficient Reasoning with {LLM}s",
    author = "Xu, Yige  and
      Guo, Xu  and
      Zeng, Zhiwei  and
      Miao, Chunyan",
    editor = "Che, Wanxiang  and
      Nabende, Joyce  and
      Shutova, Ekaterina  and
      Pilehvar, Mohammad Taher",
    booktitle = "Proceedings of the 63rd Annual Meeting of the Association for Computational Linguistics (Volume 1: Long Papers)",
    month = jul,
    year = "2025",
    address = "Vienna, Austria",
    publisher = "Association for Computational Linguistics",
    url = "https://aclanthology.org/2025.acl-long.1137/",
    doi = "10.18653/v1/2025.acl-long.1137",
    pages = "23336--23351",
    ISBN = "979-8-89176-251-0"
}

@article{codi,
  author       = {Zhenyi Shen and
                  Hanqi Yan and
                  Linhai Zhang and
                  Zhanghao Hu and
                  Yali Du and
                  Yulan He},
  title        = {{CODI:} Compressing Chain-of-Thought into Continuous Space via Self-Distillation},
  journal      = {CoRR},
  volume       = {abs/2502.21074},
  year         = {2025},
  url          = {https://doi.org/10.48550/arXiv.2502.21074},
  doi          = {10.48550/ARXIV.2502.21074},
  eprinttype    = {arXiv},
  eprint       = {2502.21074},
  bibsource    = {dblp computer science bibliography, https://dblp.org}
}

@article{ccot,
  author       = {Jeffrey Cheng and
                  Benjamin Van Durme},
  title        = {Compressed Chain of Thought: Efficient Reasoning Through Dense Representations},
  journal      = {CoRR},
  volume       = {abs/2412.13171},
  year         = {2024},
  url          = {https://doi.org/10.48550/arXiv.2412.13171},
  doi          = {10.48550/ARXIV.2412.13171},
  eprinttype    = {arXiv},
  eprint       = {2412.13171},
  bibsource    = {dblp computer science bibliography, https://dblp.org}
}

@article{CoLaR,
  author       = {Wenhui Tan and
                  Jiaze Li and
                  Jianzhong Ju and
                  Zhenbo Luo and
                  Jian Luan and
                  Ruihua Song},
  title        = {Think Silently, Think Fast: Dynamic Latent Compression of {LLM} Reasoning
                  Chains},
  journal      = {CoRR},
  volume       = {abs/2505.16552},
  year         = {2025},
  url          = {https://doi.org/10.48550/arXiv.2505.16552},
  doi          = {10.48550/ARXIV.2505.16552},
  eprinttype    = {arXiv},
  eprint       = {2505.16552},
  bibsource    = {dblp computer science bibliography, https://dblp.org}
}

@article{DBLP:journals/corr/abs-2311-01460,
  author       = {Yuntian Deng and
                  Kiran Prasad and
                  Roland Fernandez and
                  Paul Smolensky and
                  Vishrav Chaudhary and
                  Stuart M. Shieber},
  title        = {Implicit Chain of Thought Reasoning via Knowledge Distillation},
  journal      = {CoRR},
  volume       = {abs/2311.01460},
  year         = {2023},
  url          = {https://doi.org/10.48550/arXiv.2311.01460},
  doi          = {10.48550/ARXIV.2311.01460},
  eprinttype    = {arXiv},
  eprint       = {2311.01460},
  bibsource    = {dblp computer science bibliography, https://dblp.org}
}

@article{SIM-CoT,
  author       = {Xilin Wei and
                  Xiaoran Liu and
                  Yuhang Zang and
                  Xiaoyi Dong and
                  Yuhang Cao and
                  Jiaqi Wang and
                  Xipeng Qiu and
                  Dahua Lin},
  title        = {SIM-CoT: Supervised Implicit Chain-of-Thought},
  journal      = {CoRR},
  volume       = {abs/2509.20317},
  year         = {2025},
  url          = {https://doi.org/10.48550/arXiv.2509.20317},
  doi          = {10.48550/ARXIV.2509.20317},
  eprinttype    = {arXiv},
  eprint       = {2509.20317},
  bibsource    = {dblp computer science bibliography, https://dblp.org}
}

@article{cot2,
  author       = {Halil Alperen Gozeten and
                  Muhammed Emrullah Ildiz and
                  Xuechen Zhang and
                  Hrayr Harutyunyan and
                  Ankit Singh Rawat and
                  Samet Oymak},
  title        = {Continuous Chain of Thought Enables Parallel Exploration and Reasoning},
  journal      = {CoRR},
  volume       = {abs/2505.23648},
  year         = {2025},
  url          = {https://doi.org/10.48550/arXiv.2505.23648},
  doi          = {10.48550/ARXIV.2505.23648},
  eprinttype    = {arXiv},
  eprint       = {2505.23648},
  bibsource    = {dblp computer science bibliography, https://dblp.org}
}

@article{soft_token_hard_truth,
  author       = {Natasha Butt and
                  Ariel Kwiatkowski and
                  Ismail Labiad and
                  Julia Kempe and
                  Yann Ollivier},
  title        = {Soft Tokens, Hard Truths},
  journal      = {CoRR},
  volume       = {abs/2509.19170},
  year         = {2025},
  url          = {https://doi.org/10.48550/arXiv.2509.19170},
  doi          = {10.48550/ARXIV.2509.19170},
  eprinttype    = {arXiv},
  eprint       = {2509.19170},
  bibsource    = {dblp computer science bibliography, https://dblp.org}
}

@inproceedings{sglang,
  author       = {Lianmin Zheng and
                  Liangsheng Yin and
                  Zhiqiang Xie and
                  Chuyue Sun and
                  Jeff Huang and
                  Cody Hao Yu and
                  Shiyi Cao and
                  Christos Kozyrakis and
                  Ion Stoica and
                  Joseph E. Gonzalez and
                  Clark W. Barrett and
                  Ying Sheng},
  editor       = {Amir Globersons and
                  Lester Mackey and
                  Danielle Belgrave and
                  Angela Fan and
                  Ulrich Paquet and
                  Jakub M. Tomczak and
                  Cheng Zhang},
  title        = {SGLang: Efficient Execution of Structured Language Model Programs},
  booktitle    = {Advances in Neural Information Processing Systems 38: Annual Conference
                  on Neural Information Processing Systems 2024, NeurIPS 2024, Vancouver,
                  BC, Canada, December 10 - 15, 2024},
  year         = {2024},
  url          = {http://papers.nips.cc/paper\_files/paper/2024/hash/724be4472168f31ba1c9ac630f15dec8-Abstract-Conference.html},
  bibsource    = {dblp computer science bibliography, https://dblp.org}
}

@inproceedings{self-consistency,
  author       = {Xuezhi Wang and
                  Jason Wei and
                  Dale Schuurmans and
                  Quoc V. Le and
                  Ed H. Chi and
                  Sharan Narang and
                  Aakanksha Chowdhery and
                  Denny Zhou},
  title        = {Self-Consistency Improves Chain of Thought Reasoning in Language Models},
  booktitle    = {The Eleventh International Conference on Learning Representations,
                  {ICLR} 2023, Kigali, Rwanda, May 1-5, 2023},
  publisher    = {OpenReview.net},
  year         = {2023},
  url          = {https://openreview.net/forum?id=1PL1NIMMrw},
  bibsource    = {dblp computer science bibliography, https://dblp.org}
}

@article{survey1,
  author       = {Xinghao Chen and
                  Anhao Zhao and
                  Heming Xia and
                  Xuan Lu and
                  Hanlin Wang and
                  Yanjun Chen and
                  Wei Zhang and
                  Jian Wang and
                  Wenjie Li and
                  Xiaoyu Shen},
  title        = {Reasoning Beyond Language: {A} Comprehensive Survey on Latent Chain-of-Thought
                  Reasoning},
  journal      = {CoRR},
  volume       = {abs/2505.16782},
  year         = {2025},
  url          = {https://doi.org/10.48550/arXiv.2505.16782},
  doi          = {10.48550/ARXIV.2505.16782},
  eprinttype    = {arXiv},
  eprint       = {2505.16782},
  bibsource    = {dblp computer science bibliography, https://dblp.org}
}

@article{survey2,
  author       = {Ruijie Zhu and
                  Tianhao Peng and
                  Tianhao Cheng and
                  Xingwei Qu and
                  Jinfa Huang and
                  Dawei Zhu and
                  Hao Wang and
                  Kaiwen Xue and
                  Xuanliang Zhang and
                  Yong Shan and
                  Tianle Cai and
                  Taylor Kergan and
                  Assel Kembay and
                  Andrew Smith and
                  Chenghua Lin and
                  Binh Nguyen and
                  Yuqi Pan and
                  Yuhong Chou and
                  Zefan Cai and
                  Zhenhe Wu and
                  Yongchi Zhao and
                  Tianyu Liu and
                  Jian Yang and
                  Wangchunshu Zhou and
                  Chujie Zheng and
                  Chongxuan Li and
                  Yuyin Zhou and
                  Zhoujun Li and
                  Zhaoxiang Zhang and
                  Jiaheng Liu and
                  Ge Zhang and
                  Wenhao Huang and
                  Jason Eshraghian},
  title        = {A Survey on Latent Reasoning},
  journal      = {CoRR},
  volume       = {abs/2507.06203},
  year         = {2025},
  url          = {https://doi.org/10.48550/arXiv.2507.06203},
  doi          = {10.48550/ARXIV.2507.06203},
  eprinttype    = {arXiv},
  eprint       = {2507.06203},
  bibsource    = {dblp computer science bibliography, https://dblp.org}
}

@inproceedings{zhang-etal-2025-adaptthink,
    title = "{A}dapt{T}hink: Reasoning Models Can Learn When to Think",
    author = "Zhang, Jiajie  and
      Lin, Nianyi  and
      Hou, Lei  and
      Feng, Ling  and
      Li, Juanzi",
    editor = "Christodoulopoulos, Christos  and
      Chakraborty, Tanmoy  and
      Rose, Carolyn  and
      Peng, Violet",
    booktitle = "Proceedings of the 2025 Conference on Empirical Methods in Natural Language Processing",
    month = nov,
    year = "2025",
    address = "Suzhou, China",
    publisher = "Association for Computational Linguistics",
    url = "https://aclanthology.org/2025.emnlp-main.184/",
    doi = "10.18653/v1/2025.emnlp-main.184",
    pages = "3716--3730",
    ISBN = "979-8-89176-332-6"
}

@inproceedings{mmlu_pro,
  author       = {Yubo Wang and
                  Xueguang Ma and
                  Ge Zhang and
                  Yuansheng Ni and
                  Abhranil Chandra and
                  Shiguang Guo and
                  Weiming Ren and
                  Aaran Arulraj and
                  Xuan He and
                  Ziyan Jiang and
                  Tianle Li and
                  Max Ku and
                  Kai Wang and
                  Alex Zhuang and
                  Rongqi Fan and
                  Xiang Yue and
                  Wenhu Chen},
  editor       = {Amir Globersons and
                  Lester Mackey and
                  Danielle Belgrave and
                  Angela Fan and
                  Ulrich Paquet and
                  Jakub M. Tomczak and
                  Cheng Zhang},
  title        = {MMLU-Pro: {A} More Robust and Challenging Multi-Task Language Understanding
                  Benchmark},
  booktitle    = {Advances in Neural Information Processing Systems 38: Annual Conference
                  on Neural Information Processing Systems 2024, NeurIPS 2024, Vancouver,
                  BC, Canada, December 10 - 15, 2024},
  year         = {2024},
  url          = {http://papers.nips.cc/paper\_files/paper/2024/hash/ad236edc564f3e3156e1b2feafb99a24-Abstract-Datasets\_and\_Benchmarks\_Track.html},
  bibsource    = {dblp computer science bibliography, https://dblp.org}
}

@inproceedings{huang-chang-2023-towards,
    title = "Towards Reasoning in Large Language Models: A Survey",
    author = "Huang, Jie  and
      Chang, Kevin Chen-Chuan",
    editor = "Rogers, Anna  and
      Boyd-Graber, Jordan  and
      Okazaki, Naoaki",
    booktitle = "Findings of the Association for Computational Linguistics: ACL 2023",
    month = jul,
    year = "2023",
    address = "Toronto, Canada",
    publisher = "Association for Computational Linguistics",
    url = "https://aclanthology.org/2023.findings-acl.67/",
    doi = "10.18653/v1/2023.findings-acl.67",
    pages = "1049--1065"
}

@inproceedings{chu-etal-2024-navigate,
    title = "Navigate through Enigmatic Labyrinth A Survey of Chain of Thought Reasoning: Advances, Frontiers and Future",
    author = "Chu, Zheng  and
      Chen, Jingchang  and
      Chen, Qianglong  and
      Yu, Weijiang  and
      He, Tao  and
      Wang, Haotian  and
      Peng, Weihua  and
      Liu, Ming  and
      Qin, Bing  and
      Liu, Ting",
    editor = "Ku, Lun-Wei  and
      Martins, Andre  and
      Srikumar, Vivek",
    booktitle = "Proceedings of the 62nd Annual Meeting of the Association for Computational Linguistics (Volume 1: Long Papers)",
    month = aug,
    year = "2024",
    address = "Bangkok, Thailand",
    publisher = "Association for Computational Linguistics",
    url = "https://aclanthology.org/2024.acl-long.65/",
    doi = "10.18653/v1/2024.acl-long.65",
    pages = "1173--1203"
}

@inproceedings{cot,
  author       = {Jason Wei and
                  Xuezhi Wang and
                  Dale Schuurmans and
                  Maarten Bosma and
                  Brian Ichter and
                  Fei Xia and
                  Ed H. Chi and
                  Quoc V. Le and
                  Denny Zhou},
  editor       = {Sanmi Koyejo and
                  S. Mohamed and
                  A. Agarwal and
                  Danielle Belgrave and
                  K. Cho and
                  A. Oh},
  title        = {Chain-of-Thought Prompting Elicits Reasoning in Large Language Models},
  booktitle    = {Advances in Neural Information Processing Systems 35: Annual Conference
                  on Neural Information Processing Systems 2022, NeurIPS 2022, New Orleans,
                  LA, USA, November 28 - December 9, 2022},
  year         = {2022},
  url          = {http://papers.nips.cc/paper\_files/paper/2022/hash/9d5609613524ecf4f15af0f7b31abca4-Abstract-Conference.html},
  bibsource    = {dblp computer science bibliography, https://dblp.org}
}

@inproceedings{cot_mystery,
  author       = {Guhao Feng and
                  Bohang Zhang and
                  Yuntian Gu and
                  Haotian Ye and
                  Di He and
                  Liwei Wang},
  editor       = {Alice Oh and
                  Tristan Naumann and
                  Amir Globerson and
                  Kate Saenko and
                  Moritz Hardt and
                  Sergey Levine},
  title        = {Towards Revealing the Mystery behind Chain of Thought: {A} Theoretical
                  Perspective},
  booktitle    = {Advances in Neural Information Processing Systems 36: Annual Conference
                  on Neural Information Processing Systems 2023, NeurIPS 2023, New Orleans,
                  LA, USA, December 10 - 16, 2023},
  year         = {2023},
  url          = {http://papers.nips.cc/paper\_files/paper/2023/hash/dfc310e81992d2e4cedc09ac47eff13e-Abstract-Conference.html},
  bibsource    = {dblp computer science bibliography, https://dblp.org}
}

@article{openaio1,
  author       = {Aaron Jaech and
                  Adam Kalai and
                  Adam Lerer and
                  Adam Richardson and
                  Ahmed El{-}Kishky and
                  Aiden Low and
                  Alec Helyar and
                  Aleksander Madry and
                  Alex Beutel and
                  Alex Carney and
                  Alex Iftimie and
                  Alex Karpenko and
                  Alex Tachard Passos and
                  Alexander Neitz and
                  Alexander Prokofiev and
                  Alexander Wei and
                  Allison Tam and
                  Ally Bennett and
                  Ananya Kumar and
                  Andre Saraiva and
                  Andrea Vallone and
                  Andrew Duberstein and
                  Andrew Kondrich and
                  Andrey Mishchenko and
                  Andy Applebaum and
                  Angela Jiang and
                  Ashvin Nair and
                  Barret Zoph and
                  Behrooz Ghorbani and
                  Ben Rossen and
                  Benjamin Sokolowsky and
                  Boaz Barak and
                  Bob McGrew and
                  Borys Minaiev and
                  Botao Hao and
                  Bowen Baker and
                  Brandon Houghton and
                  Brandon McKinzie and
                  Brydon Eastman and
                  Camillo Lugaresi and
                  Cary Bassin and
                  Cary Hudson and
                  Chak Ming Li and
                  Charles de Bourcy and
                  Chelsea Voss and
                  Chen Shen and
                  Chong Zhang and
                  Chris Koch and
                  Chris Orsinger and
                  Christopher Hesse and
                  Claudia Fischer and
                  Clive Chan and
                  Dan Roberts and
                  Daniel Kappler and
                  Daniel Levy and
                  Daniel Selsam and
                  David Dohan and
                  David Farhi and
                  David Mely and
                  David Robinson and
                  Dimitris Tsipras and
                  Doug Li and
                  Dragos Oprica and
                  Eben Freeman and
                  Eddie Zhang and
                  Edmund Wong and
                  Elizabeth Proehl and
                  Enoch Cheung and
                  Eric Mitchell and
                  Eric Wallace and
                  Erik Ritter and
                  Evan Mays and
                  Fan Wang and
                  Felipe Petroski Such and
                  Filippo Raso and
                  Florencia Leoni and
                  Foivos Tsimpourlas and
                  Francis Song and
                  Fred von Lohmann and
                  Freddie Sulit and
                  Geoff Salmon and
                  Giambattista Parascandolo and
                  Gildas Chabot and
                  Grace Zhao and
                  Greg Brockman and
                  Guillaume Leclerc and
                  Hadi Salman and
                  Haiming Bao and
                  Hao Sheng and
                  Hart Andrin and
                  Hessam Bagherinezhad and
                  Hongyu Ren and
                  Hunter Lightman and
                  Hyung Won Chung and
                  Ian Kivlichan and
                  Ian O'Connell and
                  Ian Osband and
                  Ignasi Clavera Gilaberte and
                  Ilge Akkaya},
  title        = {OpenAI o1 System Card},
  journal      = {CoRR},
  volume       = {abs/2412.16720},
  year         = {2024},
  url          = {https://doi.org/10.48550/arXiv.2412.16720},
  doi          = {10.48550/ARXIV.2412.16720},
  eprinttype    = {arXiv},
  eprint       = {2412.16720},
  bibsource    = {dblp computer science bibliography, https://dblp.org}
}

@article{lrm_survey,
  author       = {Zhong{-}Zhi Li and
                  Duzhen Zhang and
                  Ming{-}Liang Zhang and
                  Jiaxin Zhang and
                  Zengyan Liu and
                  Yuxuan Yao and
                  Haotian Xu and
                  Junhao Zheng and
                  Pei{-}Jie Wang and
                  Xiuyi Chen and
                  Yingying Zhang and
                  Fei Yin and
                  Jiahua Dong and
                  Zhijiang Guo and
                  Le Song and
                  Cheng{-}Lin Liu},
  title        = {From System 1 to System 2: {A} Survey of Reasoning Large Language
                  Models},
  journal      = {CoRR},
  volume       = {abs/2502.17419},
  year         = {2025},
  url          = {https://doi.org/10.48550/arXiv.2502.17419},
  doi          = {10.48550/ARXIV.2502.17419},
  eprinttype    = {arXiv},
  eprint       = {2502.17419},
  bibsource    = {dblp computer science bibliography, https://dblp.org}
}

@article{lrm_survey1,
  author       = {Fengli Xu and
                  Qianyue Hao and
                  Zefang Zong and
                  Jingwei Wang and
                  Yunke Zhang and
                  Jingyi Wang and
                  Xiaochong Lan and
                  Jiahui Gong and
                  Tianjian Ouyang and
                  Fanjin Meng and
                  Chenyang Shao and
                  Yuwei Yan and
                  Qinglong Yang and
                  Yiwen Song and
                  Sijian Ren and
                  Xinyuan Hu and
                  Yu Li and
                  Jie Feng and
                  Chen Gao and
                  Yong Li},
  title        = {Towards Large Reasoning Models: {A} Survey of Reinforced Reasoning
                  with Large Language Models},
  journal      = {CoRR},
  volume       = {abs/2501.09686},
  year         = {2025},
  url          = {https://doi.org/10.48550/arXiv.2501.09686},
  doi          = {10.48550/ARXIV.2501.09686},
  eprinttype    = {arXiv},
  eprint       = {2501.09686},
  bibsource    = {dblp computer science bibliography, https://dblp.org}
}

@inproceedings{swebench,
  author       = {Carlos E. Jimenez and
                  John Yang and
                  Alexander Wettig and
                  Shunyu Yao and
                  Kexin Pei and
                  Ofir Press and
                  Karthik R. Narasimhan},
  title        = {SWE-bench: Can Language Models Resolve Real-world Github Issues?},
  booktitle    = {The Twelfth International Conference on Learning Representations,
                  {ICLR} 2024, Vienna, Austria, May 7-11, 2024},
  publisher    = {OpenReview.net},
  year         = {2024},
  url          = {https://openreview.net/forum?id=VTF8yNQM66},
  bibsource    = {dblp computer science bibliography, https://dblp.org}
}

@inproceedings{lightthinker,
    title = "{L}ight{T}hinker: Thinking Step-by-Step Compression",
    author = "Zhang, Jintian  and
      Zhu, Yuqi  and
      Sun, Mengshu  and
      Luo, Yujie  and
      Qiao, Shuofei  and
      Du, Lun  and
      Zheng, Da  and
      Chen, Huajun  and
      Zhang, Ningyu",
    editor = "Christodoulopoulos, Christos  and
      Chakraborty, Tanmoy  and
      Rose, Carolyn  and
      Peng, Violet",
    booktitle = "Proceedings of the 2025 Conference on Empirical Methods in Natural Language Processing",
    month = nov,
    year = "2025",
    address = "Suzhou, China",
    publisher = "Association for Computational Linguistics",
    url = "https://aclanthology.org/2025.emnlp-main.673/",
    doi = "10.18653/v1/2025.emnlp-main.673",
    pages = "13318--13339",
    ISBN = "979-8-89176-332-6"
}

@misc{claude3modelcard,
  title        = {The Claude 3 Model Family: Opus, Sonnet, Haiku},
  author       = {{Anthropic}},
  year         = {2024},
  url          = {https://www-cdn.anthropic.com/de8ba9b01c9ab7cbabf5c33b80b7bbc618857627/Model_Card_Claude_3.pdf}
}

@article{deepconf,
  author       = {Yichao Fu and
                  Xuewei Wang and
                  Yuandong Tian and
                  Jiawei Zhao},
  title        = {Deep Think with Confidence},
  journal      = {CoRR},
  volume       = {abs/2508.15260},
  year         = {2025},
  url          = {https://doi.org/10.48550/arXiv.2508.15260},
  doi          = {10.48550/ARXIV.2508.15260},
  eprinttype    = {arXiv},
  eprint       = {2508.15260},
  bibsource    = {dblp computer science bibliography, https://dblp.org}
}

@article{gemini,
  author       = {Gemini Team},
  title        = {Gemini 2.5: Pushing the Frontier with Advanced Reasoning, Multimodality,
                  Long Context, and Next Generation Agentic Capabilities},
  journal      = {CoRR},
  volume       = {abs/2507.06261},
  year         = {2025},
  url          = {https://doi.org/10.48550/arXiv.2507.06261},
  doi          = {10.48550/ARXIV.2507.06261},
  eprinttype    = {arXiv},
  eprint       = {2507.06261},
  bibsource    = {dblp computer science bibliography, https://dblp.org}
}

@article{l1,
  author       = {Pranjal Aggarwal and
                  Sean Welleck},
  title        = {{L1:} Controlling How Long {A} Reasoning Model Thinks With Reinforcement
                  Learning},
  journal      = {CoRR},
  volume       = {abs/2503.04697},
  year         = {2025},
  url          = {https://doi.org/10.48550/arXiv.2503.04697},
  doi          = {10.48550/ARXIV.2503.04697},
  eprinttype    = {arXiv},
  eprint       = {2503.04697},
  bibsource    = {dblp computer science bibliography, https://dblp.org}
}

@article{longcot_survey,
  author       = {Qiguang Chen and
                  Libo Qin and
                  Jinhao Liu and
                  Dengyun Peng and
                  Jiannan Guan and
                  Peng Wang and
                  Mengkang Hu and
                  Yuhang Zhou and
                  Te Gao and
                  Wanxiang Che},
  title        = {Towards Reasoning Era: {A} Survey of Long Chain-of-Thought for Reasoning
                  Large Language Models},
  journal      = {CoRR},
  volume       = {abs/2503.09567},
  year         = {2025},
  url          = {https://doi.org/10.48550/arXiv.2503.09567},
  doi          = {10.48550/ARXIV.2503.09567},
  eprinttype    = {arXiv},
  eprint       = {2503.09567},
  bibsource    = {dblp computer science bibliography, https://dblp.org}
}

@inproceedings{qiao-etal-2025-agentic,
    title = "Agentic Knowledgeable Self-awareness",
    author = "Qiao, Shuofei  and
      Qiu, Zhisong  and
      Ren, Baochang  and
      Wang, Xiaobin  and
      Ru, Xiangyuan  and
      Zhang, Ningyu  and
      Chen, Xiang  and
      Jiang, Yong  and
      Xie, Pengjun  and
      Huang, Fei  and
      Chen, Huajun",
    editor = "Che, Wanxiang  and
      Nabende, Joyce  and
      Shutova, Ekaterina  and
      Pilehvar, Mohammad Taher",
    booktitle = "Proceedings of the 63rd Annual Meeting of the Association for Computational Linguistics (Volume 1: Long Papers)",
    month = jul,
    year = "2025",
    address = "Vienna, Austria",
    publisher = "Association for Computational Linguistics",
    url = "https://aclanthology.org/2025.acl-long.619/",
    doi = "10.18653/v1/2025.acl-long.619",
    pages = "12601--12625",
    ISBN = "979-8-89176-251-0"
}

@inproceedings{loop1,
  author       = {Nikunj Saunshi and
                  Nishanth Dikkala and
                  Zhiyuan Li and
                  Sanjiv Kumar and
                  Sashank J. Reddi},
  title        = {Reasoning with Latent Thoughts: On the Power of Looped Transformers},
  booktitle    = {The Thirteenth International Conference on Learning Representations,
                  {ICLR} 2025, Singapore, April 24-28, 2025},
  publisher    = {OpenReview.net},
  year         = {2025},
  url          = {https://openreview.net/forum?id=din0lGfZFd},
  bibsource    = {dblp computer science bibliography, https://dblp.org}
}

@article{loop2,
  author       = {Rui{-}Jie Zhu and
                  Zixuan Wang and
                  Kai Hua and
                  Tianyu Zhang and
                  Ziniu Li and
                  Haoran Que and
                  Boyi Wei and
                  Zixin Wen and
                  Fan Yin and
                  He Xing and
                  Lu Li and
                  Jiajun Shi and
                  Kaijing Ma and
                  Shanda Li and
                  Taylor Kergan and
                  Andrew Smith and
                  Xingwei Qu and
                  Mude Hui and
                  Bohong Wu and
                  Qiyang Min and
                  Hongzhi Huang and
                  Xun Zhou and
                  Wei Ye and
                  Jiaheng Liu and
                  Jian Yang and
                  Yunfeng Shi and
                  Chenghua Lin and
                  Enduo Zhao and
                  Tianle Cai and
                  Ge Zhang and
                  Wenhao Huang and
                  Yoshua Bengio and
                  Jason Eshraghian},
  title        = {Scaling Latent Reasoning via Looped Language Models},
  journal      = {CoRR},
  volume       = {abs/2510.25741},
  year         = {2025},
  url          = {https://doi.org/10.48550/arXiv.2510.25741},
  doi          = {10.48550/ARXIV.2510.25741},
  eprinttype    = {arXiv},
  eprint       = {2510.25741},
  bibsource    = {dblp computer science bibliography, https://dblp.org}
}

@article{DBLP:journals/bstj/Shannon48,
  author       = {Claude E. Shannon},
  title        = {A mathematical theory of communication},
  journal      = {Bell Syst. Tech. J.},
  volume       = {27},
  number       = {3},
  pages        = {379--423},
  year         = {1948},
  url          = {https://doi.org/10.1002/j.1538-7305.1948.tb01338.x},
  doi          = {10.1002/J.1538-7305.1948.TB01338.X},
  bibsource    = {dblp computer science bibliography, https://dblp.org}
}

@inproceedings{DBLP:conf/iclr/HendrycksG17,
  author       = {Dan Hendrycks and
                  Kevin Gimpel},
  title        = {A Baseline for Detecting Misclassified and Out-of-Distribution Examples
                  in Neural Networks},
  booktitle    = {5th International Conference on Learning Representations, {ICLR} 2017,
                  Toulon, France, April 24-26, 2017, Conference Track Proceedings},
  publisher    = {OpenReview.net},
  year         = {2017},
  url          = {https://openreview.net/forum?id=Hkg4TI9xl},
  bibsource    = {dblp computer science bibliography, https://dblp.org}
}

@inproceedings{DBLP:conf/icml/GuoPSW17,
  author       = {Chuan Guo and
                  Geoff Pleiss and
                  Yu Sun and
                  Kilian Q. Weinberger},
  editor       = {Doina Precup and
                  Yee Whye Teh},
  title        = {On Calibration of Modern Neural Networks},
  booktitle    = {Proceedings of the 34th International Conference on Machine Learning,
                  {ICML} 2017, Sydney, NSW, Australia, 6-11 August 2017},
  series       = {Proceedings of Machine Learning Research},
  volume       = {70},
  pages        = {1321--1330},
  publisher    = {{PMLR}},
  year         = {2017},
  url          = {http://proceedings.mlr.press/v70/guo17a.html},
  bibsource    = {dblp computer science bibliography, https://dblp.org}
}

@article{DBLP:journals/tacl/JiangXAN20,
  author       = {Zhengbao Jiang and
                  Frank F. Xu and
                  Jun Araki and
                  Graham Neubig},
  title        = {How Can We Know What Language Models Know},
  journal      = {Trans. Assoc. Comput. Linguistics},
  volume       = {8},
  pages        = {423--438},
  year         = {2020},
  url          = {https://doi.org/10.1162/tacl\_a\_00324},
  doi          = {10.1162/TACL\_A\_00324},
  bibsource    = {dblp computer science bibliography, https://dblp.org}
}

@inproceedings{DBLP:conf/icml/GalG16,
  author       = {Yarin Gal and
                  Zoubin Ghahramani},
  editor       = {Maria{-}Florina Balcan and
                  Kilian Q. Weinberger},
  title        = {Dropout as a Bayesian Approximation: Representing Model Uncertainty
                  in Deep Learning},
  booktitle    = {Proceedings of the 33nd International Conference on Machine Learning,
                  {ICML} 2016, New York City, NY, USA, June 19-24, 2016},
  series       = {{JMLR} Workshop and Conference Proceedings},
  volume       = {48},
  pages        = {1050--1059},
  publisher    = {JMLR.org},
  year         = {2016},
  url          = {http://proceedings.mlr.press/v48/gal16.html},
  bibsource    = {dblp computer science bibliography, https://dblp.org}
}

@article{DBLP:journals/corr/abs-2510-01123,
  author       = {Lovish Madaan and
                  Aniket Didolkar and
                  Suchin Gururangan and
                  John Quan and
                  Ruan Silva and
                  Ruslan Salakhutdinov and
                  Manzil Zaheer and
                  Sanjeev Arora and
                  Anirudh Goyal},
  title        = {Rethinking Thinking Tokens: LLMs as Improvement Operators},
  journal      = {CoRR},
  volume       = {abs/2510.01123},
  year         = {2025},
  url          = {https://doi.org/10.48550/arXiv.2510.01123},
  doi          = {10.48550/ARXIV.2510.01123},
  eprinttype    = {arXiv},
  eprint       = {2510.01123},
  bibsource    = {dblp computer science bibliography, https://dblp.org}
}

@article{DBLP:journals/corr/abs-2502-06772,
  author       = {Ling Yang and
                  Zhaochen Yu and
                  Bin Cui and
                  Mengdi Wang},
  title        = {ReasonFlux: Hierarchical {LLM} Reasoning via Scaling Thought Templates},
  journal      = {CoRR},
  volume       = {abs/2502.06772},
  year         = {2025},
  url          = {https://doi.org/10.48550/arXiv.2502.06772},
  doi          = {10.48550/ARXIV.2502.06772},
  eprinttype    = {arXiv},
  eprint       = {2502.06772},
  bibsource    = {dblp computer science bibliography, https://dblp.org}
}

@article{gemma3,
  author       = {Gemma Team},
  title        = {Gemma 3 Technical Report},
  journal      = {CoRR},
  volume       = {abs/2503.19786},
  year         = {2025},
  url          = {https://doi.org/10.48550/arXiv.2503.19786},
  doi          = {10.48550/ARXIV.2503.19786},
  eprinttype    = {arXiv},
  eprint       = {2503.19786},
  bibsource    = {dblp computer science bibliography, https://dblp.org}
}

@article{conditional_memory,
  author       = {Xin Cheng and Wangding Zeng and Damai Dai and Qinyu Chen and Bingxuan Wang and Zhenda Xie and Kezhao Huang and Xingkai Yu and Zhewen Hao and Yukun Li and Han Zhang and Huishuai Zhang and Dongyan Zhao and Wenfeng Liang
},
  title        = {Conditional Memory via Scalable Lookup: A New Axis of Sparsity for Large Language Models},
  journal      = {CoRR},
  volume       = {abs/2601.07372},
  year         = {2026},
  url          = {https://arxiv.org/pdf/2601.07372},
  eprinttype    = {arXiv},
  eprint       = {2601.07372}
}

@article{DBLP:journals/corr/abs-2510-14901,
  author       = {Aayush Karan and
                  Yilun Du},
  title        = {Reasoning with Sampling: Your Base Model is Smarter Than You Think},
  journal      = {CoRR},
  volume       = {abs/2510.14901},
  year         = {2025},
  url          = {https://doi.org/10.48550/arXiv.2510.14901},
  doi          = {10.48550/ARXIV.2510.14901},
  eprinttype    = {arXiv},
  eprint       = {2510.14901},
  bibsource    = {dblp computer science bibliography, https://dblp.org}
}

@article{DBLP:journals/corr/abs-2310-02170,
  author       = {Zijun Liu and
                  Yanzhe Zhang and
                  Peng Li and
                  Yang Liu and
                  Diyi Yang},
  title        = {Dynamic LLM-Agent Network: An LLM-agent Collaboration Framework with
                  Agent Team Optimization},
  journal      = {CoRR},
  volume       = {abs/2310.02170},
  year         = {2023},
  url          = {https://doi.org/10.48550/arXiv.2310.02170},
  doi          = {10.48550/ARXIV.2310.02170},
  eprinttype    = {arXiv},
  eprint       = {2310.02170},
  bibsource    = {dblp computer science bibliography, https://dblp.org}
}

@article{llama,
  author       = {Hugo Touvron and
                  Thibaut Lavril and
                  Gautier Izacard and
                  Xavier Martinet and
                  Marie{-}Anne Lachaux and
                  Timoth{\'{e}}e Lacroix and
                  Baptiste Rozi{\`{e}}re and
                  Naman Goyal and
                  Eric Hambro and
                  Faisal Azhar and
                  Aur{\'{e}}lien Rodriguez and
                  Armand Joulin and
                  Edouard Grave and
                  Guillaume Lample},
  title        = {LLaMA: Open and Efficient Foundation Language Models},
  journal      = {CoRR},
  volume       = {abs/2302.13971},
  year         = {2023},
  url          = {https://doi.org/10.48550/arXiv.2302.13971},
  doi          = {10.48550/ARXIV.2302.13971},
  eprinttype    = {arXiv},
  eprint       = {2302.13971},
  bibsource    = {dblp computer science bibliography, https://dblp.org}
}

@article{DBLP:journals/tacl/VashurinFVRVTPXSGPBNPS25,
  author       = {Roman Vashurin and
                  Ekaterina Fadeeva and
                  Artem Vazhentsev and
                  Lyudmila Rvanova and
                  Daniil Vasilev and
                  Akim Tsvigun and
                  Sergey Petrakov and
                  Rui Xing and
                  Abdelrahman Boda Sadallah and
                  Kirill Grishchenkov and
                  Alexander Panchenko and
                  Timothy Baldwin and
                  Preslav Nakov and
                  Maxim Panov and
                  Artem Shelmanov},
  title        = {Benchmarking Uncertainty Quantification Methods for Large Language
                  Models with LM-Polygraph},
  journal      = {Trans. Assoc. Comput. Linguistics},
  volume       = {13},
  pages        = {220--248},
  year         = {2025},
  url          = {https://doi.org/10.1162/tacl\_a\_00737},
  doi          = {10.1162/TACL\_A\_00737},
  bibsource    = {dblp computer science bibliography, https://dblp.org}
}

@inproceedings{top-k,
  author       = {Angela Fan and
                  Mike Lewis and
                  Yann N. Dauphin},
  editor       = {Iryna Gurevych and
                  Yusuke Miyao},
  title        = {Hierarchical Neural Story Generation},
  booktitle    = {Proceedings of the 56th Annual Meeting of the Association for Computational
                  Linguistics, {ACL} 2018, Melbourne, Australia, July 15-20, 2018, Volume
                  1: Long Papers},
  pages        = {889--898},
  publisher    = {Association for Computational Linguistics},
  year         = {2018},
  url          = {https://aclanthology.org/P18-1082/},
  doi          = {10.18653/V1/P18-1082},
  bibsource    = {dblp computer science bibliography, https://dblp.org}
}

@inproceedings{top-p,
  author       = {Ari Holtzman and
                  Jan Buys and
                  Li Du and
                  Maxwell Forbes and
                  Yejin Choi},
  title        = {The Curious Case of Neural Text Degeneration},
  booktitle    = {8th International Conference on Learning Representations, {ICLR} 2020,
                  Addis Ababa, Ethiopia, April 26-30, 2020},
  publisher    = {OpenReview.net},
  year         = {2020},
  url          = {https://openreview.net/forum?id=rygGQyrFvH},
  bibsource    = {dblp computer science bibliography, https://dblp.org}
}

@inproceedings{min-p,
  author       = {Nguyen Nhat Minh and
                  Andrew Baker and
                  Clement Neo and
                  Allen G. Roush and
                  Andreas Kirsch and
                  Ravid Shwartz{-}Ziv},
  title        = {Turning Up the Heat: Min-p Sampling for Creative and Coherent {LLM}
                  Outputs},
  booktitle    = {The Thirteenth International Conference on Learning Representations,
                  {ICLR} 2025, Singapore, April 24-28, 2025},
  publisher    = {OpenReview.net},
  year         = {2025},
  url          = {https://openreview.net/forum?id=FBkpCyujtS},
  bibsource    = {dblp computer science bibliography, https://dblp.org}
}

@inproceedings{DBLP:conf/iclr/YangDSC18,
  author       = {Zhilin Yang and
                  Zihang Dai and
                  Ruslan Salakhutdinov and
                  William W. Cohen},
  title        = {Breaking the Softmax Bottleneck: {A} High-Rank {RNN} Language Model},
  booktitle    = {6th International Conference on Learning Representations, {ICLR} 2018,
                  Vancouver, BC, Canada, April 30 - May 3, 2018, Conference Track Proceedings},
  publisher    = {OpenReview.net},
  year         = {2018},
  url          = {https://openreview.net/forum?id=HkwZSG-CZ},
  bibsource    = {dblp computer science bibliography, https://dblp.org}
}

@inproceedings{DBLP:conf/iclr/AroraLM17,
  author       = {Sanjeev Arora and
                  Yingyu Liang and
                  Tengyu Ma},
  title        = {A Simple but Tough-to-Beat Baseline for Sentence Embeddings},
  booktitle    = {5th International Conference on Learning Representations, {ICLR} 2017,
                  Toulon, France, April 24-26, 2017, Conference Track Proceedings},
  publisher    = {OpenReview.net},
  year         = {2017},
  url          = {https://openreview.net/forum?id=SyK00v5xx},
  bibsource    = {dblp computer science bibliography, https://dblp.org}
}

@inproceedings{DBLP:conf/emnlp/HaoGMHWWH23,
  author       = {Shibo Hao and
                  Yi Gu and
                  Haodi Ma and
                  Joshua Jiahua Hong and
                  Zhen Wang and
                  Daisy Zhe Wang and
                  Zhiting Hu},
  editor       = {Houda Bouamor and
                  Juan Pino and
                  Kalika Bali},
  title        = {Reasoning with Language Model is Planning with World Model},
  booktitle    = {Proceedings of the 2023 Conference on Empirical Methods in Natural
                  Language Processing, {EMNLP} 2023, Singapore, December 6-10, 2023},
  pages        = {8154--8173},
  publisher    = {Association for Computational Linguistics},
  year         = {2023},
  url          = {https://doi.org/10.18653/v1/2023.emnlp-main.507},
  doi          = {10.18653/V1/2023.EMNLP-MAIN.507},
  bibsource    = {dblp computer science bibliography, https://dblp.org}
}

@inproceedings{DBLP:conf/nips/BengioVJS15,
  author       = {Samy Bengio and
                  Oriol Vinyals and
                  Navdeep Jaitly and
                  Noam Shazeer},
  editor       = {Corinna Cortes and
                  Neil D. Lawrence and
                  Daniel D. Lee and
                  Masashi Sugiyama and
                  Roman Garnett},
  title        = {Scheduled Sampling for Sequence Prediction with Recurrent Neural Networks},
  booktitle    = {Advances in Neural Information Processing Systems 28: Annual Conference
                  on Neural Information Processing Systems 2015, December 7-12, 2015,
                  Montreal, Quebec, Canada},
  pages        = {1171--1179},
  year         = {2015},
  url          = {https://proceedings.neurips.cc/paper/2015/hash/e995f98d56967d946471af29d7bf99f1-Abstract.html},
  bibsource    = {dblp computer science bibliography, https://dblp.org}
}

@inproceedings{DBLP:conf/emnlp/Schmidt19,
  author       = {Florian Schmidt},
  editor       = {Alexandra Birch and
                  Andrew M. Finch and
                  Hiroaki Hayashi and
                  Ioannis Konstas and
                  Thang Luong and
                  Graham Neubig and
                  Yusuke Oda and
                  Katsuhito Sudoh},
  title        = {Generalization in Generation: {A} closer look at Exposure Bias},
  booktitle    = {Proceedings of the 3rd Workshop on Neural Generation and Translation@EMNLP-IJCNLP
                  2019, Hong Kong, November 4, 2019},
  pages        = {157--167},
  publisher    = {Association for Computational Linguistics},
  year         = {2019},
  url          = {https://doi.org/10.18653/v1/D19-5616},
  doi          = {10.18653/V1/D19-5616},
  bibsource    = {dblp computer science bibliography, https://dblp.org}
}

@inproceedings{DBLP:conf/nips/LimEHM21,
  author       = {Soon Hoe Lim and
                  N. Benjamin Erichson and
                  Liam Hodgkinson and
                  Michael W. Mahoney},
  editor       = {Marc'Aurelio Ranzato and
                  Alina Beygelzimer and
                  Yann N. Dauphin and
                  Percy Liang and
                  Jennifer Wortman Vaughan},
  title        = {Noisy Recurrent Neural Networks},
  booktitle    = {Advances in Neural Information Processing Systems 34: Annual Conference
                  on Neural Information Processing Systems 2021, NeurIPS 2021, December
                  6-14, 2021, virtual},
  pages        = {5124--5137},
  year         = {2021},
  url          = {https://proceedings.neurips.cc/paper/2021/hash/29301521774ff3cbd26652b2d5c95996-Abstract.html},
  bibsource    = {dblp computer science bibliography, https://dblp.org}
}

@article{DBLP:journals/corr/abs-2509-06770,
  author       = {Shashidhar Reddy Javaji and
                  Bhavul Gauri and
                  Zining Zhu},
  title        = {Another Turn, Better Output? {A} Turn-Wise Analysis of Iterative {LLM}
                  Prompting},
  journal      = {CoRR},
  volume       = {abs/2509.06770},
  year         = {2025},
  url          = {https://doi.org/10.48550/arXiv.2509.06770},
  doi          = {10.48550/ARXIV.2509.06770},
  eprinttype    = {arXiv},
  eprint       = {2509.06770},
  bibsource    = {dblp computer science bibliography, https://dblp.org}
}

@article{DBLP:journals/tois/HuangYMZFWCPFQL25,
  author       = {Lei Huang and
                  Weijiang Yu and
                  Weitao Ma and
                  Weihong Zhong and
                  Zhangyin Feng and
                  Haotian Wang and
                  Qianglong Chen and
                  Weihua Peng and
                  Xiaocheng Feng and
                  Bing Qin and
                  Ting Liu},
  title        = {A Survey on Hallucination in Large Language Models: Principles, Taxonomy,
                  Challenges, and Open Questions},
  journal      = {{ACM} Trans. Inf. Syst.},
  volume       = {43},
  number       = {2},
  pages        = {42:1--42:55},
  year         = {2025},
  url          = {https://doi.org/10.1145/3703155},
  doi          = {10.1145/3703155},
  bibsource    = {dblp computer science bibliography, https://dblp.org}
}

@article{DBLP:journals/corr/abs-2509-04664,
  author       = {Adam Tauman Kalai and
                  Ofir Nachum and
                  Santosh S. Vempala and
                  Edwin Zhang},
  title        = {Why Language Models Hallucinate},
  journal      = {CoRR},
  volume       = {abs/2509.04664},
  year         = {2025},
  url          = {https://doi.org/10.48550/arXiv.2509.04664},
  doi          = {10.48550/ARXIV.2509.04664},
  eprinttype    = {arXiv},
  eprint       = {2509.04664},
  bibsource    = {dblp computer science bibliography, https://dblp.org}
}

@article{DBLP:journals/corr/abs-2505-13143,
  author       = {Haolang Lu and
                  Yilian Liu and
                  Jingxin Xu and
                  Guoshun Nan and
                  Yuanlong Yu and
                  Zhican Chen and
                  Kun Wang},
  title        = {Auditing Meta-Cognitive Hallucinations in Reasoning Large Language
                  Models},
  journal      = {CoRR},
  volume       = {abs/2505.13143},
  year         = {2025},
  url          = {https://doi.org/10.48550/arXiv.2505.13143},
  doi          = {10.48550/ARXIV.2505.13143},
  eprinttype    = {arXiv},
  eprint       = {2505.13143},
  bibsource    = {dblp computer science bibliography, https://dblp.org}
}

@article{swireasoning,
  author       = {Dachuan Shi and
                  Abedelkadir Asi and
                  Keying Li and
                  Xiangchi Yuan and
                  Leyan Pan and
                  Wenke Lee and
                  Wen Xiao},
  title        = {SwiReasoning: Switch-Thinking in Latent and Explicit for Pareto-Superior
                  Reasoning LLMs},
  journal      = {CoRR},
  volume       = {abs/2510.05069},
  year         = {2025},
  url          = {https://doi.org/10.48550/arXiv.2510.05069},
  doi          = {10.48550/ARXIV.2510.05069},
  eprinttype    = {arXiv},
  eprint       = {2510.05069},
  bibsource    = {dblp computer science bibliography, https://dblp.org}
}

@article{DBLP:journals/corr/abs-2508-03440,
  author       = {Ch{\"{u}}nhung Wu and
                  Jinliang Lu and
                  Zixuan Ren and
                  Gangqiang Hu and
                  Zhi Wu and
                  Dai Dai and
                  Hua Wu},
  title        = {LLMs are Single-threaded Reasoners: Demystifying the Working Mechanism
                  of Soft Thinking},
  journal      = {CoRR},
  volume       = {abs/2508.03440},
  year         = {2025},
  url          = {https://doi.org/10.48550/arXiv.2508.03440},
  doi          = {10.48550/ARXIV.2508.03440},
  eprinttype    = {arXiv},
  eprint       = {2508.03440},
  bibsource    = {dblp computer science bibliography, https://dblp.org}
}

@article{DBLP:journals/corr/abs-2505-12514,
  author       = {Hanlin Zhu and
                  Shibo Hao and
                  Zhiting Hu and
                  Jiantao Jiao and
                  Stuart Russell and
                  Yuandong Tian},
  title        = {Reasoning by Superposition: {A} Theoretical Perspective on Chain of
                  Continuous Thought},
  journal      = {CoRR},
  volume       = {abs/2505.12514},
  year         = {2025},
  url          = {https://doi.org/10.48550/arXiv.2505.12514},
  doi          = {10.48550/ARXIV.2505.12514},
  eprinttype    = {arXiv},
  eprint       = {2505.12514},
  bibsource    = {dblp computer science bibliography, https://dblp.org}
}

@inproceedings{lars,
    title = "{L}a{RS}: Latent Reasoning Skills for Chain-of-Thought Reasoning",
    author = "Xu, Zifan  and
      Wang, Haozhu  and
      Bespalov, Dmitriy  and
      Wu, Xian  and
      Stone, Peter  and
      Qi, Yanjun",
    editor = "Al-Onaizan, Yaser  and
      Bansal, Mohit  and
      Chen, Yun-Nung",
    booktitle = "Findings of the Association for Computational Linguistics: EMNLP 2024",
    month = nov,
    year = "2024",
    address = "Miami, Florida, USA",
    publisher = "Association for Computational Linguistics",
    url = "https://aclanthology.org/2024.findings-emnlp.206/",
    doi = "10.18653/v1/2024.findings-emnlp.206",
    pages = "3624--3643"
}
\bibliographystyle{icml2026}

\newpage
\appendix
\onecolumn

\section{Preliminary Experiments}
\label{app:pre_exp}

\begin{figure}[htbp]
\centering
    \begin{subfigure}[b]{0.41\textwidth}
        \centering
        \includegraphics[width=\textwidth]{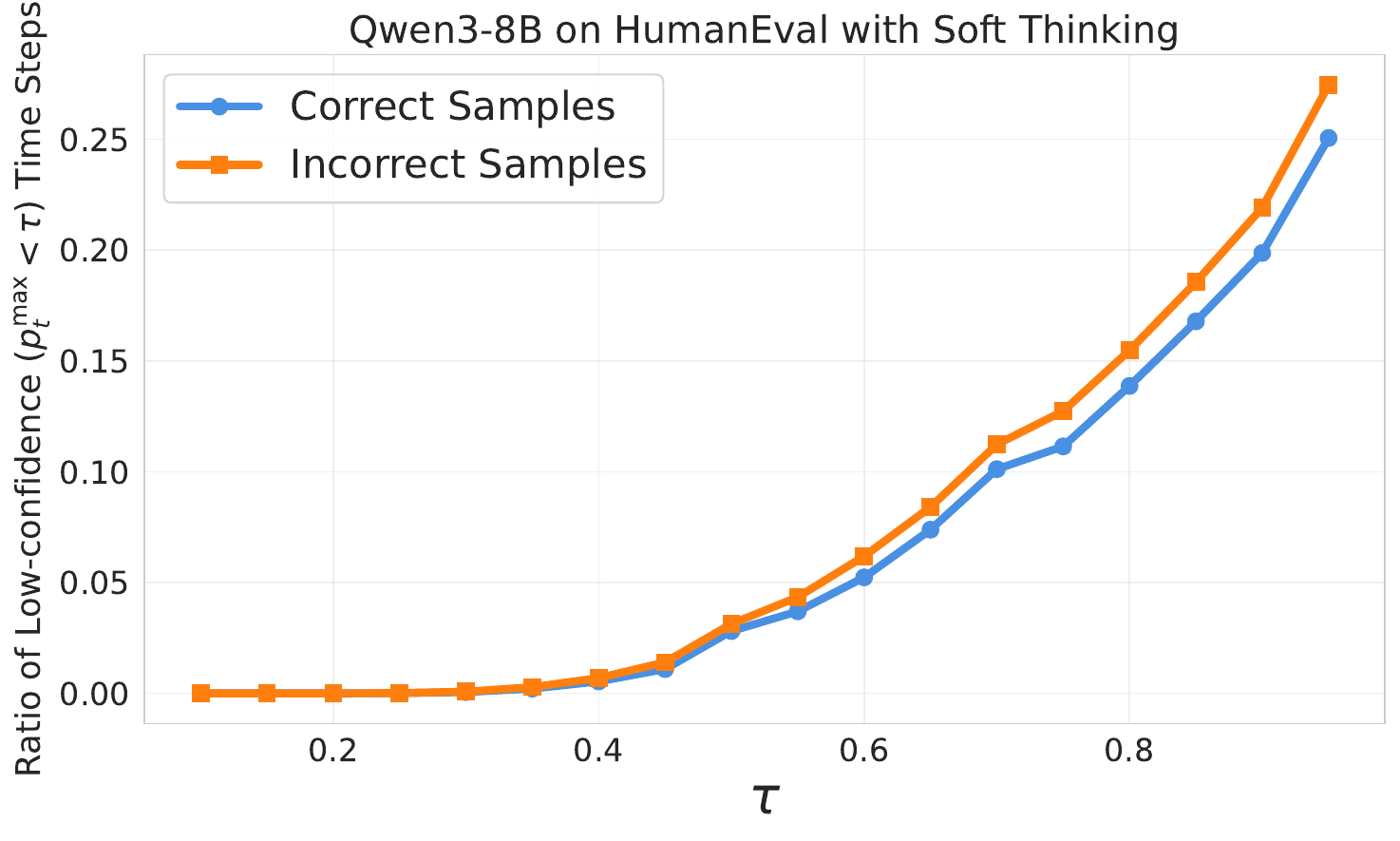}
    \end{subfigure}
    \hspace{0.05\textwidth}
    \begin{subfigure}[b]{0.41\textwidth}
        \centering
        \includegraphics[width=\textwidth]{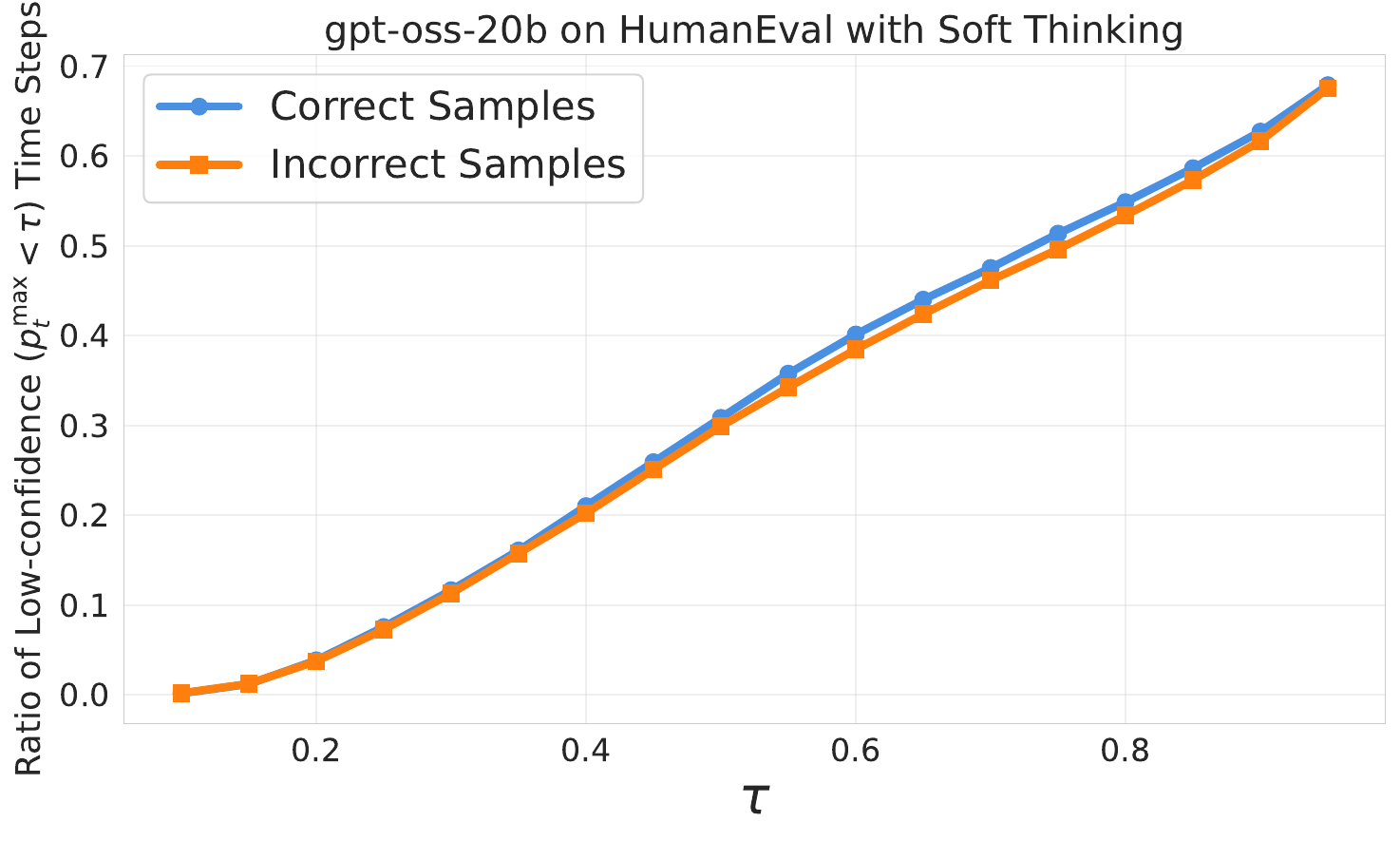}
    \end{subfigure}
    \caption{Ratio of low-confidence time steps of HumanEval within reasoning trajectories under latent-only reasoning (\textit{Soft Thinking}).}
    \label{fig:low_ratio_gptoss}
\end{figure}

One time step $t$ is classified as a low-confidence time step if the maximum next-token 
probability $p_t^{\max}$ is lower than $\tau$.
Figure \ref{fig:low_ratio_gptoss} shows the ratio of low-confidence time steps in reasoning trajectories evaluated on HumanEval.
We follow the same setting in \S \ref{sec:experi_setup}.
For each LRM, we run each sample in each dataset three times.
We record the next-token distribution at every time step for all three runs, and use the collected next-token distributions from all three runs to construct Figures \ref{fig:low_confidence_ratio} and \ref{fig:low_ratio_gptoss}.
Specifically, let $\mathcal{C}$ denote the set of maximum next-token probabilities from all thinking time steps ending in correct answers, and $\mathcal{I}$ denote such a set for incorrect ones.
The ratio of low-confidence time steps for correct samples is $\frac{1}{|\mathcal{C}|}\sum\limits_{p_t^{\max} \in \mathcal{C}} \mathbb{I}[p_t^{\max} < \tau]$.
The ratio of low-confidence time steps for correct samples is $\frac{1}{|\mathcal{I}|}\sum\limits_{p_t^{\max} \in \mathcal{I}} \mathbb{I}[p_t^{\max} < \tau]$.




\section{Main Experiments}
\label{app:exper}
\subsection{More Details in \ours\ Algorithm}
\label{app:algorithm}
\paragraph{\textsc{Sample}} This is renormalization filtering with the sampling strategies, which uses the same sampling parameters of CoT (sampling) in Table \ref{tab:sampling_param}.

\paragraph{\textsc{MultinomialSample}} 
Following \textit{Soft Thinking}, we use multinomial sampling with \texttt{torch.multinomial}\footnote{\url{https://docs.pytorch.org/docs/stable/generated/torch.multinomial.html}} if thinking performs in a discrete token space. 

\paragraph{\textsc{ColdStop}}
We use \textit{Cold Stop} from \textit{Soft Thinking} to stop intermediate thinking when the model becomes continuously overconfident. 
At each time step $t$, we compute the entropy of the next-token probability over the vocabulary:
\begin{equation}
    H(p_t)=-\sum\limits_{v \in \mathcal{V}}p_t[v]\log p_t[v]
\end{equation}
Low entropy suggests that the model is of high confidence in its prediction \citep{DBLP:journals/bstj/Shannon48}.
If $H(p_t)$ is lower than an entropy threshold $\delta$, a low-entropy step counter increases; otherwise, the counter is reset.
When the counter reaches $l$ consecutive confident steps, the end-of-thinking token is inserted to stop thinking, and then final answer generation begins.
\textsc{ColdStop} is applied for \ours, Random Routing, and \textit{Soft Thinking} in this work.

\paragraph{\textsc{Decode}} After thinking, the LRMs perform the standard autoregressive decoding in a discrete token space with the official sampling strategy in Table \ref{tab:sampling_param}.

\subsection{Datasets}
\label{app:datasets}
To comprehensively evaluate \ours, three STEM reasoning benchmarks and two coding benchmarks are used, spanning different domains and scales to assess robustness and generality.

\paragraph{STEM Reasoning Benchmarks}
AIME 2024 and AIME 2025 are challenging mathematical reasoning benchmarks derived from the American Invitational Mathematics Examination (AIME), a prestigious U.S. mathematics competition.
Each AIME exam consists of 15 problems requiring an exact integer answer between 000 and 999, with no partial credit, emphasizing precise multi-step reasoning.
Due to their high difficulty and strict answer format, AIME 2024 and AIME 2025 are widely adopted to evaluate advanced mathematical reasoning \citep{deepconf, DBLP:journals/corr/abs-2502-06772, DBLP:journals/corr/abs-2510-01123}.
GPQA Diamond \citep{gpqa} is the most challenging subset of the GPQA benchmark, consisting of graduate-level, Google-proof multiple-choice questions that require deep domain knowledge (STEM) and multi-step reasoning, and is widely used to stress-test the reasoning robustness of large language models \cite{gemma3}.

\paragraph{Coding Benchmarks}
HumanEval \citep{humaneval} is a widely used code generation benchmark consisting of hand-written Python programming problems that require function-level reasoning and exact-match execution correctness \citep{DBLP:journals/corr/abs-2510-14901, conditional_memory, DBLP:journals/corr/abs-2310-02170}.
MBPP (Mostly Basic Programming Problems) is also a code generation benchmark composed of short Python programming tasks with natural language descriptions and test cases, designed to evaluate basic algorithmic reasoning and functional correctness of large language models \citep{llama}.

The number of samples in each dataset is shown in Table \ref{tab:dataset_number}.
The datasets and the implementation of the evaluation in our experiments follow \url{https://github.com/eric-ai-lab/Soft-Thinking/tree/main}.

\begin{table}[htbp]
  \centering
  \caption{The statistic distributions of the datasets.}
    \begin{tabular}{lc}
    \toprule
    \textbf{Dataset} & \textbf{\# Sample} \\
    \midrule
    AIME 2024 & 30 \\
    AIME 2025 & 30 \\
    GPQA Diamond & 198 \\
    HumanEval & 164 \\
    MBPP  & 257 \\
    \bottomrule
    \end{tabular}%
  \label{tab:dataset_number}
\end{table}

\subsection{Baselines and Hyper-parameters}
\label{app:baseline}
\paragraph{CoT (sampling)} This is a standard decoding baseline only in the discrete token space with chain-of-thought thinking and sampling strategies, including top-k \citep{top-k}, top-p \citep{top-p}, and min-p \citep{min-p}.
We use the official sampling strategies of Qwen3-8B\footnote{\url{https://huggingface.co/Qwen/Qwen3-8B}} and gpt-oss-20b\footnote{\url{https://github.com/openai/gpt-oss}} for the CoT (sampling) baseline.
The specific sampling parameters are shown in Table \ref{tab:sampling_param}.

\begin{table}[ht!]
  \centering
  \caption{The sampling parameters.}
    \begin{tabular}{lccccc}
    \toprule
          & \textbf{Temperature} & \textbf{Top-k} & \textbf{Top-p} & \textbf{Min-p} & \textbf{Max output length} \\
    \midrule
    \textbf{Qwen3-8B} & 0.6   & 20    & 0.95  & 0.0   & 32,768 \\
    \textbf{gpt-oss-20b} & 1.0   & 20    & 1.00  & 0.0   & 32,768 \\
    \bottomrule
    \end{tabular}%
  \label{tab:sampling_param}%
\end{table}%

\paragraph{CoT (greedy)} This is a standard decoding baseline only in the discrete token space with chain-of-thought thinking and greedy decoding.

\paragraph{\textit{Soft Thinking}} We use the official code\footnote{\url{https://github.com/eric-ai-lab/Soft-Thinking/tree/main}} to conduct \textit{Soft Thinking}.
\textit{Soft Thinking} introduce four hyper-parameters: top-j (Equation \ref{eq:3}), entropy threshold $\delta$, and maximum consecutive confident step $l$.
Corresponding to the \textit{Soft Thinking} paper~\citep{soft-thinking}, grid search over these parameters would require evaluating $5\times4\times4$ configurations per model per dataset, leading to substantial computational overhead. 
Since latent-space reasoning implementation in \ours\ directly follows \textit{Soft Thinking}, the comparison between \ours\ and \textit{Soft Thinking} primarily aims to isolate the effect of confidence-aware hybrid-space reasoning versus latent-only reasoning, rather than to exhaustively optimize\textit{ Soft Thinking} itself.
Therefore, hyperparameter tuning for \textit{Soft Thinking} is not necessary.
Instead, we directly adopt the hyperparameter configuration from the official GitHub codebase and apply it uniformly across all models and datasets, i.e., $j=10, \delta=0.01, l=256$ without hyperparameter tuning.
After thinking, the sampling strategies for generating final answers are same as CoT (sampling).

\paragraph{Random Routing} For each step during thinking, we uniformly sample a binary routing decision from $\{0,1\}$ using \texttt{torch.randint}\footnote{\url{https://docs.pytorch.org/docs/stable/generated/torch.randint.html}} to choose between a latent space and a discrete token space.

\subsection{\ours\ Implementation with \textit{Soft Thinking}}
\label{app:implementation}

\begin{figure}[htbp]
\centering
    \begin{subfigure}[b]{0.43\textwidth}
        \centering
        \includegraphics[width=\textwidth]{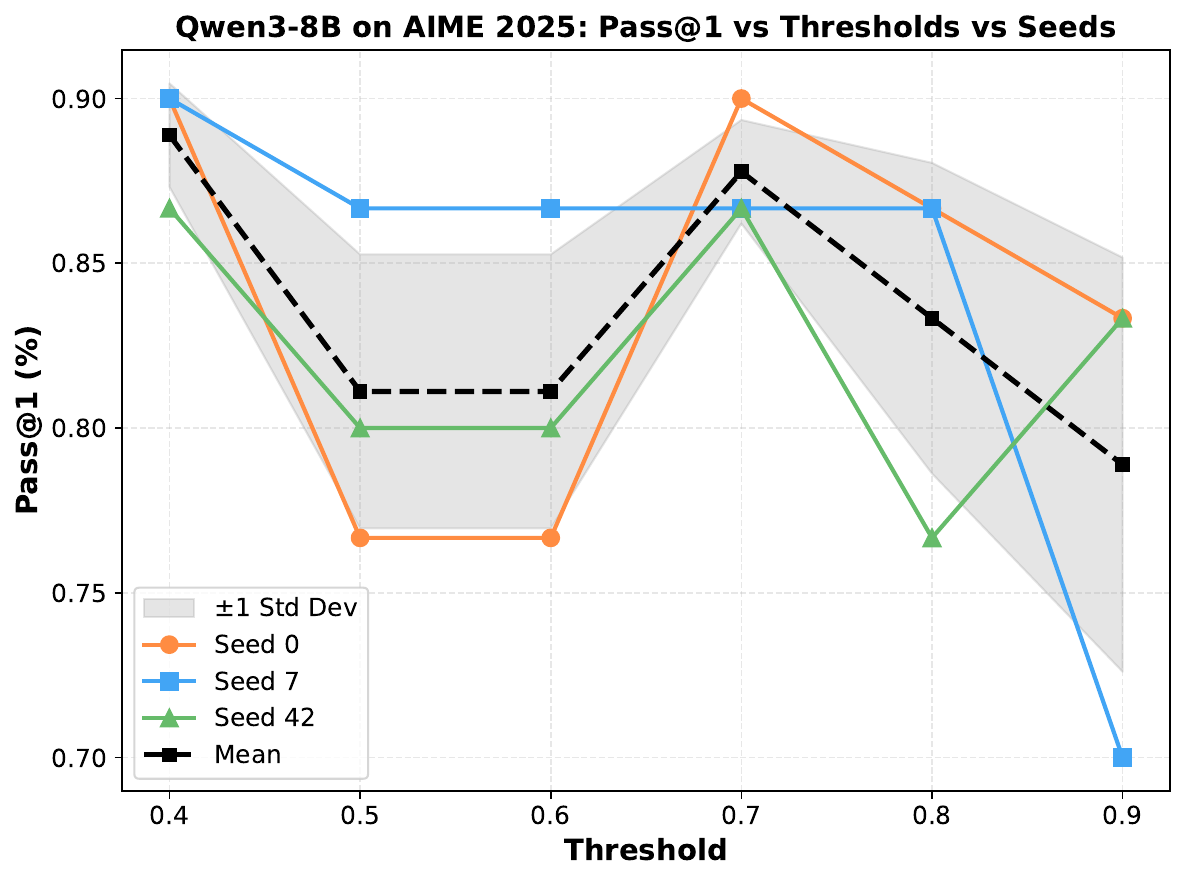}
    \end{subfigure}
    \hspace{0.1\textwidth}
    \begin{subfigure}[b]{0.43\textwidth}
        \centering
        \includegraphics[width=\textwidth]{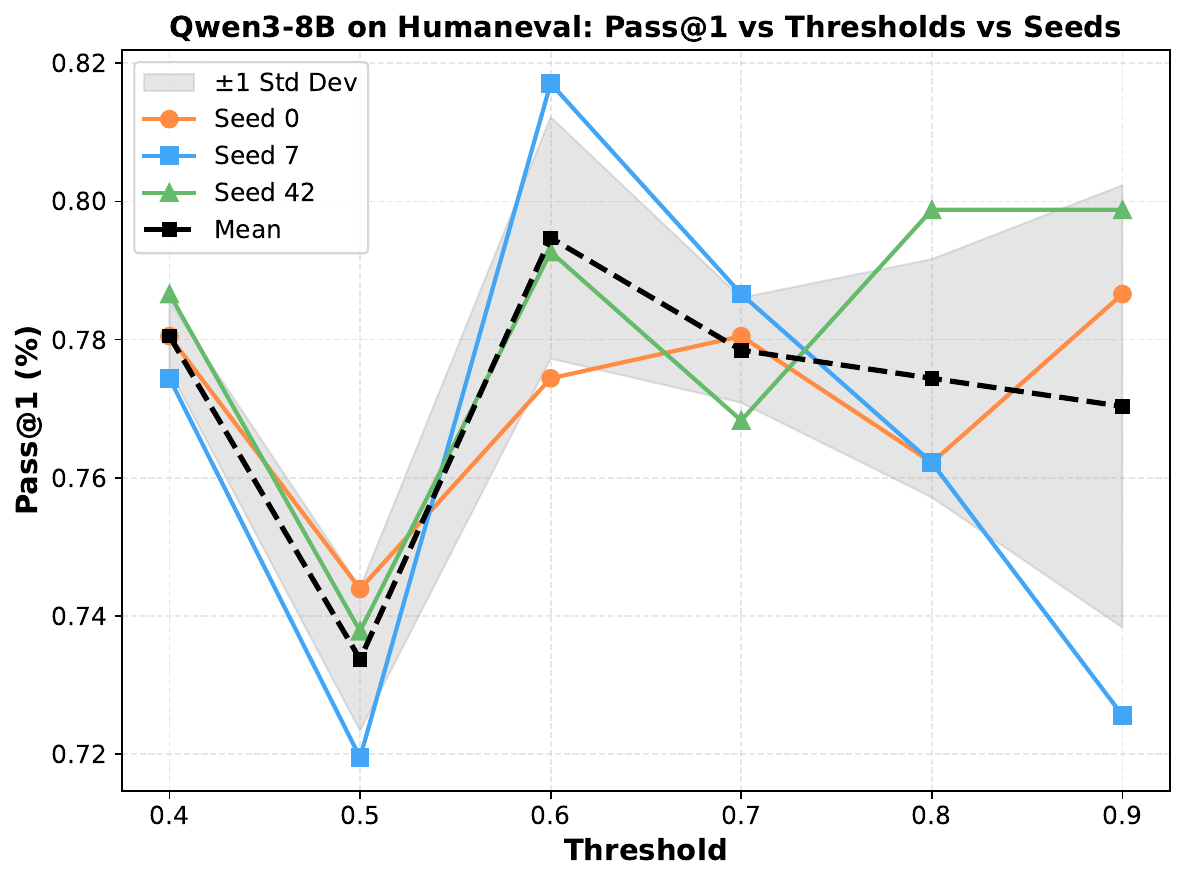}
    \end{subfigure}
    \caption{Impact of the routing thresholds $\tau$ and random seeds on performance.}
    \label{fig:seed}
\end{figure}

To facilitate reproduction, in this section we describe in detail the implementation of \ours.
\ours\ is implemented on top of \textit{Soft Thinking}, using SGLang \citep{sglang} as the inference backend and NVIDIA H100 80GB GPUs.
As shown in Figure \ref{fig:seed}, performance varies primarily with the routing threshold $\tau$. Although different random seeds introduce stochastic fluctuations, the overall performance trend with respect to $\tau$ remains consistent.
According to this observation, we perform a grid search over $\tau$ to select the optimal routing threshold, and use multiple random seeds solely to perform repeated runs and average out randomness, reducing the impact of chance outcomes from a single run.
Three random seeds $\{0, 7, 42\}$ are used. 
For each model–dataset pair, we run inference three times with these three seeds, and report Pass@1 with $n = 3$ generations per sample.
All hyperparameters except for $\tau$ are provided in Appendix \ref{app:baseline} and the scripts in our public codebase after review.
To obtain the optimal $\tau$, we randomly sample 10 instances from each model-benchmark pair as a small validation set. 
We perform a grid search over $\tau \in \{0.4,0.5,0.6,0.7,0.8,0.9\}$ and compute Pass@1 on these 10 samples. 
The value of $\tau$ that achieves the highest Pass@1 and the shortest generation length is selected as the optimal $\tau$ for that model-benchmark pair.
Finally, we evaluate \ours\ on the remaining test samples of the benchmark on the model, excluding the 10 validation instances, and report Pass@1 and generation length in Table \ref{tab:math} and \ref{tab:code} ($\tau$ used in Table \ref{tab:math} and \ref{tab:code} are shown in Table \ref{tab:tau}.) 
This protocol ensures that $\tau$ selection does not leak test information while keeping the overall procedure lightweight and easy to reproduce.
The evaluation implementation follows \textit{Soft Thinking}, which uses Math-Verify\footnote{\url{https://github.com/huggingface/Math-Verify}} and GPT-4.1 for STEM reasoning, and the official code evaluations from the coding benchmarks.

\begin{table}[htbp]
  \centering
  \caption{$\tau$ used in Table \ref{tab:math}, \ref{tab:code}, \ref{tab:math_all}, and \ref{tab:code_all}.}
    \begin{tabular}{lccccc}
    \toprule
    \textbf{$\tau$} & \textbf{AIME 2024} & \textbf{AIME 2025} & \textbf{GPQA Diamond} & \textbf{HumanEval} & \textbf{MBPP} \\
    \midrule
    \textbf{Qwen3-1.7B} & 0.4   & 0.9   & 0.5   & 0.6   & 0.7 \\
    \textbf{Qwen3-8B} & 0.8   & 0.7   & 0.5   & 0.9   & 0.9 \\
    \textbf{Qwen3-32B} & 0.7   & 0.4   & 0.5   & 0.5   & 0.6 \\
    \textbf{gpt-oss-20b} &  0.5     &  0.9     &   0.9    &   0.4    & 0.4 \\
    \bottomrule
    \end{tabular}%
  \label{tab:tau}%
\end{table}%

\subsection{Main Results on Whole Datasets}
\label{app:fulldataset}
Table \ref{tab:math_all} and Table \ref{tab:code_all} report \ours's performance on the original all samples in each benchmark to ensure the robustness of our evaluation.
We observe that for STEM reasoning benchmarks, \ours\ outperforms all of the baselines on Pass@1, and has the shortest generation length in most cases.
For coding benchmarks, \ours\ has the highest Pass@1 accuracy in most cases, even when \textit{Soft Thinking} drops the accuracy compared with CoT (sampling).
Meanwhile, \ours\ exhibits competitive performance on reducing generation length with \textit{Soft Thinking}.
All the findings are the same as those concluded from Table \ref{tab:math} and \ref{tab:code} in  Section \ref{sec:result}.

\begin{table}[htbp]
  \centering
  \caption{Pass@1 (\%) and generation length on all data from STEM reasoning benchmarks for \ours\ and the baselines across
different models.}
\scalebox{0.7}{
    \begin{tabular}{l|cccccccc|ccccc}
    \toprule
          & \multicolumn{8}{c}{\textbf{Pass@1 (\%)}}                      & \multicolumn{5}{c}{\textbf{Generation Length}} \\
\cmidrule{2-14}          & \multicolumn{2}{c}{\textbf{AIME 2024}} & \multicolumn{2}{c}{\textbf{AIME 2025}} & \multicolumn{2}{c}{\textbf{GPQA Diamond}} & \multicolumn{2}{c}{\textbf{Average}} & \textbf{AIME 2024} & \textbf{AIME 2025} & \textbf{GPQA Diamond} & \multicolumn{2}{c}{\textbf{Average}} \\
    \midrule
          & \multicolumn{13}{c}{\textbf{Qwen3-1.7B}} \\
    \midrule
    CoT (sampling) & 45.56 &       & 33.33 &       & 38.98 &       & 39.29 &       & 17954.63 & 17684.26 & \textbf{8928.78} & 14855.89 &  \\
    CoT (greedy) & 58.89 & \textcolor{textblue}{$\uparrow$13.33} & 37.78 & \textcolor{textblue}{$\uparrow$4.44}  & 26.86 & \textcolor{textred}{$\downarrow$12.13} & 41.17 & \textcolor{textblue}{$\uparrow$1.88}  & 19101.63 & 19577.00 & 12877.72 & 17185.45 & \textcolor{textred}{$\uparrow$15.68\%} \\
    \textit{Soft Thinking} & 52.22 & \textcolor{textblue}{$\uparrow$6.67}  & 56.67 & \textcolor{textblue}{$\uparrow$23.33} & 43.29 & \textcolor{textblue}{$\uparrow$4.30}  & 50.72 & \textcolor{textblue}{$\uparrow$11.43} & 17516.83 & \textbf{16333.40} & 9300.22 & 14383.48 & \textcolor{textblue}{$\downarrow$3.18\%} \\
    Random Routing & 55.56 & \textcolor{textblue}{$\uparrow$10.00} & 46.67 & \textcolor{textblue}{$\uparrow$13.33} & 44.00 & \textcolor{textblue}{$\uparrow$5.01}  & 48.74 & \textcolor{textblue}{$\uparrow$9.45}  & 17255.43 & 18237.87 & 9331.43 & 14941.58 & \textcolor{textred}{$\uparrow$0.58\%} \\
    \textbf{\ours} & \textbf{67.78} & \textcolor{textblue}{\textbf{$\uparrow$22.22}} & \textbf{62.22} & \textcolor{textblue}{\textbf{$\uparrow$28.89}} & \textbf{49.50} & \textcolor{textblue}{\textbf{$\uparrow$10.52}} & \textbf{59.83} & \textcolor{textblue}{\textbf{$\uparrow$20.54}} & \textbf{16222.60} & 16576.42 & 9196.06 & \textbf{13998.36} & \textcolor{textblue}{\textbf{$\downarrow$5.77\%}} \\
    \midrule
          & \multicolumn{13}{c}{\textbf{Qwen3-8B}} \\
    \midrule
    CoT (sampling) & 75.56 &       & 74.44 &       & 62.70 &       & 70.90 &       & 14686.38 & 18062.92 & 9007.57 & 13918.96 &  \\
    CoT (greedy) & 81.11 & \textcolor{textblue}{$\uparrow$5.56}  & 81.11 & \textcolor{textblue}{$\uparrow$6.67}  & 63.76 & \textcolor{textblue}{$\uparrow$1.06}  & 75.33 & \textcolor{textblue}{$\uparrow$4.43}  & 16443.03 & 17630.60 & 11508.81 & 15194.15 & \textcolor{textred}{$\uparrow$9.16\%} \\
    \textit{Soft Thinking} & 84.44 & \textcolor{textblue}{$\uparrow$8.89}  & 80.00 & \textcolor{textblue}{$\uparrow$5.56}  & 68.63 & \textcolor{textblue}{$\uparrow$5.93}  & 77.69 & \textcolor{textblue}{$\uparrow$6.79}  & \textbf{13823.70} & \textbf{16973.40} & 9015.03 & \textbf{13270.71} & \textcolor{textblue}{$\downarrow$\textbf{4.66\%}} \\
    Random Routing & 87.78 & \textcolor{textblue}{$\uparrow$12.22} & 83.33 & \textcolor{textblue}{$\uparrow$8.89}  & 69.53 & \textcolor{textblue}{$\uparrow$6.83}  & 80.21 & \textcolor{textblue}{$\uparrow$9.31}  & 15131.58 & 17674.14 & 9080.94 & 13962.22 & \textcolor{textred}{$\uparrow$0.31\%} \\
    \textbf{\ours} & \textbf{88.89} & \textcolor{textblue}{$\uparrow$\textbf{13.33}} & \textbf{83.33} & \textcolor{textblue}{$\uparrow$\textbf{8.89}} & \textbf{82.10} & \textcolor{textblue}{$\uparrow$\textbf{19.40}} & \textbf{84.77} & \textcolor{textblue}{$\uparrow$\textbf{13.88}} & 14406.19 & 17497.14 & \textbf{8128.78} & 13344.04 & \textcolor{textblue}{$\downarrow$4.13\%} \\
    \midrule
          & \multicolumn{13}{c}{\textbf{Qwen3-32B}} \\
    \midrule    
    CoT (sampling) & 78.89 &       & 80.00 &       & 75.56 &       & 78.15 &       & 12528.44 & 15532.58 & 5546.56 & 11202.53 &  \\
    CoT (greedy) & 83.33 & \textcolor{textblue}{$\uparrow$4.44}  & 77.78 & \textcolor{textred}{$\downarrow$2.22} & 70.52 & \textcolor{textred}{$\downarrow$5.04} & 77.21 & \textcolor{textred}{$\downarrow$0.94} & 12681.27 & 14135.13 & 8127.01 & 11647.80 & \textcolor{textred}{$\uparrow$3.97\%} \\
    \textit{Soft Thinking} & 91.11 & \textcolor{textblue}{$\uparrow$12.22} & 83.33 & \textcolor{textblue}{$\uparrow$3.33}  & 75.25 & \textcolor{textred}{$\downarrow$0.31} & 83.23 & \textcolor{textblue}{$\uparrow$5.08}  & 12882.30 & 15011.17 & \textbf{5278.91} & 11057.46 & \textcolor{textblue}{$\downarrow$1.29\%} \\

    Random Routing & 88.89 & \textcolor{textblue}{$\uparrow$10.00} & 85.56 & \textcolor{textblue}{$\uparrow$5.56}  & 80.12 & \textcolor{textblue}{$\uparrow$4.56}  & 84.85 & \textcolor{textblue}{$\uparrow$6.71}  & \textbf{11704.17} & 15350.50 & 6174.77 & 11076.48 & \textcolor{textblue}{$\downarrow$1.13\%} \\ 
    \textbf{\ours} & \textbf{92.22} & \textcolor{textblue}{$\uparrow$\textbf{13.33}} & \textbf{92.22} & \textcolor{textblue}{$\uparrow$\textbf{12.22}} & \textbf{82.10} & \textcolor{textblue}{$\uparrow$\textbf{6.55}} & \textbf{88.85} & \textcolor{textblue}{$\uparrow$\textbf{10.70}} & 12291.54 & \textbf{11994.26} & 5475.92 & \textbf{9920.57} & \textcolor{textblue}{\textbf{$\downarrow$11.44\%}} \\
    \midrule
          & \multicolumn{13}{c}{\textbf{gpt-oss-20b}} \\
    \midrule
    CoT (sampling) & 82.22 &       & 80.00 &       & 72.79 &       & 78.34 &       & 9718.70 & 12224.48 & 3471.98 & 8471.72 &  \\
    CoT (greedy) & 76.67 & \textcolor{textred}{$\downarrow$5.56} & 75.56 & \textcolor{textred}{$\downarrow$4.44} & 71.23 & \textcolor{textred}{$\downarrow$1.56} & 74.48 & \textcolor{textred}{$\downarrow$3.85} & 12557.50 & \textbf{5215.20} & 6151.36 & 7974.69 & \textcolor{textred}{$\uparrow$5.87\%} \\
    \textit{Soft Thinking} & 76.67 & \textcolor{textred}{$\downarrow$5.56} & 73.33 & \textcolor{textred}{$\downarrow$6.67} & 72.39 & \textcolor{textred}{$\downarrow$0.40} & 74.13 & \textcolor{textred}{$\downarrow$4.21} & \textbf{5592.63} & 15671.90 & \textbf{2557.33} & 7940.62 & \textcolor{textblue}{$\downarrow$6.27\%} \\
    Random Routing & 93.33 & \textcolor{textblue}{$\uparrow$11.11} & 84.44 & \textcolor{textblue}{$\uparrow$4.44}  & 73.50 & \textcolor{textblue}{$\uparrow$0.71}  & 83.76 & \textcolor{textblue}{$\uparrow$5.42}  & 6588.63 & 9076.46 & 2571.38 & \textbf{6078.82} & \textcolor{textblue}{$\downarrow$\textbf{28.25\%}} \\
    \textbf{\ours} & \textbf{94.44} & \textcolor{textblue}{$\uparrow$\textbf{12.22}} & \textbf{92.22} & \textcolor{textblue}{$\uparrow$\textbf{12.22}} & \textbf{79.98} & \textcolor{textblue}{$\uparrow$\textbf{7.19}} & \textbf{88.88} & \textcolor{textblue}{$\uparrow$\textbf{10.54}} & 8269.57 & 11308.31 & 3288.05 & 7621.98 & \textcolor{textblue}{$\downarrow$10.03\%} \\

    \bottomrule
    \end{tabular}}
  \label{tab:math_all}%
\end{table}%

\begin{table}[htbp]
  \centering
  \caption{Pass@1 (\%) and generation length on all data from coding benchmarks for \ours and the baselines across different
models.}
\scalebox{0.78}{
    \begin{tabular}{l|cccccc|cccc}
    \toprule
          & \multicolumn{6}{c}{\textbf{Pass@1 (\%)}}           & \multicolumn{4}{c}{\textbf{Generation Length}} \\
\cmidrule{2-11}          & \multicolumn{2}{c}{\textbf{HumanEval}} & \multicolumn{2}{c}{\textbf{MBPP}} & \multicolumn{2}{c}{\textbf{Average}} & \textbf{HumanEval} & \textbf{MBPP} & \multicolumn{2}{c}{\textbf{Average}} \\
    \midrule
          & \multicolumn{10}{c}{\textbf{Qwen3-1.7B}} \\
    \midrule
    CoT (sampling) & 81.27 &       & 79.48 &       & 80.38 &       & 3853.75 & 3716.10 & 3784.93 &  \\
    CoT (greedy) & 81.82 & \textcolor{textblue}{$\uparrow$0.55}  & 70.84 & \textcolor{textred}{$\downarrow$8.65} & 76.33 & \textcolor{textred}{$\downarrow$4.05} & 4917.32 & 5727.78 & 5322.55 & \textcolor{textred}{$\uparrow$40.62\%} \\
    \textit{Soft Thinking} & 81.95 & \textcolor{textblue}{$\uparrow$0.68}  & 71.92 & \textcolor{textred}{$\downarrow$7.57} & 76.93 & \textcolor{textred}{$\downarrow$3.44} & \textbf{3045.71} & 3889.74 & \textbf{3467.72} & \textcolor{textblue}{\textbf{$\downarrow$8.38\%}} \\
    Random Routing & 83.06 & \textcolor{textblue}{$\uparrow$1.79}  & 71.38 & \textcolor{textred}{$\downarrow$8.11} & 77.22 & \textcolor{textred}{$\downarrow$3.16} & 3548.99 & 3799.67 & 3674.33 & \textcolor{textblue}{$\downarrow$2.92\%} \\
    \textbf{\ours} & \textbf{84.55} & \textcolor{textblue}{\textbf{$\uparrow$3.28}} & \textbf{79.51} & \textcolor{textblue}{\textbf{$\uparrow$0.03}} & \textbf{82.03} & \textcolor{textblue}{\textbf{$\uparrow$1.65}} & 3531.70 & \textbf{3652.33} & 3592.01 & \textcolor{textblue}{$\downarrow$0.05\%} \\
    \midrule
          & \multicolumn{10}{c}{\textbf{Qwen3-8B}} \\
    \midrule
    CoT (sampling) & 76.35 &       & 96.04 &       & 86.20 &       & 3951.80 & 2986.37 & 3469.08 &  \\
    CoT (greedy) & 74.29 & \textcolor{textred}{$\downarrow$2.06} & 92.71 & \textcolor{textred}{$\downarrow$3.33} & 83.50 & \textcolor{textred}{$\downarrow$2.70} & 5559.28 & 3994.14 & 4776.71 & \textcolor{textred}{$\uparrow$37.69\%} \\
    \textit{Soft Thinking} & 81.82 & \textcolor{textblue}{$\uparrow$5.47}  & 94.33 & \textcolor{textred}{$\downarrow$1.71} & 88.08 & \textcolor{textblue}{$\uparrow$1.88}  & \textbf{3029.38} & \textbf{2609.40} & \textbf{2819.39} & \textcolor{textblue}{\textbf{$\downarrow$18.73\%}} \\

    Random Routing & 81.95 & \textcolor{textblue}{$\uparrow$5.60}  & 96.30 & \textcolor{textblue}{$\uparrow$0.26}  & 89.12 & \textcolor{textblue}{$\uparrow$2.93}  & 3480.59 & 2917.61 & 3199.10 & \textcolor{textblue}{$\downarrow$7.78\%} \\

    \textbf{\ours} & \textbf{82.07} & \textcolor{textblue}{\textbf{$\uparrow$5.72}} & \textbf{96.31} & \textcolor{textblue}{\textbf{$\uparrow$0.27}} & \textbf{89.19} & \textcolor{textblue}{\textbf{$\uparrow$2.99}} & 3445.92 & 2840.10 & 3143.01 & \textcolor{textblue}{$\downarrow$9.40\%} \\
    \midrule
          & \multicolumn{10}{c}{\textbf{Qwen3-32B}} \\
    \midrule
    CoT (sampling) & 71.54 &       & 96.76 &       & 84.15 &       & 2998.65 & 2291.46 & 2645.06 &  \\
    CoT (greedy) & \textbf{78.05} & \textcolor{textblue}{\textbf{$\uparrow$6.51}} & 97.03 & \textcolor{textblue}{$\uparrow$0.27}  & \textbf{87.54} & \textcolor{textblue}{\textbf{$\uparrow$3.39}} & 3046.14 & 2342.95 & 2694.55 & \textcolor{textred}{$\uparrow$1.87\%} \\
    \textit{Soft Thinking} & 66.32 & \textcolor{textred}{$\downarrow$5.22} & 97.57 & \textcolor{textred}{$\uparrow$0.81}  & 81.95 & \textcolor{textred}{$\downarrow$2.21} & \textbf{2922.66} & \textbf{2178.35} & \textbf{2550.50} & \textcolor{textblue}{\textbf{$\downarrow$3.57\%}} \\
    Random Routing & 69.94 & \textcolor{textred}{$\downarrow$1.60} & 97.57 & \textcolor{textblue}{0.81}  & 83.76 & \textcolor{textred}{$\downarrow$0.40} & 3062.56 & 2313.64 & 2688.10 & \textcolor{textred}{$\uparrow$1.63\%} \\
    \textbf{\ours} & 75.15 & \textcolor{textblue}{$\uparrow$3.61}  & \textbf{97.66} & \textcolor{textblue}{\textbf{$\uparrow$0.90}} & 86.41 & \textcolor{textblue}{$\uparrow$2.26}  & 2944.79 & 2287.75 & 2616.27 & \textcolor{textblue}{$\downarrow$1.09\%} \\
    \midrule
          & \multicolumn{10}{c}{\textbf{gpt-oss-20b}} \\
    \midrule

    CoT (sampling) & 86.15 &       & 98.11 &       & 92.13 &       & 842.35 & 739.92 & 791.13 &  \\
    CoT (greedy) & 81.82 & \textcolor{textred}{$\downarrow$4.33} & 97.30 & \textcolor{textred}{$\downarrow$0.81} & 89.56 & \textcolor{textred}{$\downarrow$2.57} & 1481.55 & 1069.33 & 1275.44 & \textcolor{textred}{$\uparrow$61.22\%} \\
    \textit{Soft Thinking} & 86.15 & 0.00  & 96.22 & \textcolor{textred}{$\downarrow$1.89} & 91.18 & \textcolor{textred}{$\downarrow$0.94} & \textbf{842.35} & 659.69 & \textbf{751.02} & \textcolor{textblue}{$\downarrow$5.07\%} \\
    Random Routing & 80.59 & \textcolor{textred}{$\downarrow$5.56} & 97.48 & \textcolor{textred}{$\downarrow$0.63} & 89.04 & \textcolor{textred}{$\downarrow$3.09} & 852.17 & \textbf{595.03} & \textbf{723.60} & \textcolor{textblue}{$\downarrow$\textbf{8.54\%}} \\
    \textbf{\ours} & \textbf{86.29} & \textcolor{textblue}{$\uparrow$\textbf{0.14}} & \textbf{98.83} & \textcolor{textblue}{$\uparrow$\textbf{0.72}} & \textbf{92.56} & \textcolor{textblue}{$\uparrow$\textbf{0.43}} & \textbf{859.45} & \textbf{648.43} & 753.94 & \textcolor{textblue}{$\downarrow$4.70\%} \\
    \bottomrule
    \end{tabular}}
  \label{tab:code_all}%
\end{table}%

\section{Further Analysis}
We follow the same setting in \S \ref{sec:experi_setup}.
For each LRM, we run each sample in each dataset three times. 
We record the next-token distribution at every time step for all three runs, and use the collected next-token distributions across all runs to construct all figures in this section.

\subsection{Low-confidence Steps as Thinking Progresses}
\label{app:low-confidence}

\begin{figure*}[ht]
    \centering
    \begin{subfigure}[b]{0.47\textwidth}
        \centering
        \includegraphics[width=\textwidth]{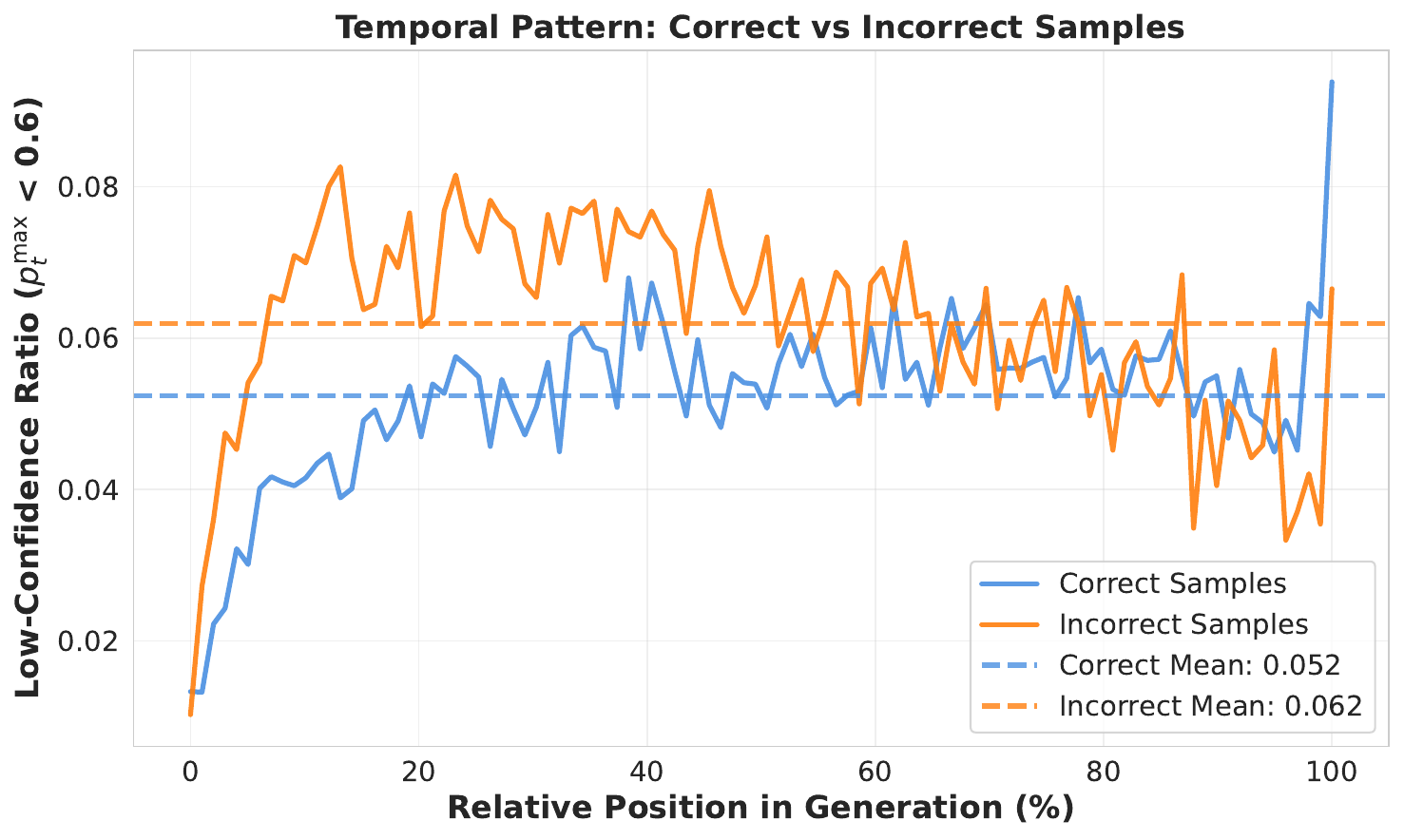}
        \caption{Latent-only Reasoning (Soft Thinking).}
    \end{subfigure}
    \begin{subfigure}[b]{0.47\textwidth}
        \centering
        \includegraphics[width=\textwidth]{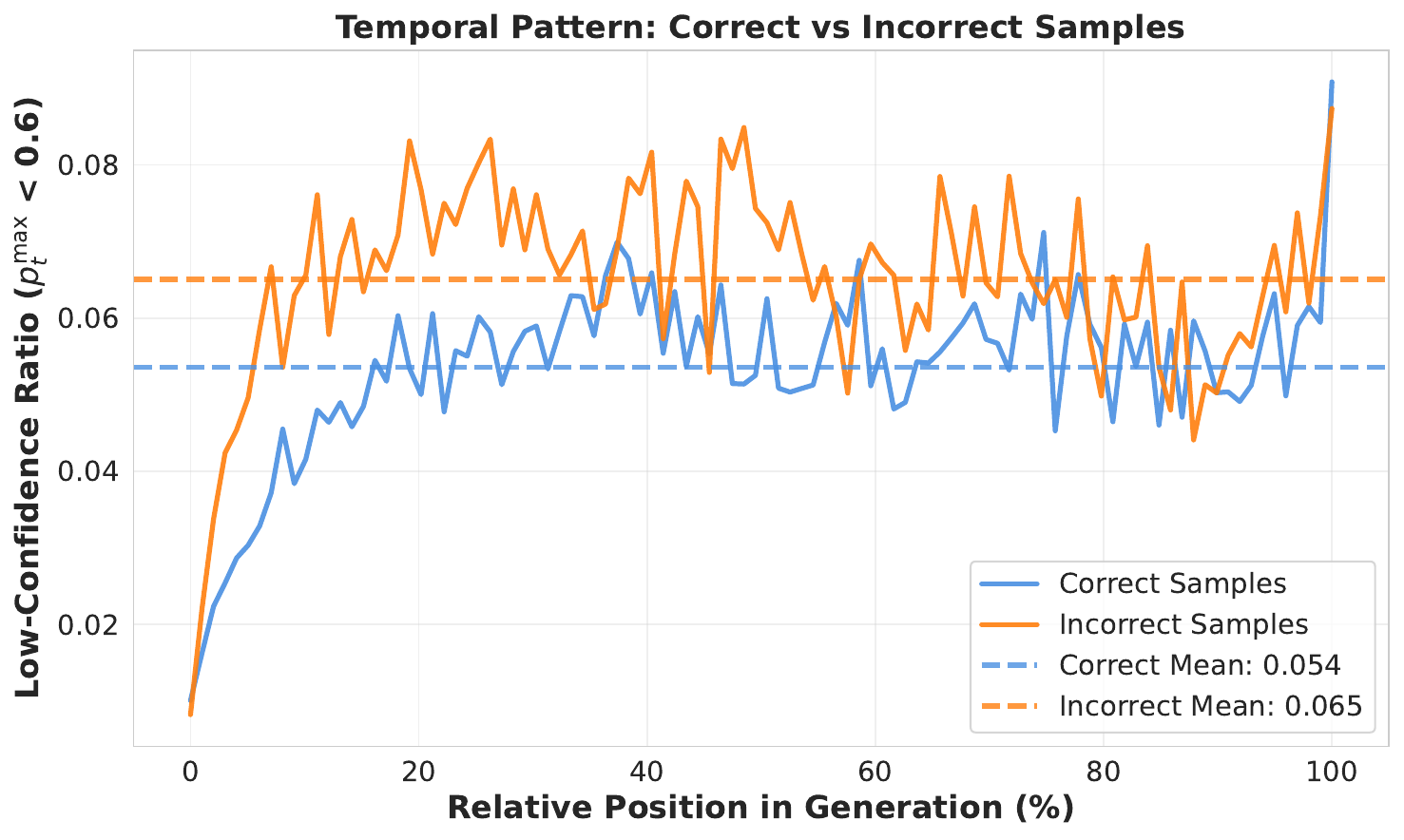}
        \caption{\ours.}
    \end{subfigure}
    \caption{Low-confidence time step ratio (\%) across generation steps on Qwen3-8B with HumanEval.}
    \label{fig:temporal_pattern2}
\end{figure*}

\begin{figure*}[hb]
    \centering
    \begin{subfigure}[b]{0.47\textwidth}
        \centering
        \includegraphics[width=\textwidth]{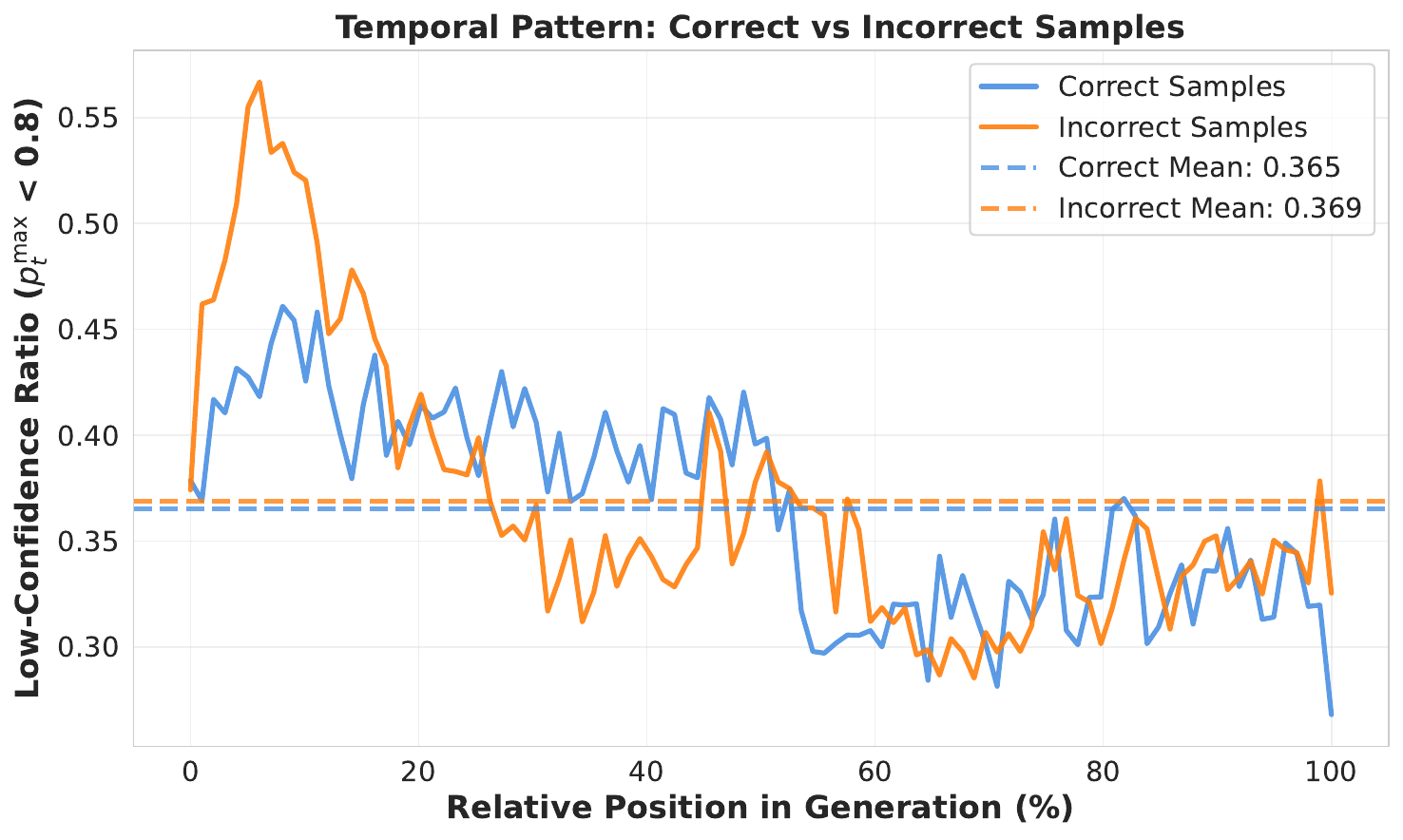}
        \caption{Latent-only Reasoning (Soft Thinking).}
    \end{subfigure}
    \begin{subfigure}[b]{0.47\textwidth}
        \centering
        \includegraphics[width=\textwidth]{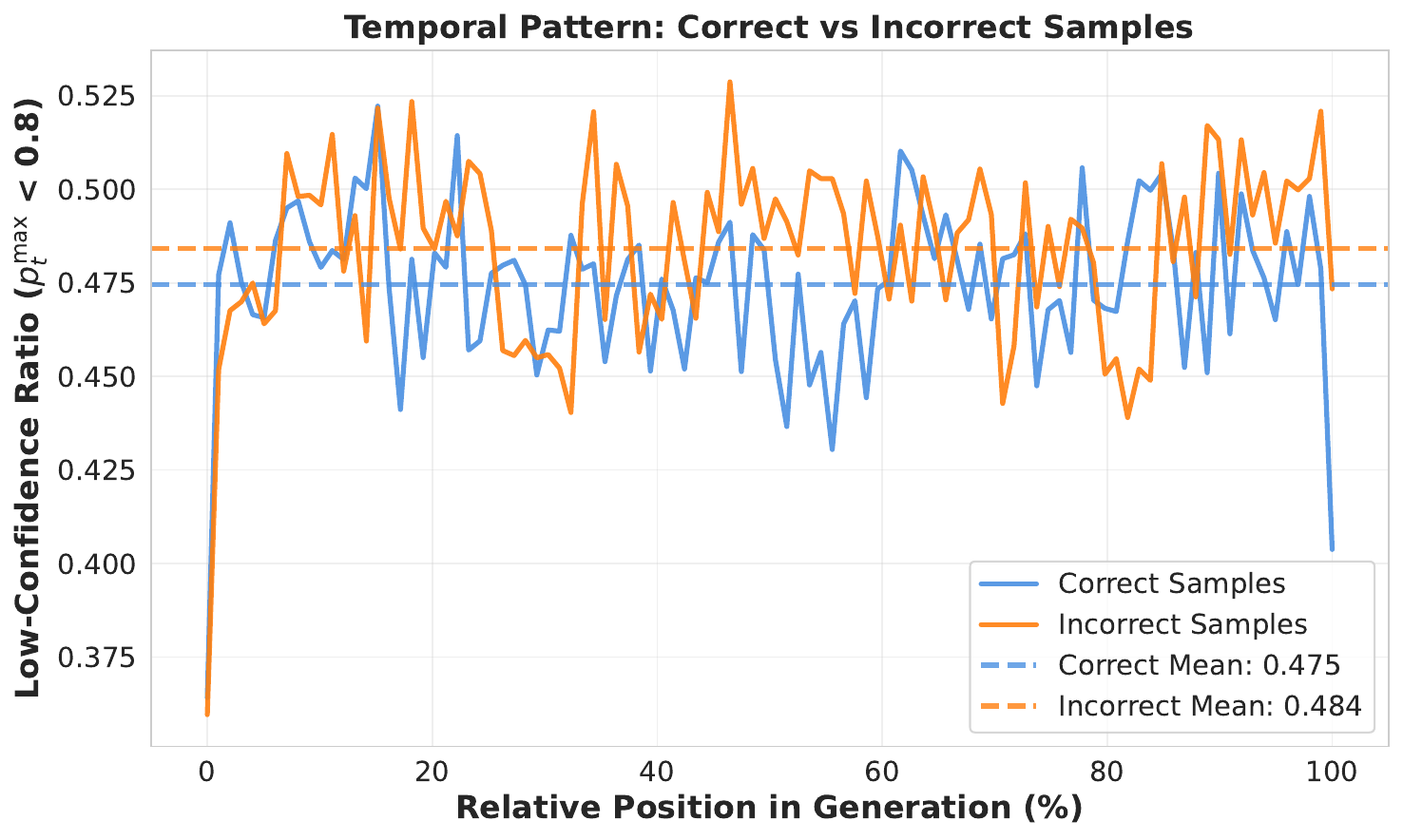}
        \caption{\ours.}
    \end{subfigure}
    \caption{Low-confidence time step ratio (\%) across generation steps on gpt-oss-20b with GPQA Diamond.}
    \label{fig:temporal_pattern3}
\end{figure*}

\begin{figure*}[ht]
    \centering
    \begin{subfigure}[b]{0.47\textwidth}
        \centering
        \includegraphics[width=\textwidth]{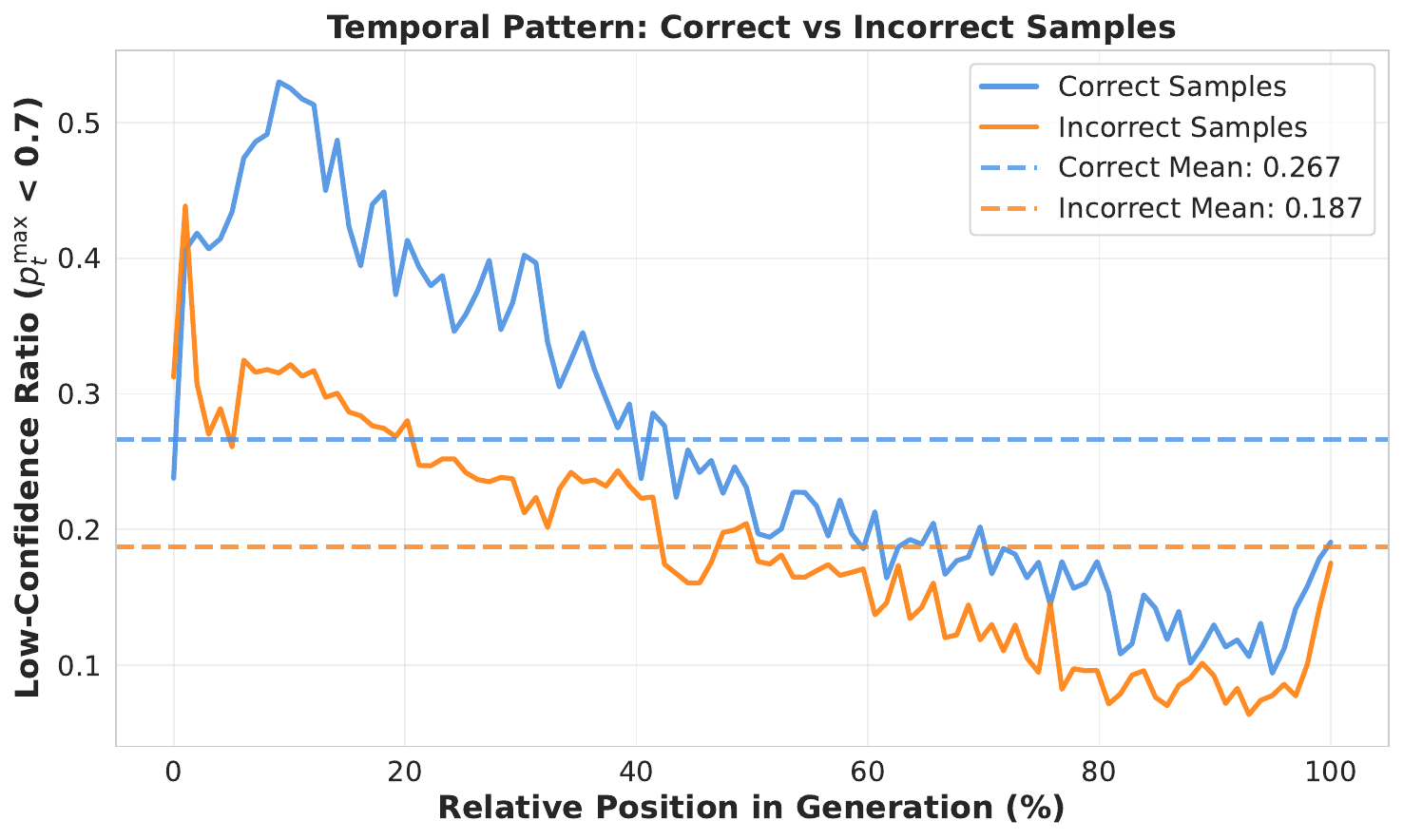}
        \caption{Latent-only Reasoning (Soft Thinking).}
    \end{subfigure}
    \begin{subfigure}[b]{0.47\textwidth}
        \centering
        \includegraphics[width=\textwidth]{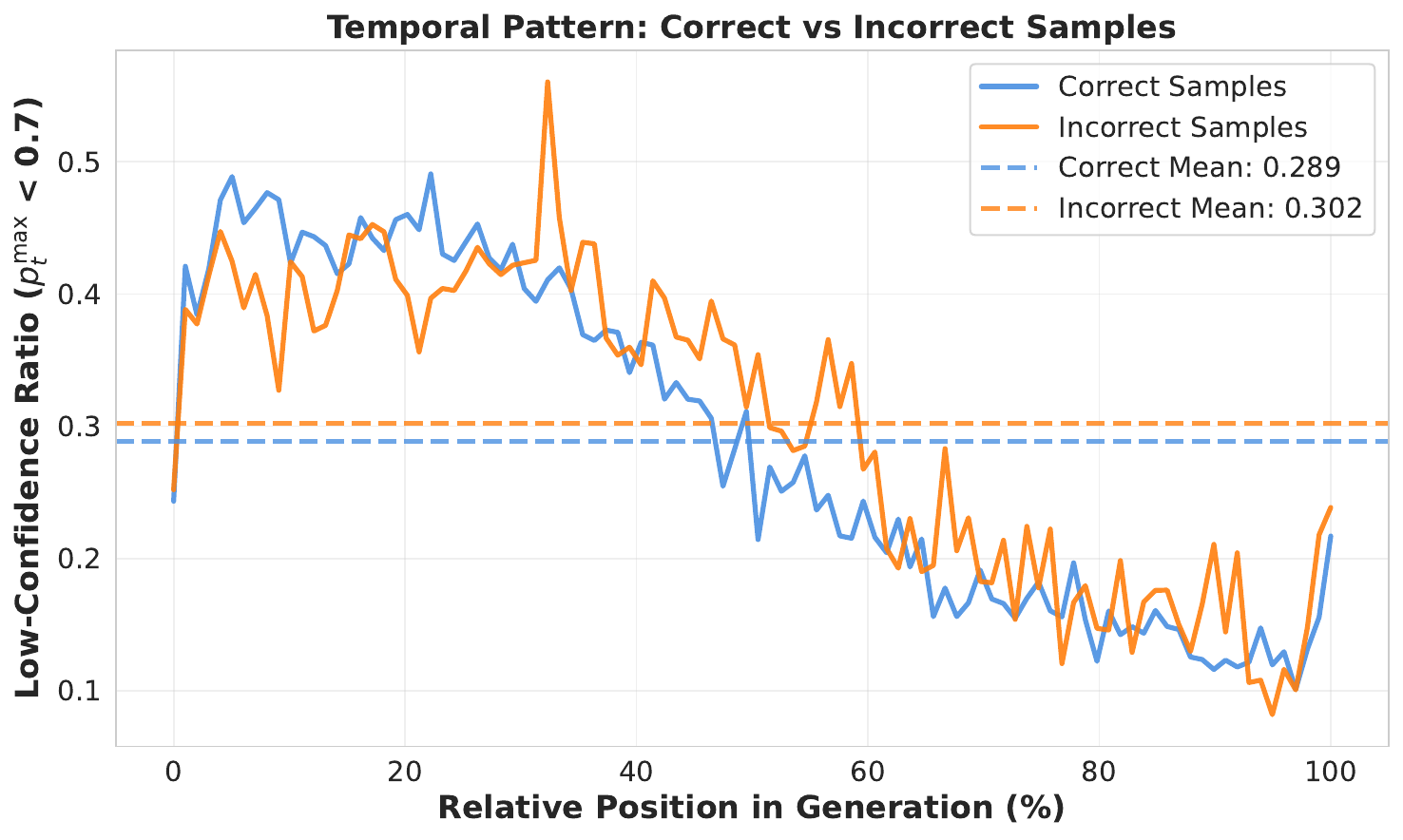}
        \caption{\ours.}
    \end{subfigure}
    \caption{Low-confidence time step ratio (\%) across generation steps on gpt-oss-20b with HumanEval.}
    \label{fig:temporal_pattern4}
\end{figure*}

Since different samples have varying generation lengths, we normalize each sample's reasoning trajectory to relative positions and discretize it into 
100 bins.
For each bin $b$, we collect all time steps whose relative positions fall within it and calculate the ratio of low-confidence time steps for both correct $\frac{1}{|S_{\text{correct}}^{(b)}|} \sum\limits_{t \in S_{\text{correct}}^{(b)}} \mathbb{I}[p_t^{\max} < \tau]$ and incorrect samples $\frac{1}{|S_{\text{incorrect}}^{(b)}|} \sum\limits_{t \in S_{\text{incorrect}}^{(b)}} \mathbb{I}[p_t^{\max} < \tau]$, where $S_{\text{incorrect}}^{(b)}$ and $S_{\text{incorrect}}^{(b)}$ denote the sets of time steps from correct and incorrect samples falling into bin $b$, respectively.
Figure \ref{fig:temporal_pattern1}, \ref{fig:temporal_pattern2}, \ref{fig:temporal_pattern3}, and \ref{fig:temporal_pattern4} show the ratios of low-confidence time steps for Qwen3-8B and gpt-oss-20b evaluated on GPQA Diamond and HumanEval.

\subsection{Thinking Stops}
\begin{figure}
    \centering
    \begin{subfigure}[b]{0.32\textwidth}
        \centering
        \includegraphics[width=\textwidth]{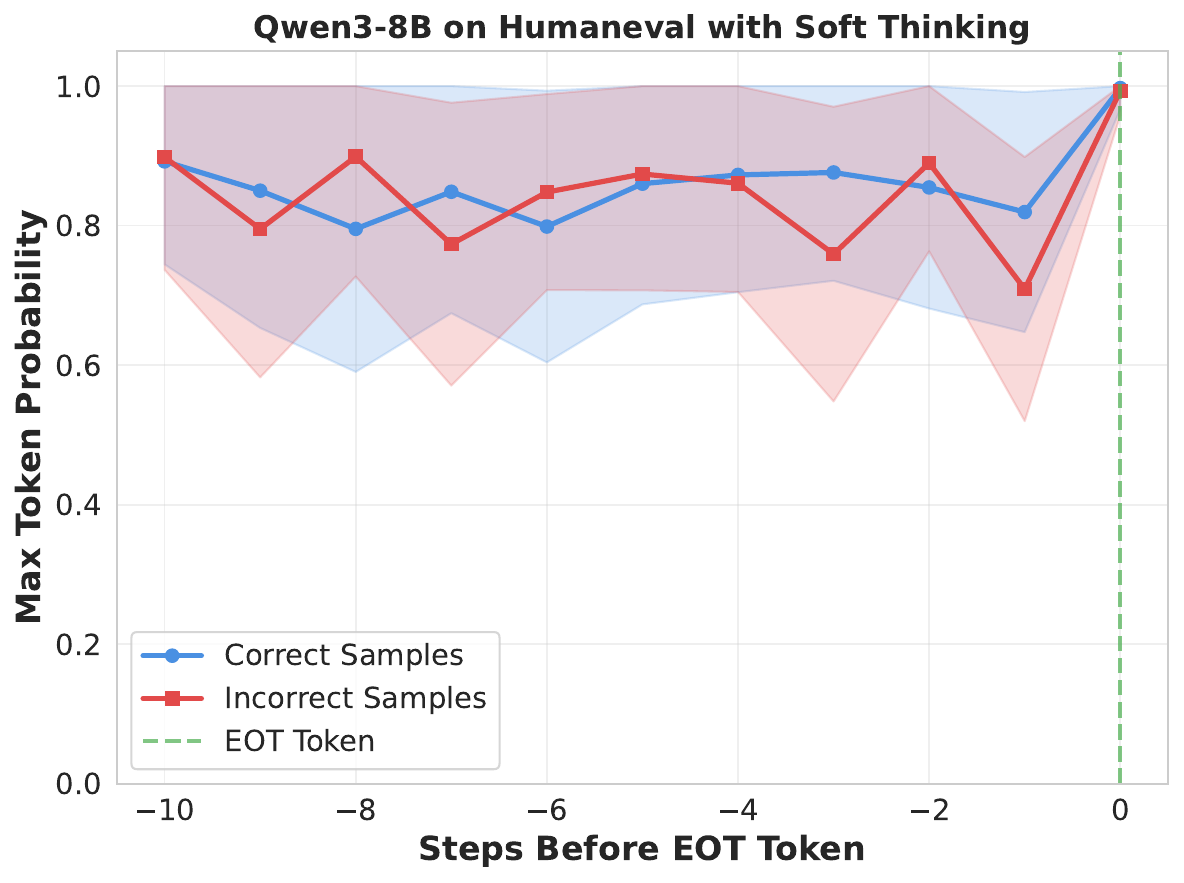}
    \end{subfigure}
    \hspace{0.01\textwidth}
    \begin{subfigure}[b]{0.32\textwidth}
        \centering
        \includegraphics[width=\textwidth]{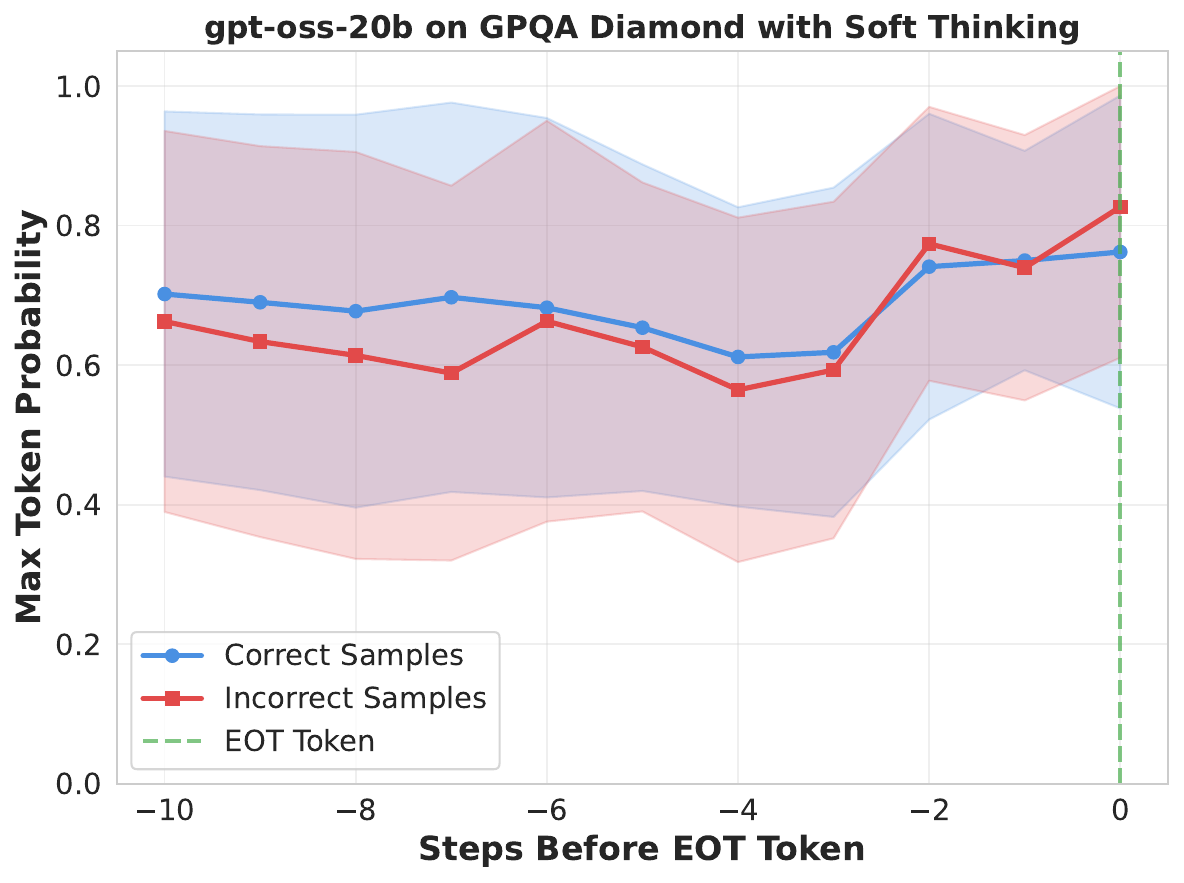}
    \end{subfigure}
    \hspace{0.01\textwidth}
    \begin{subfigure}[b]{0.32\textwidth}
        \centering
        \includegraphics[width=\textwidth]{figs/thinking_length/eot_token_evolution_st_gptoss_gpqa.pdf}
    \end{subfigure}
    \begin{subfigure}[b]{0.345\textwidth}
        \centering
        \includegraphics[width=\textwidth]{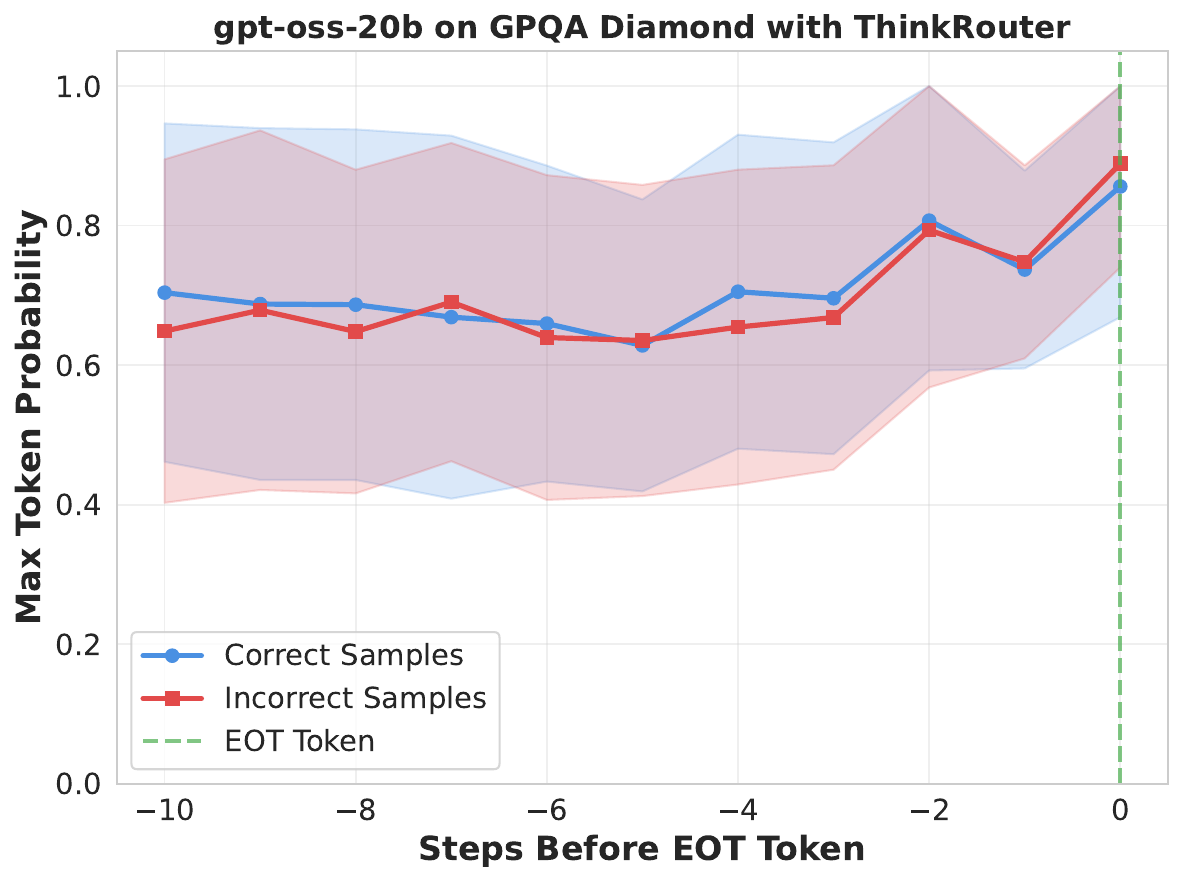}
    \end{subfigure}
\hspace{0.01\textwidth}
    \begin{subfigure}[b]{0.345\textwidth}
        \centering
        \includegraphics[width=\textwidth]{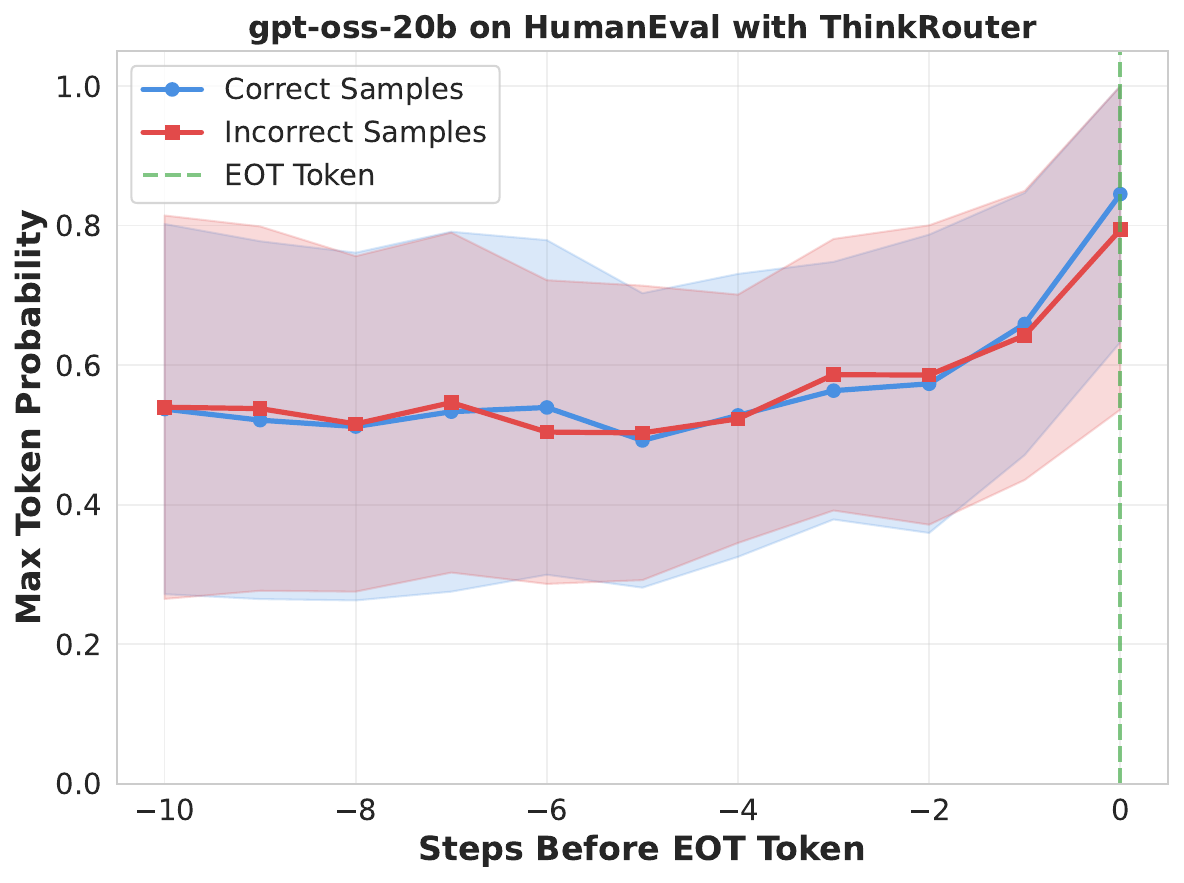}
    \end{subfigure}
    \caption{$p_t^{\max}$ of last 10 time steps before the end-of-thinking token.}
    \label{fig:eot2}
\end{figure}

\begin{figure}
    \centering
    \begin{subfigure}[b]{0.45\textwidth}
        \centering
        \includegraphics[width=\textwidth]{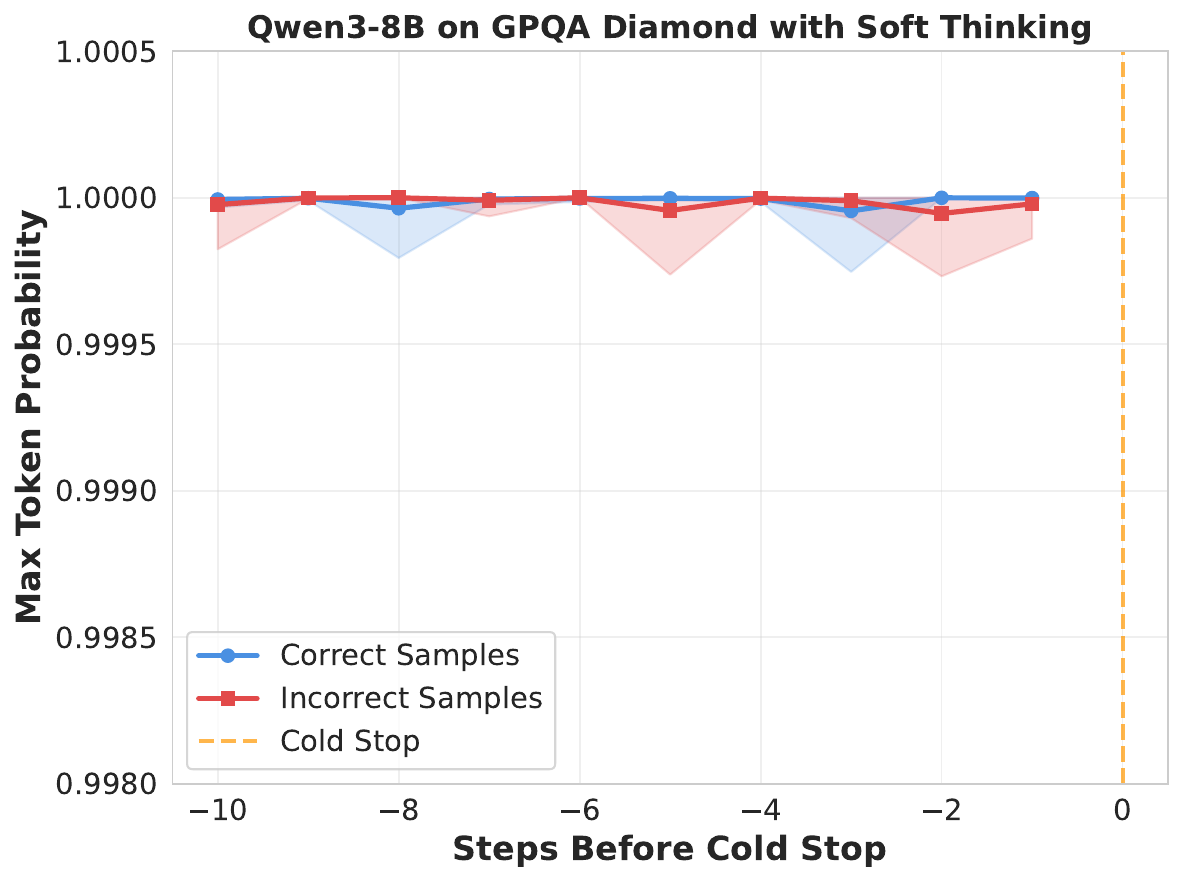}
    \end{subfigure}
    \hspace{0.01\textwidth}
    \begin{subfigure}[b]{0.45\textwidth}
        \centering
        \includegraphics[width=\textwidth]{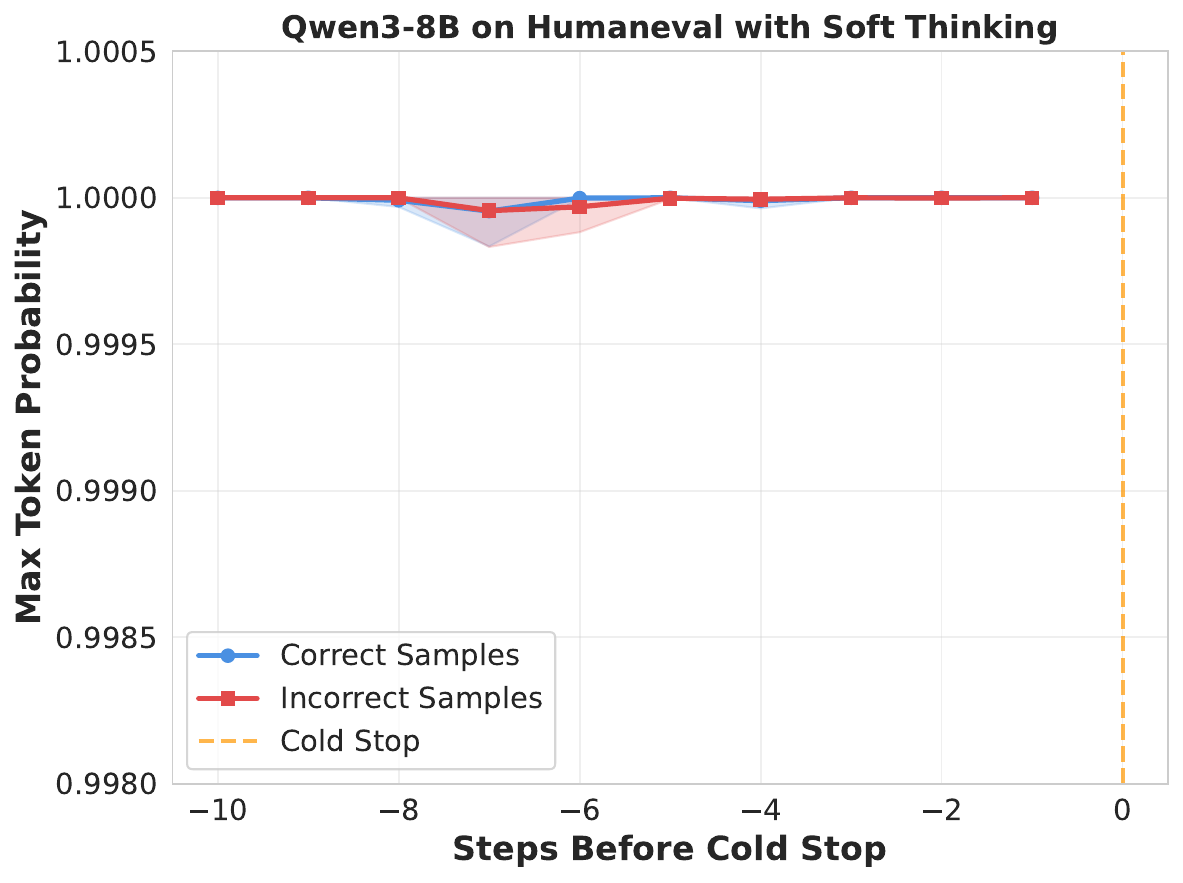}
    \end{subfigure}
    \hspace{0.01\textwidth}
    \begin{subfigure}[b]{0.45\textwidth}
        \centering
        \includegraphics[width=\textwidth]{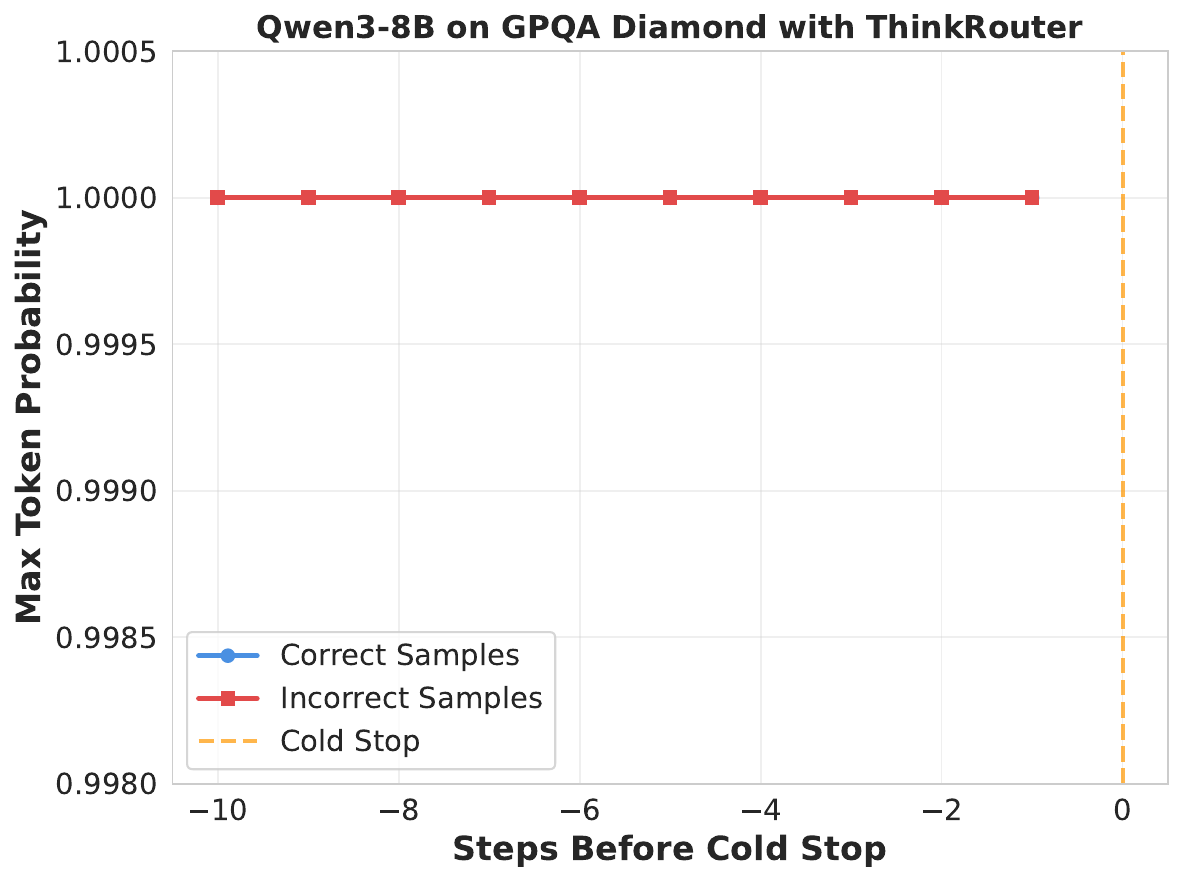}
    \end{subfigure}
    \begin{subfigure}[b]{0.45\textwidth}
        \centering
        \includegraphics[width=\textwidth]{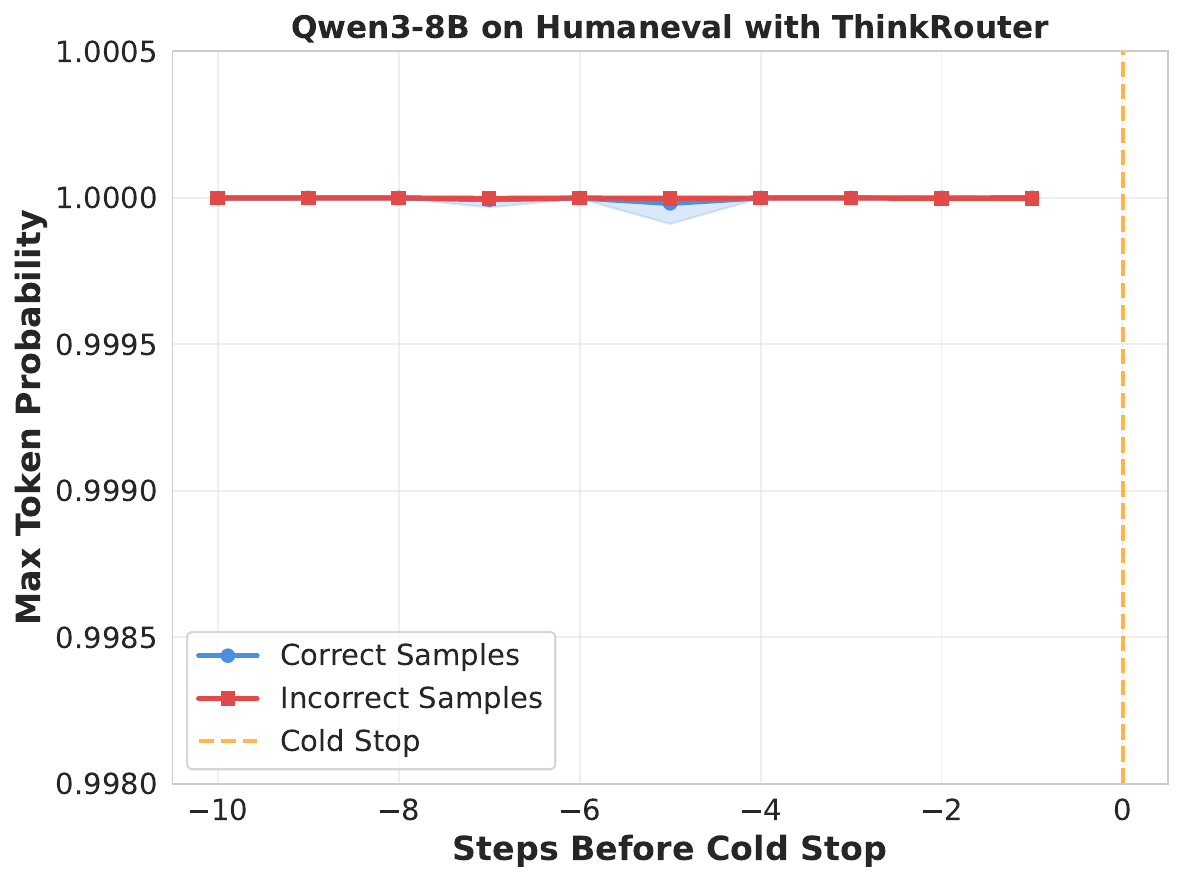}
    \end{subfigure}
    \caption{$p_t^{\max}$ of last 10 time steps before Cold Stop for Qwen3-8B.}
    \label{fig:codestop_qwen3}
\end{figure}

\begin{figure}
    \centering
    \begin{subfigure}[b]{0.45\textwidth}
        \centering
        \includegraphics[width=\textwidth]{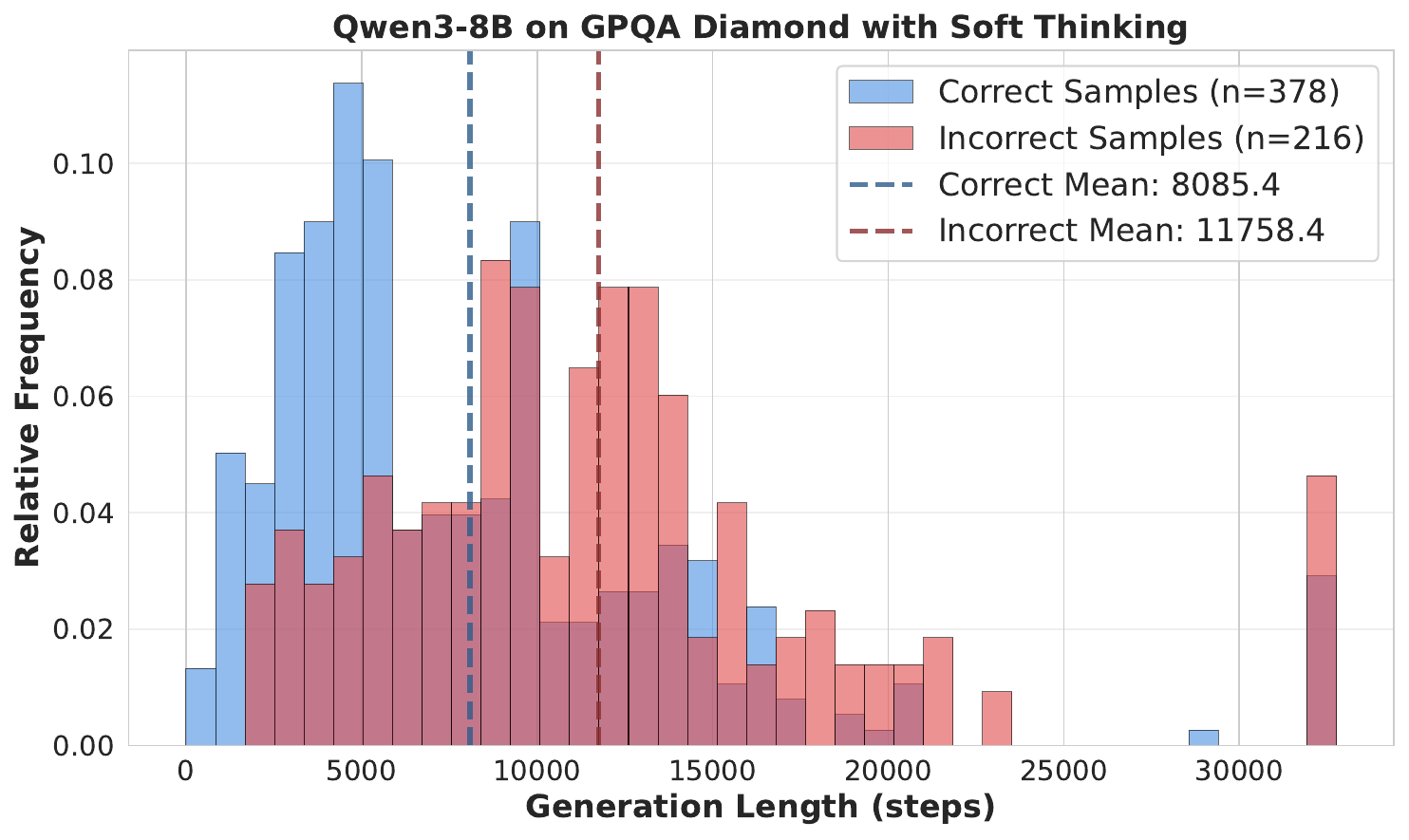}
    \end{subfigure}
    \hspace{0.01\textwidth}
    \begin{subfigure}[b]{0.45\textwidth}
        \centering
        \includegraphics[width=\textwidth]{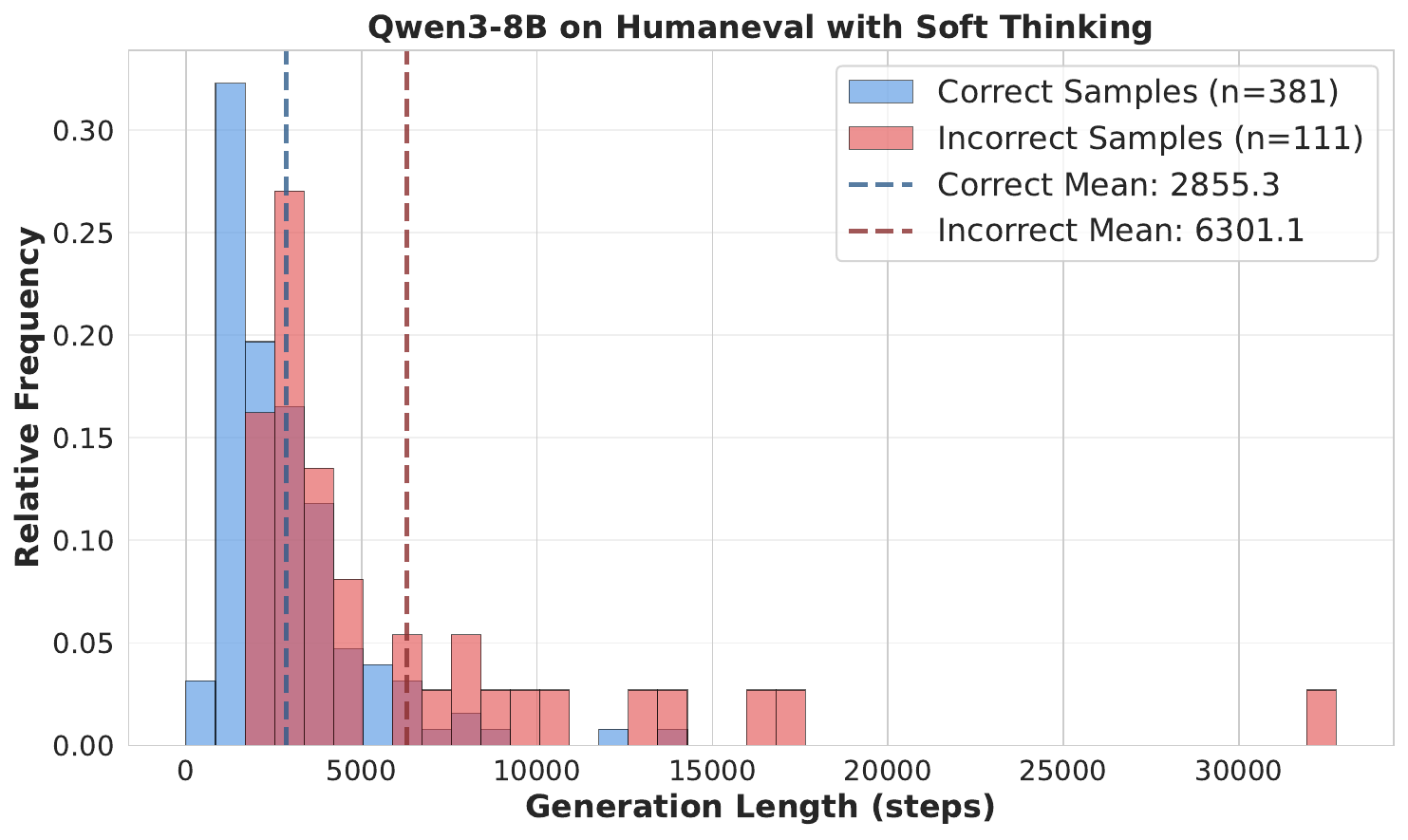}
    \end{subfigure}
    \hspace{0.01\textwidth}
    \begin{subfigure}[b]{0.45\textwidth}
        \centering
        \includegraphics[width=\textwidth]{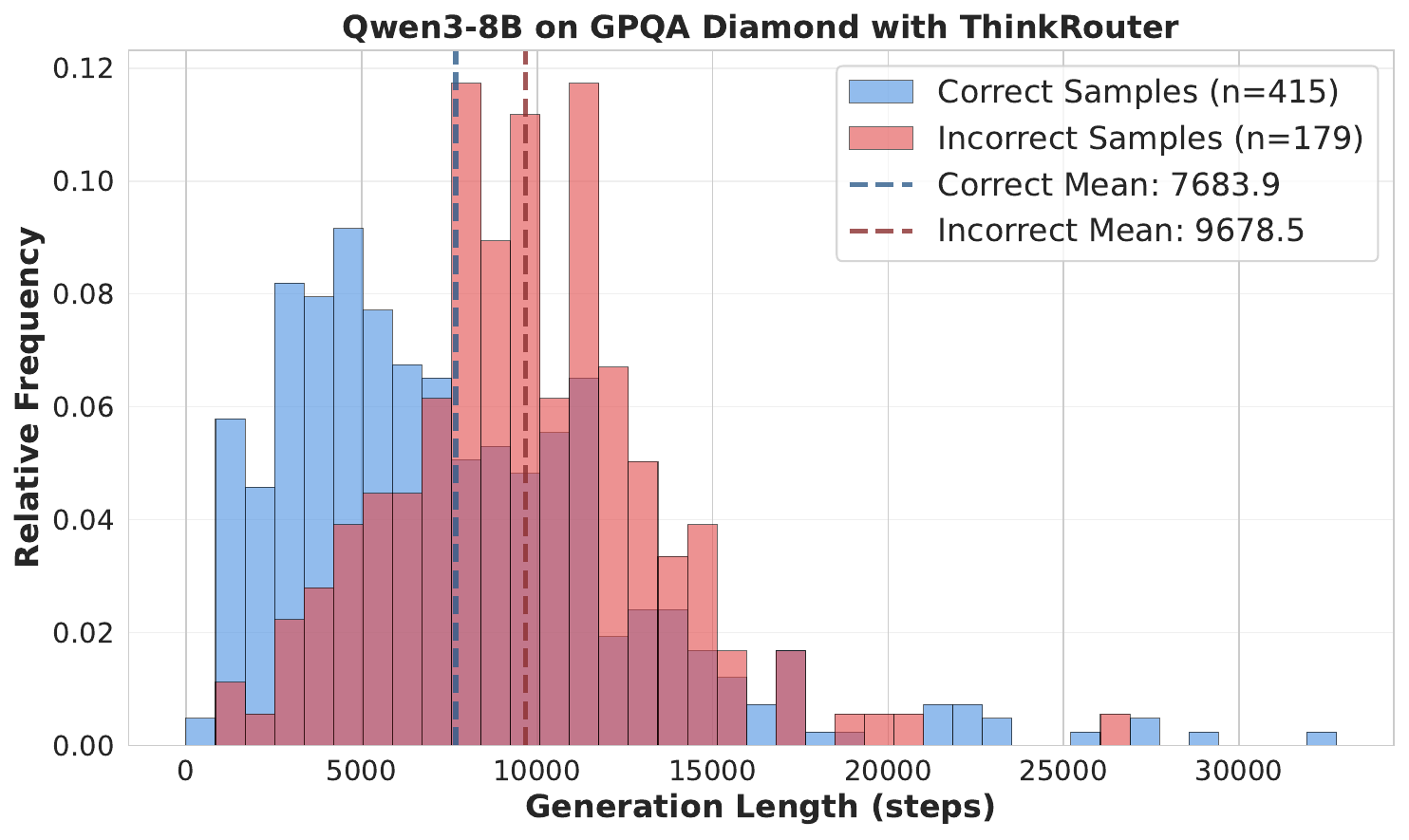}
    \end{subfigure}
    \begin{subfigure}[b]{0.45\textwidth}
        \centering
        \includegraphics[width=\textwidth]{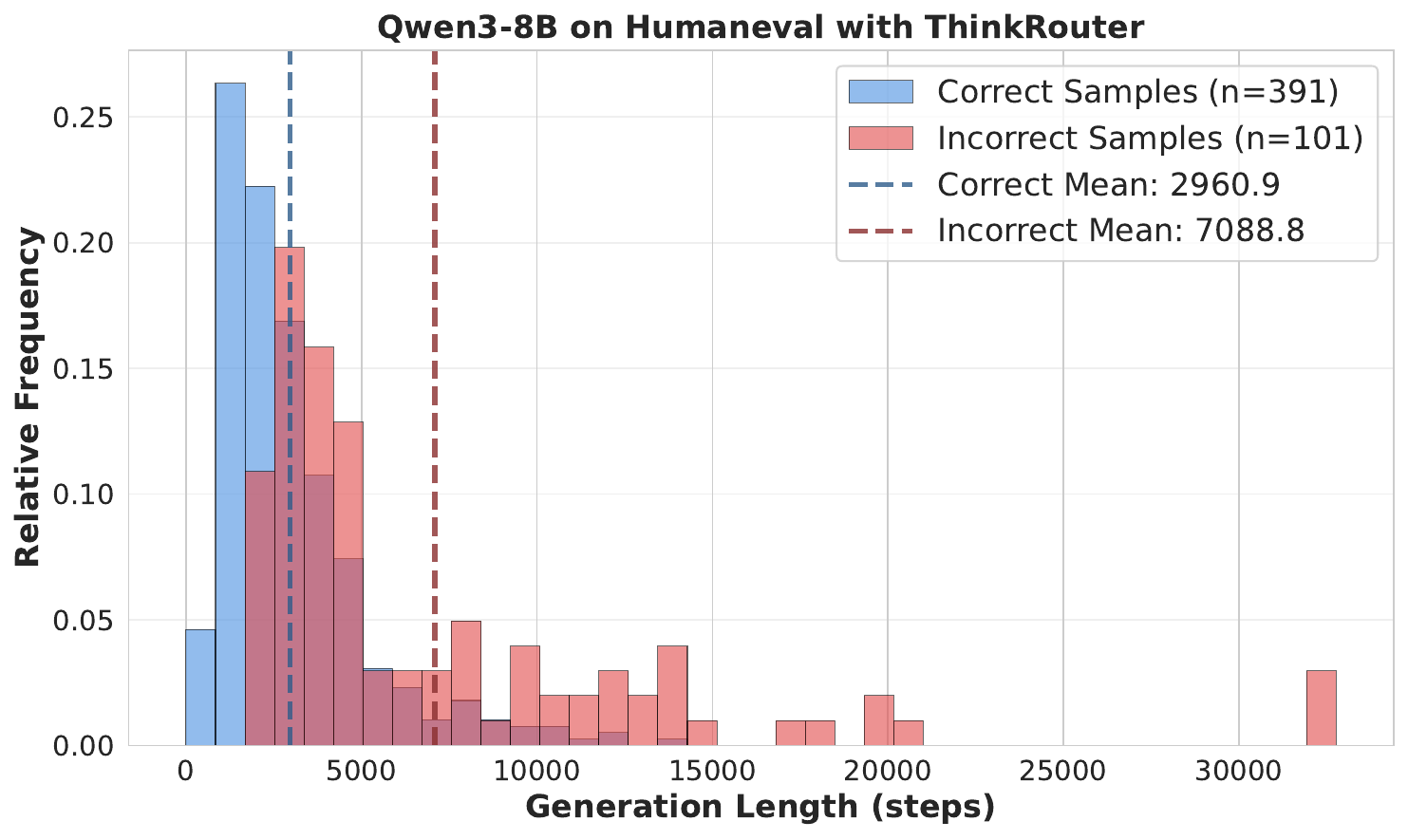}
    \end{subfigure}
    \caption{Distributions of the generation lengths for Qwen3-8B.}
    \label{fig:generation_length_qwen3}
\end{figure}

\begin{figure}
    \centering
    \begin{subfigure}[b]{0.45\textwidth}
        \centering
        \includegraphics[width=\textwidth]{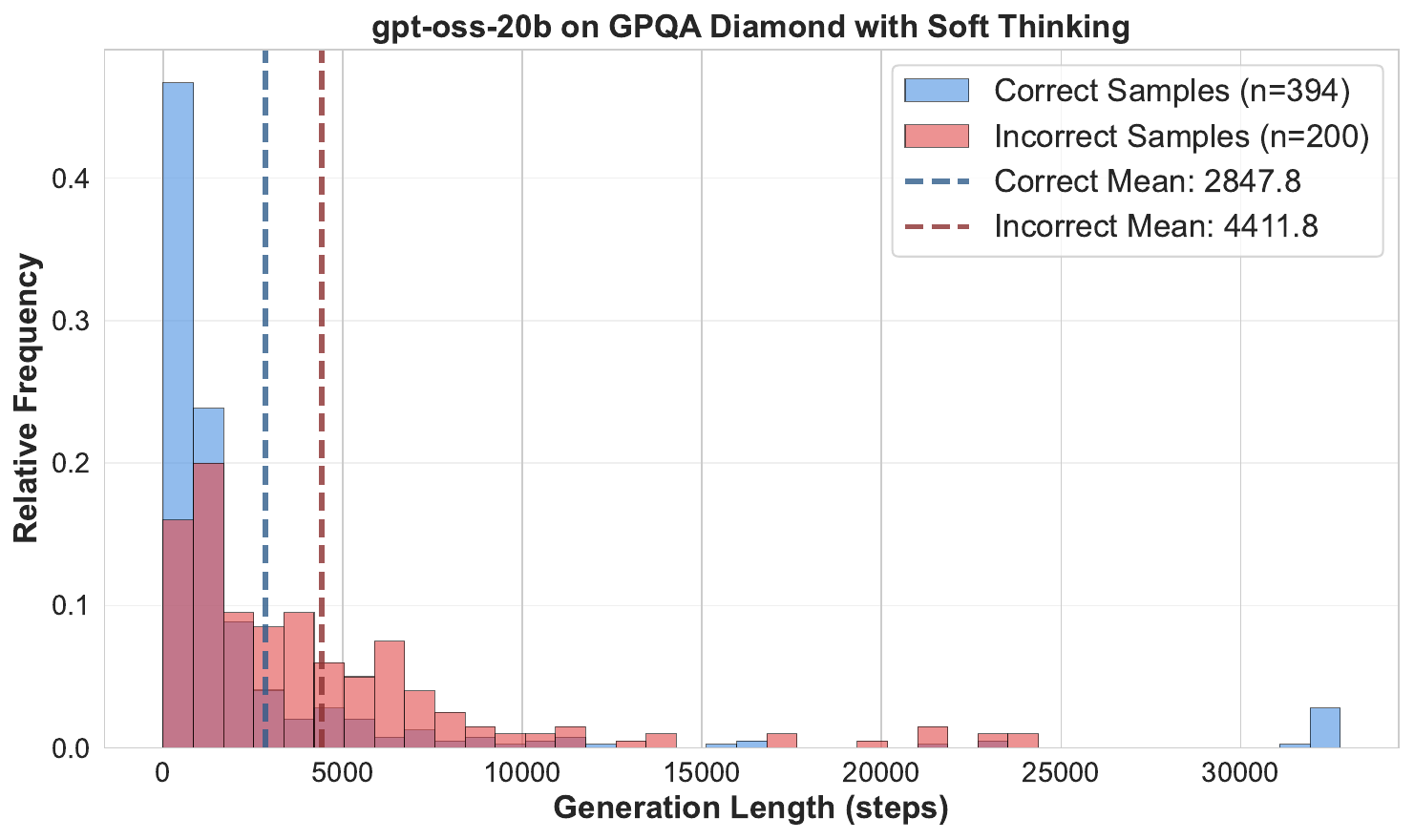}
    \end{subfigure}
    \hspace{0.01\textwidth}
    \begin{subfigure}[b]{0.45\textwidth}
        \centering
        \includegraphics[width=\textwidth]{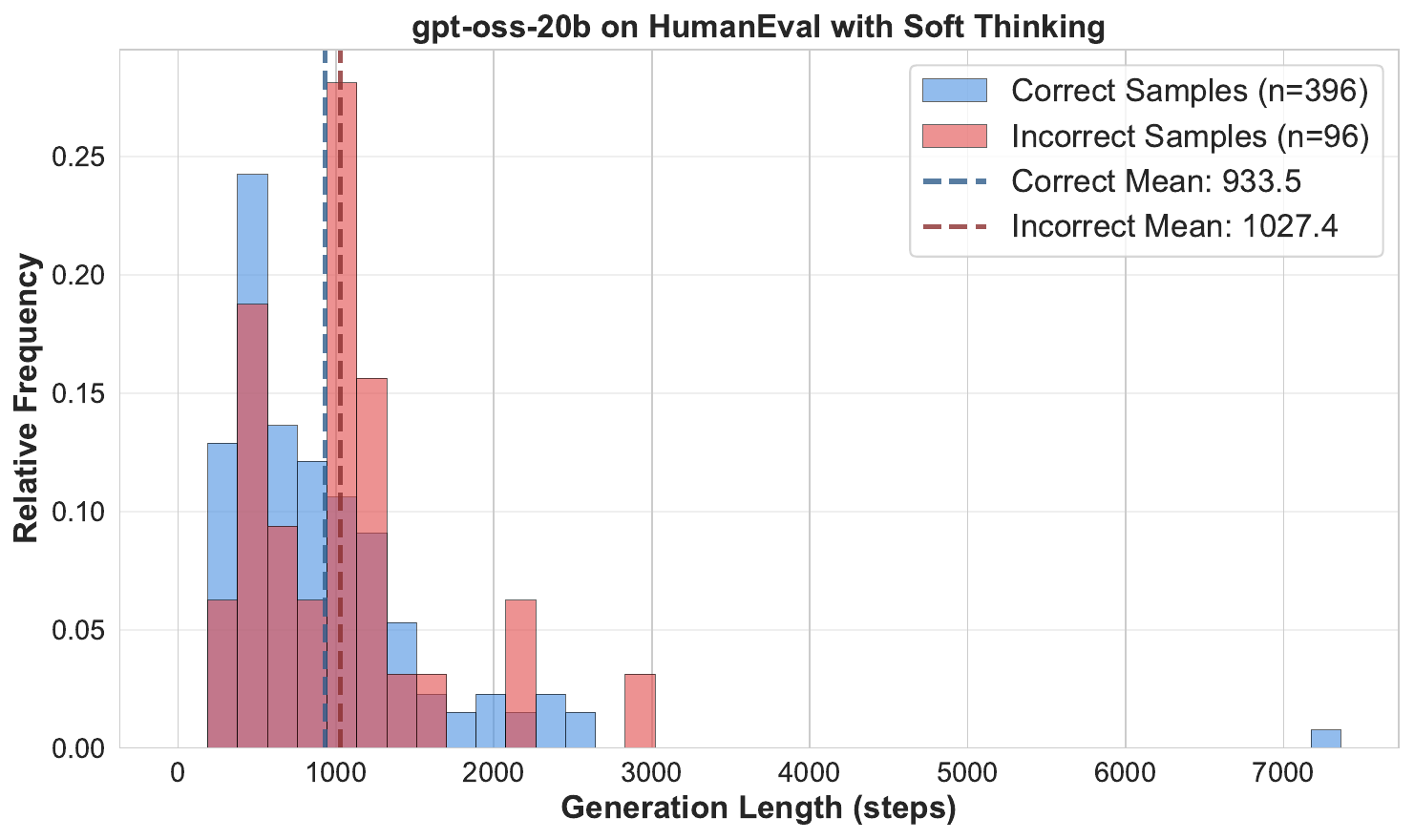}
    \end{subfigure}
    \hspace{0.01\textwidth}
    \begin{subfigure}[b]{0.45\textwidth}
        \centering
        \includegraphics[width=\textwidth]{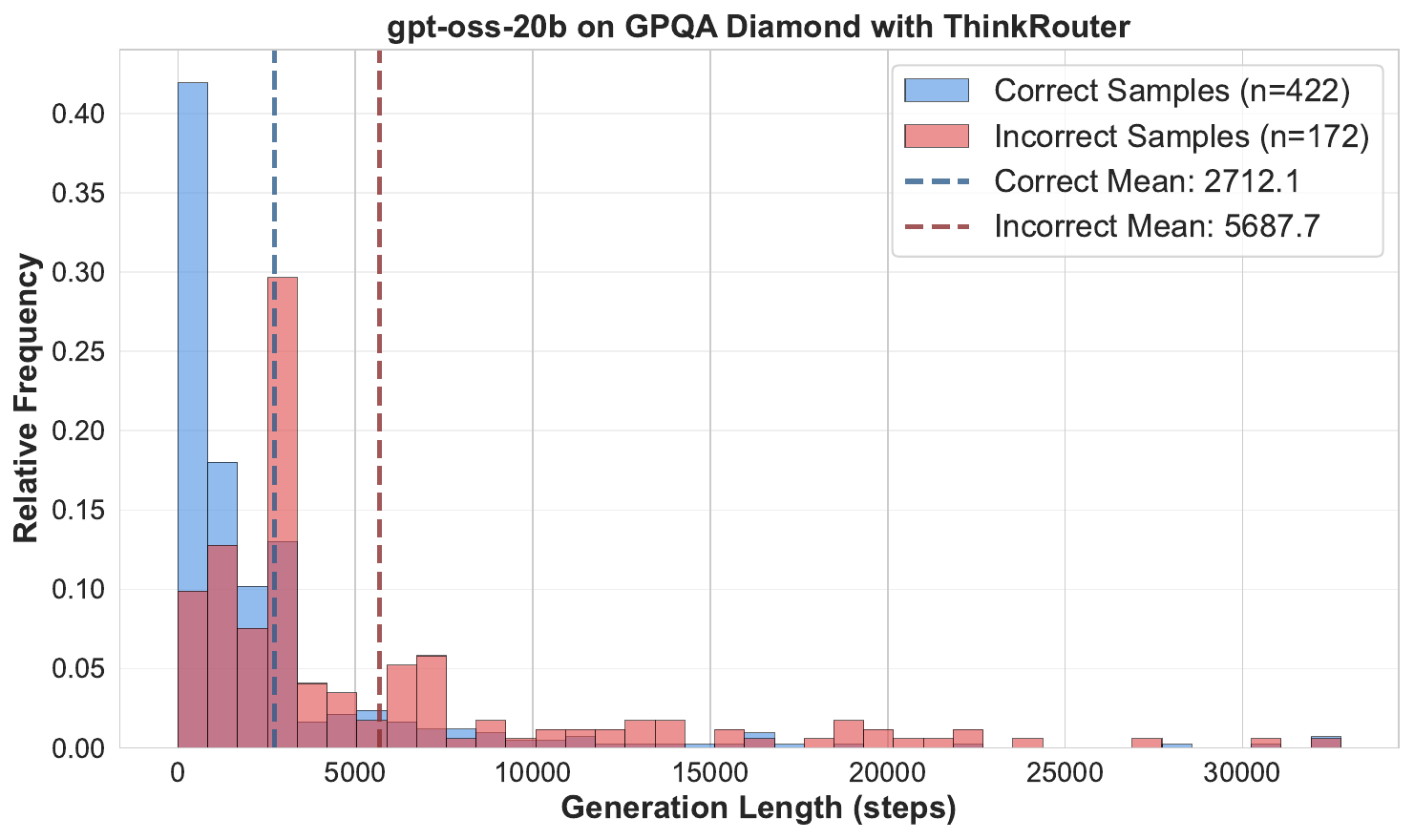}
    \end{subfigure}
    \begin{subfigure}[b]{0.45\textwidth}
        \centering
        \includegraphics[width=\textwidth]{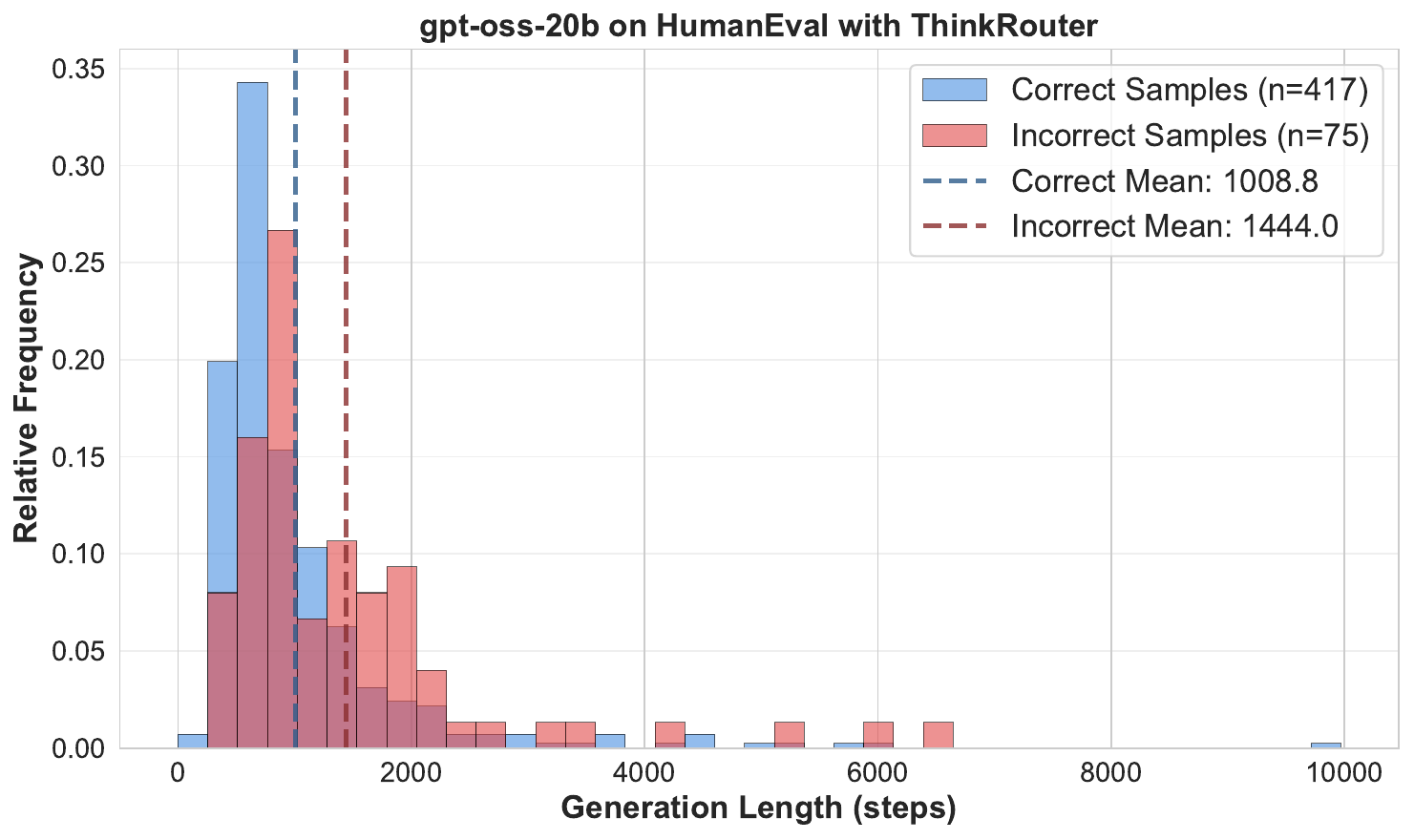}
    \end{subfigure}
    \caption{Distributions of the generation lengths for gptoss.}
    \label{fig:generation_length_gptoss}
\end{figure}

Table \ref{tab:thinking_stop} compare the thinking stop modes of \textit{Soft thinking} and \ours, which illustrates that \ours\ can help trigger EOT token generation in most cases.
Figure \ref{fig:eot2} and \ref{fig:codestop_qwen3} show the maximum next-token probabilities of the last 10 time steps before the EOT token and \textit{Cold Stop}.
Figure \ref{fig:generation_length_qwen3} and \ref{fig:generation_length_gptoss} show the generation length distributions.
We can find that incorrect predictions generally have longer outputs than correct predictions.

\begin{table}[htbp]
  \centering
  \caption{Comparisons of thinking stop modes. EOT: End-of-thinking. }
  \scalebox{0.9}{
    \begin{tabular}{l|cc|cc}
    \toprule
          & \multicolumn{2}{c}{\textbf{Qwen3-8B}} & \multicolumn{2}{c}{\textbf{gpt-oss-20b}} \\
    \midrule
          & GPQA Diamond & HumanEval & \multicolumn{1}{c}{GPQA Diamond} & \multicolumn{1}{c}{HumanEval} \\
    \midrule
    \multicolumn{5}{c}{\textbf{Latent-only Reasoning (Soft Thinking)}} \\
    \midrule
    \midrule
    EOT Token & 79.6\%  & 89.6\%  & 98.5\% & 100.0\% \\
    \textit{Cold Stop}  & 20.4\%  & 10.4\%  & 0.0\%  & 0.0\% \\
    Reached Maximum Output Length & 0.0\%  & 0.0\%  & 1.5\%  & 0.0\% \\
    \midrule
    \multicolumn{5}{c}{\textbf{\ours}} \\
    \midrule
    \midrule
    EOT Token & 98.8\%  & 93.9\%  & 99.2\% & 100.0\%  \\
    \textit{Cold Stop} & 1.2\%   & 6.1\%   & 0.1\%  & 0.0\%  \\
    Reached Maximum Output Length & 0.0\%  & 0.0\%  & 0.7\%  & 0.0\% \\
    \bottomrule
    \end{tabular}}
  \label{tab:thinking_stop}%
\end{table}%

\subsection{Probability Distributions at Routing Times}
\begin{figure*}[ht!]
    \centering
    \includegraphics[width=1.0\linewidth]{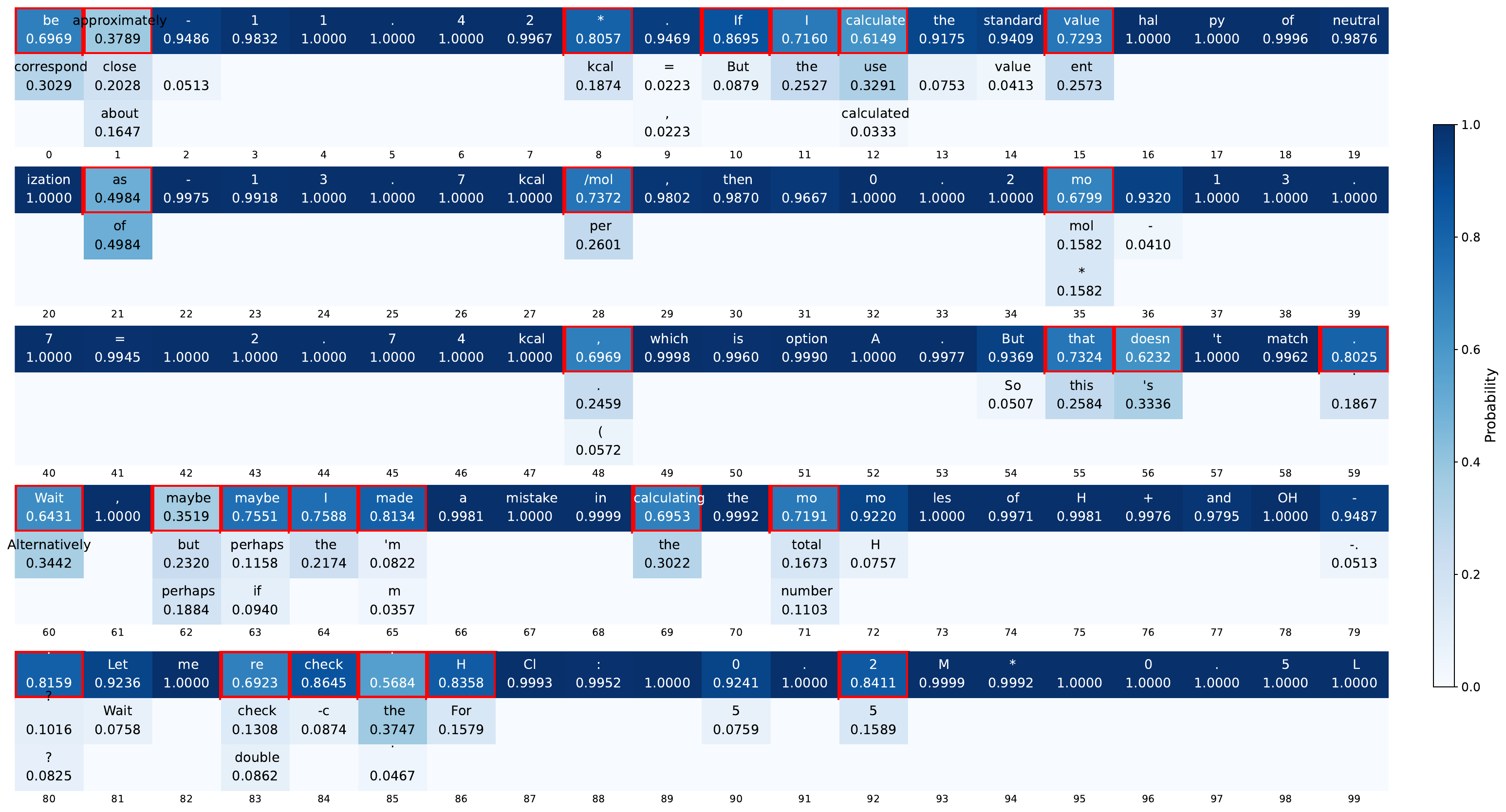}
    \caption{$p_t$ of Qwen3-8B on GPQA Diamond with \ours\ ($\tau$=0.9). \textcolor{red}{Red} boxes indicate routing thinking to the discrete token space; otherwise, to the latent space.}
    \label{fig:prob_qwen3_gpqa}
\end{figure*}

\begin{figure*}[ht!]
    \centering
    \includegraphics[width=1.0\linewidth]{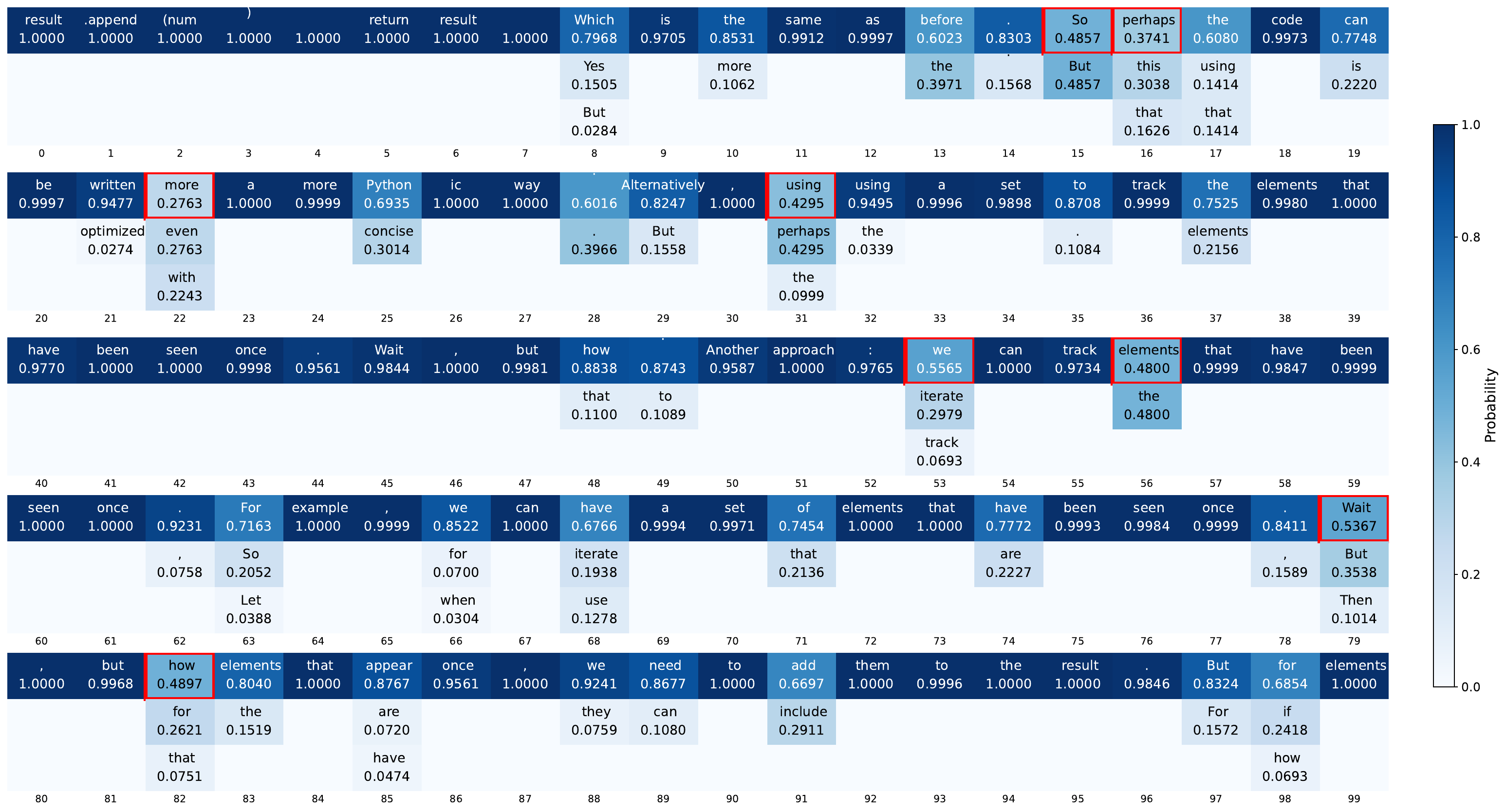}
    \caption{$p_t$ of Qwen3-8B on HumanEval with \ours\ ($\tau$=0.6). \textcolor{red}{Red} boxes indicate routing thinking to the discrete token space; otherwise, to the latent space.}
    \label{fig:prob_qwen3_humaneval}
\end{figure*}

\begin{figure*}[ht!]
    \centering
    \includegraphics[width=1.0\linewidth]{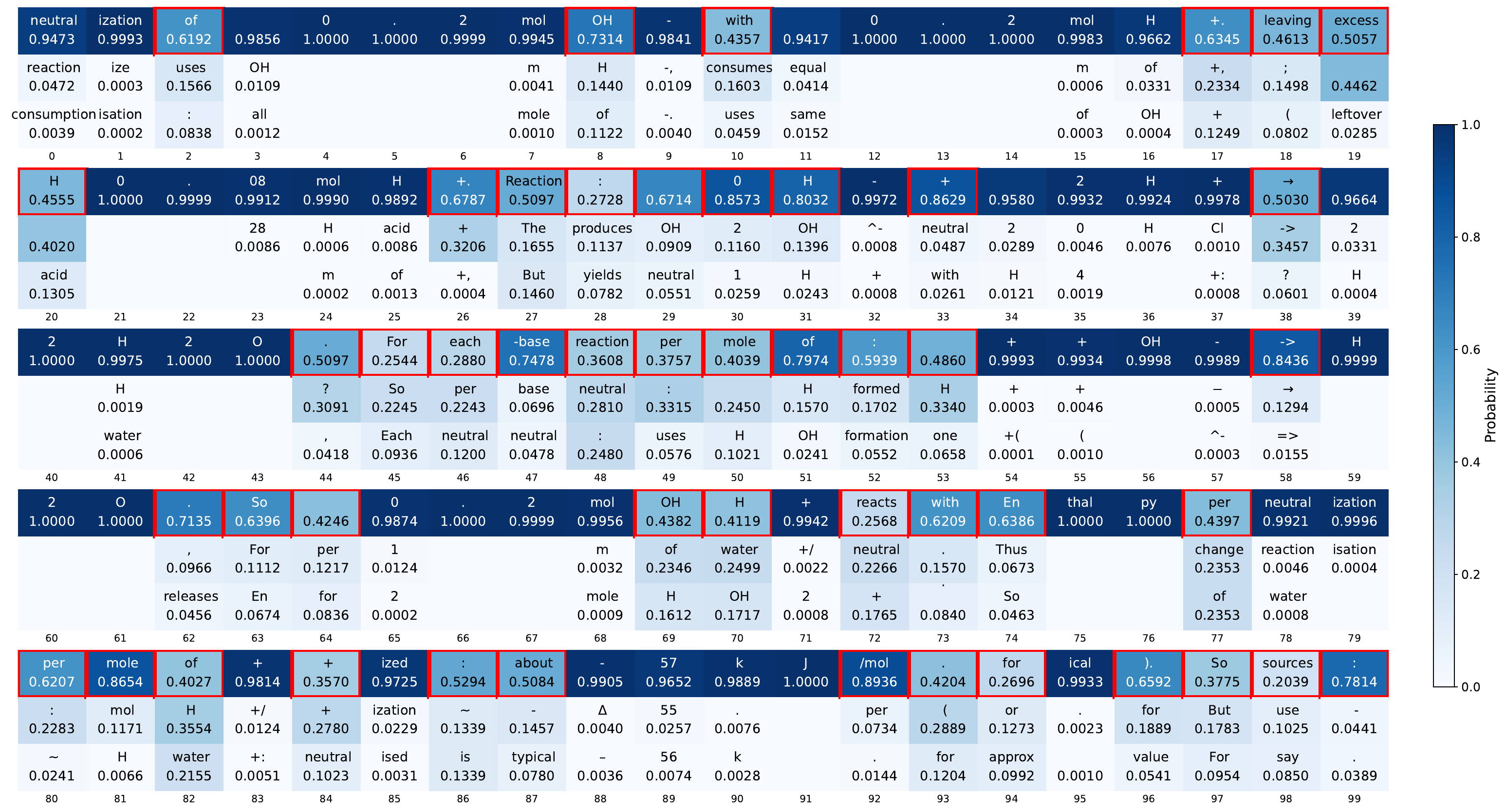}
    \caption{$p_t$ of gpt-oss-20b on GPQA Diamond with \ours\ ($\tau$=0.9). \textcolor{red}{Red} boxes indicate routing thinking to the discrete token space; otherwise, to the latent space.}
    \label{fig:prob_gptoss_humaneval}
\end{figure*}

\begin{figure*}[ht!]
    \centering
    \includegraphics[width=1.0\linewidth]{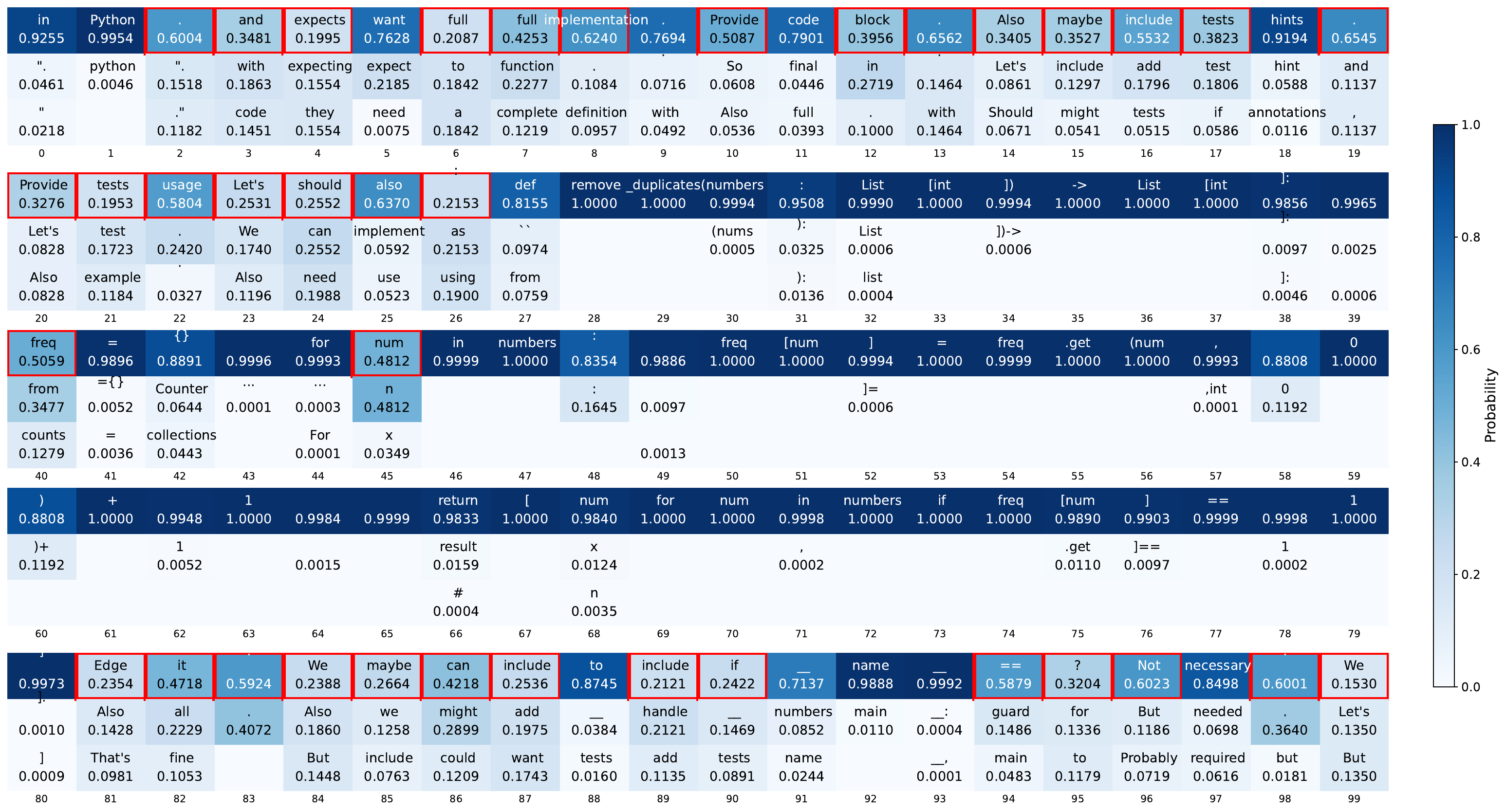}
    \caption{$p_t$ of gpt-oss-20b on HumanEval with \ours\ ($\tau$=0.7). \textcolor{red}{Red} boxes indicate routing thinking to the discrete token space; otherwise, to the latent space.}
    \label{fig:prob_gptoss_gpqa}
\end{figure*}

Here are some examples to show the top-3 next-token probability in $p_t$ along 100 time steps during thinking in Figure \ref{fig:prob_qwen3_gpqa}, \ref{fig:prob_qwen3_humaneval}, \ref{fig:prob_gptoss_humaneval}, and \ref{fig:prob_gptoss_gpqa}.

\end{document}